\documentclass[runningheads]{llncs}

\usepackage[table,xcdraw]{xcolor}

\usepackage{graphicx}

\usepackage{tikz}
\usepackage{comment}
\usepackage{amsmath,amssymb} 
\usepackage{color}

\usepackage[accsupp]{axessibility}  

\usepackage[width=122mm,left=12mm,paperwidth=146mm,height=193mm,top=12mm,paperheight=217mm]{geometry}

\usepackage{hyperref}
\hypersetup{colorlinks=true}
\usepackage{orcidlink}
\usepackage{bm}
\usepackage{booktabs}
\usepackage{multirow}
\usepackage[normalem]{ulem}
\useunder{\uline}{\ul}{}
\usepackage{mathtools}
\usepackage{bbm}
\usepackage{nicefrac}
\usepackage{subcaption}
\newcommand*{\ie}{\textit{i.e. }}
\newcommand*{\eg}{\textit{e.g. }}

\newcommand{\ra}[1]{\renewcommand{\arraystretch}{#1}}
\newcommand*{\bfPar}[1]{{\scriptsize $\ $ \par}\noindent\textbf{#1 }}
\newcommand*{\itPar}[1]{{\tiny $\ $ \par}\noindent\textit{#1 }}
\DeclarePairedDelimiter\abs{\lvert}{\rvert}
\DeclarePairedDelimiter\norm{\lVert}{\rVert}
\graphicspath{ {./images/} }

\begin{document}
\pagestyle{headings}
\mainmatter
\def\ECCVSubNumber{0}  

\title{Minimal Neural Atlas: Parameterizing Complex Surfaces with Minimal Charts and Distortion}

\titlerunning{Minimal Neural Atlas}
%
\author{Weng Fei Low\orcidlink{0000-0001-7022-5713}\index{Low, Weng Fei}\and
Gim Hee Lee\orcidlink{0000-0002-1583-0475}}\index{Lee, Gim Hee}

\authorrunning{W. F. Low and G. H. Lee}
%
\institute{
Institute of Data Science (IDS), National University of Singapore\\
NUS Graduate School's Integrative Sciences and Engineering Programme (ISEP)\\
Department of Computer Science, National University of Singapore\\
\email{\{wengfei.low, gimhee.lee\}@comp.nus.edu.sg}\\
\url{https://github.com/low5545/minimal-neural-atlas}
}
\maketitle

\begin{abstract}
Explicit neural surface representations allow for exact and efficient extraction of the encoded surface at arbitrary precision, as well as analytic derivation of differential geometric properties such as surface normal and curvature. Such desirable properties, which are absent in its implicit counterpart, makes it ideal for various applications in computer vision, graphics and robotics. However, SOTA works are limited in terms of the topology it can effectively describe, distortion it introduces to reconstruct complex surfaces and model efficiency. In this work, we present \textit{Minimal Neural Atlas}, a novel atlas-based explicit neural surface representation. At its core is a fully learnable parametric domain, given by an implicit probabilistic occupancy field defined on an open square of the parametric space. In contrast, prior works generally predefine the parametric domain. The added flexibility enables charts to admit arbitrary topology and boundary. Thus, our representation can learn a minimal atlas of 3 charts with distortion-minimal parameterization for surfaces of arbitrary topology, including closed and open surfaces with arbitrary connected components. Our experiments support the hypotheses and show that our reconstructions are more accurate in terms of the overall geometry, due to the separation of concerns on topology and geometry.

\end{abstract}

\section{Introduction}
\label{sec:intro}

\par
An explicit neural surface representation that can faithfully describe surfaces of arbitrary topology at arbitrary precision is highly coveted for various downstream applications. This is attributed to some of its intrinsic properties that are absent in implicit neural surface representations.

\par
Specifically, the explicit nature of such representations entail that the encoded surface can be sampled exactly and efficiently, irrespective of its scale and complexity. This is particularly useful for inference-time point cloud generation, mesh generation and rendering directly from the representation. In contrast, implicit representations rely on expensive and approximate isosurface extraction and ray casting. Furthermore, differential geometric properties of the surface can also be derived analytically in an efficient manner \cite{bednarik2020_dsp}. Some notable examples of such properties include surface normal, surface area, mean curvature and Gaussian curvature. Implicit neural representations can at most infer such quantities at approximated surface points. Moreover, explicit representations are potentially more scalable since a surface is merely an embedded 2D submanifold of the 3D Euclidean space.

\par
Despite the advantages of explicit representations, implicit neural surface representations have attracted most of the research attention in recent years. Nevertheless, this is not unwarranted, given its proven ability to describe general surfaces at high quality and aptitude for deep learning. This suggests that explicit representations still have a lot of potential yet to be discovered. In this work, we aim to tackle various shortcomings of existing explicit neural representations, in an effort to advance it towards the goal of a truly faithful surface representation.

\par
State-of-the-art explicit neural surface representations \cite{groueix2018_atlasnet,bednarik2020_dsp,deng2020_bps,pang2021_tearingnet} mainly consists of neural \textit{atlas}-based representations, where each \textit{chart} is given by a \textit{parameterization} modeled with neural networks, as well as a predefined open square \textit{parametric domain}. 
In other words, such representations describe a surface with a collection of neural network-deformed planar square patches.

\par
In theory, these representations cannot describe surfaces of arbitrary topology, especially for surfaces with arbitrary connected components. This is clear from the fact that an atlas with 25 deformed square patches cannot represent a surface with 26 connected components. In practice, these works also cannot faithfully represent single-object or single-connected component surfaces of arbitrary topology, although it is theoretically capable given sufficient number of charts. Furthermore, these atlas-based representations generally admits a distortion-minimal surface parameterization at the expense of representation accuracy. For instance, distortion is inevitable to deform a square patch into a circular patch. Some of these works also require a large number of charts to accurately represent general surfaces, which leads to a representation with low model efficiency.

\par
The root cause of all limitations mentioned above lies in predefining the parametric domain, which unnecessarily constrains its boundary and topology, and hence also that of the chart. While \cite{pang2021_tearingnet} has explored 
``tearing'' an initial open square parametric domain at regions of high distortion, the limitation on distortion remains unaddressed. Our experiments also show that its reconstructions still incur a relatively high topological error on general single-object surfaces.

\bfPar{Contributions.}
We propose a novel representation, 
\textit{Minimal Neural Atlas}, 
where the core idea is to model the parametric domain of each chart via an implicit \textit{probabilistic occupancy field} \cite{mescheder2019_onet} defined on the $(-1, 1)^2$ open square of the \textit{parametric space}. 
As a result, each chart is free to admit any topology and boundary, as we only restrict the bounds of the parametric domain. This enables the learning of a distortion-minimal parameterization, which is important for high quality texture mapping and efficient uniform point cloud sampling. A separation of concerns can also be established between the occupancy field and parameterization, where the former focuses on topology and the latter on geometry and distortion. 
This enables the proposed representation to describe surfaces of arbitrary topology, including closed and open surfaces with arbitrary connected components, using a \textit{minimal atlas} of 3 charts. Our experiments on ShapeNet and CLOTH3D++ support this theoretical finding and show that our reconstructions are more accurate in terms of the overall geometry.

\section{Related Work}
\label{sec:related}

\par
Point clouds, meshes and voxels have long been the \textit{de facto} standard for surface representation. Nonetheless, these \textit{discrete surface representations} describe the surface only at sampled locations with limited precision. First explored in \cite{groueix2018_atlasnet,yang2018_foldingnet}, \textit{neural surface representations} exploits the universal approximation capabilities of neural networks to describe surfaces continuously at a low memory cost.

\bfPar{Explicit Neural Surface Representations.}
Such representations provide a closed form expression describing exact points on the surface. \cite{groueix2018_atlasnet,yang2018_foldingnet} first proposed to learn an atlas for a surface by modeling the chart parameterizations with a neural network and predefining the parametric domain of each chart to the open unit square. Building on \cite{groueix2018_atlasnet}, \cite{bednarik2020_dsp} introduced novel training losses to regularize for chart degeneracy, distortion and the amount of overlap between charts. \cite{deng2020_bps} additionally optimizes for the quality of overlaps between charts. Such atlas-based representations have also been specialized for surface reconstruction \cite{williams2019_dgp,badki2020_meshlet,morreale2021_neural_surface_maps}. However, these representations suffer from various limitations outlined in Sec.~\ref{sec:intro}, as a consequence of predefining the parametric domain. Hence, \cite{pang2021_tearingnet} proposed to adapt to the target surface topology by “tearing” an initial unit square parametric domain. In addition to the drawbacks mentioned in Sec.~\ref{sec:intro}, this single-chart atlas representation also theoretically cannot describe general single-object surfaces. Moreover, the optimal tearing hyperparameters are instance-dependent, as they are determined by the scale, sampling density and area of the surface, which cannot be easily normalized. 



\bfPar{Implicit Neural Surface Representations.}
These representations generally encode the surface as a level set of a scalar field defined on the 3D space, which is parameterized by a neural network. Some of the first implicit representations proposed include the \textit{Probabilistic Occupancy Field} (POF) \cite{mescheder2019_onet,chen2019_im-net} and \textit{Signed Distance Field} (SDF) \cite{park2019_deepsdf}. These representations can theoretically describe closed surfaces of arbitrary topology and they yield accurate watertight reconstructions in practice. However, these works require ample access to watertight meshes for training, which might not always be possible. 
\cite{boulch2021_needrop,gropp2020_igr,atzmon2020_sal,atzmon2021_sald,ma2021_neural_pull} proposed various approaches to learn such representations from unoriented point clouds. Nevertheless, POF and SDF are only restricted to representing closed surfaces. 
\cite{chibane2020_ndf} proposed to model an \textit{Unsigned Distance Field} (UDF) so that both open and closed surfaces can be represented as the zero level set. While this is true in theory, surface extraction is generally performed with respect to a small epsilon level set, which leads to a double or crusted surface, since there is no guarantee that the zero level set exists in practice. Consequently, UDF cannot truly represent general surfaces. 




\section{Our Method}
\label{sec:method}

\begin{figure}[t]
\centering 
		\includegraphics[width=1\linewidth,trim={1.2cm 0.3cm 1.9cm 0.4cm},clip]{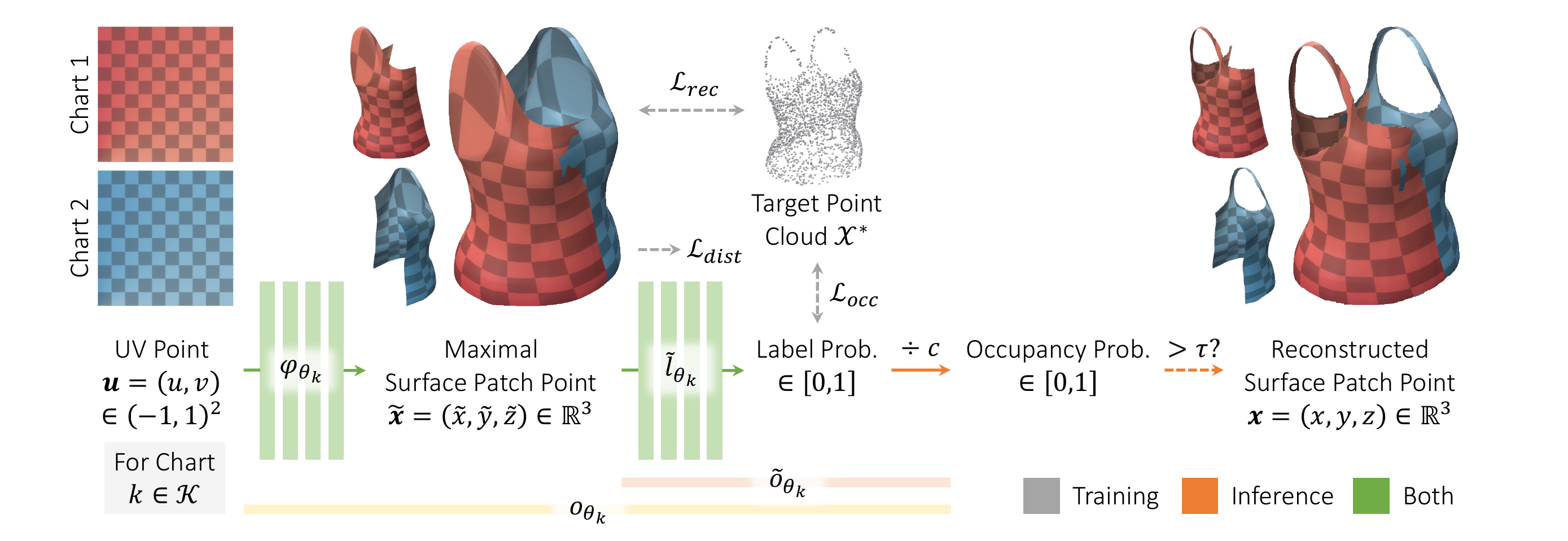}
        \caption{Overview of our proposed method. }
        \label{fig:method:pipeline}
\end{figure}

\par
We first present our proposed surface representation and its theoretical motivation (Sec.~\ref{subsec:method:background},~\ref{subsec:method:surface_repr}). Next, we detail how to learn this representation (Sec.~\ref{subsec:method:training}) and describe an approach for extracting point clouds and meshes of a specific size during inference (Sec.~\ref{subsec:method:inference}). An illustration of our method is given in Fig.~\ref{fig:method:pipeline}.

\subsection{Background}
\label{subsec:method:background}

\par
A \textit{manifold} $\mathcal{M}$ is a topological space that locally resembles an Euclidean space. A \textit{surface} $\mathcal{S}$ is merely a 2-dimensional manifold, or 2-manifold in short. In general, a manifold can be explicitly described using an \textit{atlas}, which consists of \textit{charts} that each describe different regions of the manifold. Formally, a chart on an $n$-manifold $\mathcal{M}$ can be denoted by an ordered pair $(U, \varphi)$, whereby $U \subset \mathbb{R}^n$ is an open subset of the $n$-dimensional Euclidean space and $\varphi_k : U \mapsto \mathcal{M}$ is a \textit{homeomorphism} or \textit{parameterization} from $U$ to an open subset of $\mathcal{M}$. An atlas for $\mathcal{M}$ is given by an indexed family of charts $\{ (U_k, \varphi_k) \mid k \in \mathcal{K} \}$ which forms an \textit{open cover} of $\mathcal{M}$ (\ie $\bigcup_{k \in \mathcal{K}} \varphi_k(U_k) = \mathcal{M}$).

\par
It is well-known that the \textit{Lusternik-Schnirelmann category} \cite{fox1941_ls_cat1,james1978_ls_cat2,cornea2003_ls_cat3} of a general $n$-manifold is at most $n+1$. This implies that irrespective of its complexity, a general $n$-manifold always admits an atlas of $n+1$ charts. 
Consequently, this defines the notion of a \textit{minimal atlas} for a general $n$-manifold. 

\subsection{Surface Representation}
\label{subsec:method:surface_repr}

\par
Motivated by such a theoretical guarantee, we propose to represent general surfaces $\mathcal{S}$ with a minimal atlas of 3 charts modeled using neural networks. Specifically, we model the surface parameterization of each chart $k$ with a \textit{Multi-Layer Perceptron} (MLP) parameterized by $\theta_k$, which we denote as $\varphi_{\theta_k}$. Furthermore, we employ a \textit{probabilistic occupancy field} \cite{mescheder2019_onet} defined on the $\mathbb{R}^2$ \textit{parametric space} to implicitly model the \textit{parametric domain} $U_{\theta_k}$ of each chart $k$.

\par
More precisely, we model a probabilistic occupancy field $o_{\theta_k}$ with an MLP parameterized by $\theta_k$ on the $(-1, 1)^2$ open square of the parametric space. This allows us to implicitly represent the parametric domain $U_{\theta_k}$ as regions in the parametric space with occupancy probability larger than a specific threshold $\tau$, or \textit{occupied} regions in short. 
Our proposed \textit{Minimal Neural Atlas} surface representation is formally given as:
\begin{equation}
	\{ (U_{\theta_k}, \varphi_{\theta_k}) \mid k \in \mathcal{K} \} \ ,
\end{equation}
where:
\begin{align}
	U_{\theta_k} &= \{ \bm{u} \in (-1, 1)^2 \mid o_{\theta_k} (\bm{u}) > \tau \} \ ,\\
   \varphi_{\theta_k} &: U_{\theta_k} \mapsto \mathcal{S} \ ,\\
   o_{\theta_k} &: (-1, 1)^2 \mapsto [0, 1] \ .
\end{align}
\par
While we have formulated the proposed representation in the context of representing a single surface, conditioning the representation on a \textit{latent code} $z \in \mathcal{Z}$ encoding any surface of interest facilitates the modeling of a family of surfaces. The latent code $z$ can be inferred from various forms of inputs describing the associated surface, such as a point cloud or an image, via an appropriate encoder.

\par
The key component that contrasts this atlas-based representation from the others is the flexibility of the parametric domain. 
In contrast to predefining the parametric domain, we only restrict its bounds. This eliminates redundant constraints on the boundaries and topology of the parametric domain and hence 
the chart. 
As a result, the proposed representation can learn a minimal atlas for general surfaces with arbitrary topology, including closed and open surfaces with arbitrary connected components. This also enables the learning of a distortion-minimal surface parameterization. 
A \textit{separation of concerns} can thus be achieved, where $o_{\theta_k}$ mainly addresses the concern of discovering and representing the appropriate topology, and $\varphi_{\theta_k}$ addresses the concern of accurately representing the geometry with minimum distortion.

\bfPar{Decoupling Homeomorphic Ambiguity.}
Learning a minimal neural atlas in the present form possesses some difficulties. For a given \textit{surface patch} described by a chart $(U, \varphi)$, there exists infinitely many other charts $(U', \varphi')$ such that $U' = \phi(U)$ 
and $\varphi' = \varphi \circ \phi^{-1}$, where $\phi$ is a homeomorphism in the open square of the parametric space, that can describe 
the same surface patch. This statement is true because $\varphi' (U') = (\varphi \circ \phi^{-1}) (\phi(U)) = \varphi (U)$. 
This coupled ambiguity of $\phi$ presents a great challenge during the learning of the two relatively independent components $o_{\theta_k}$ and $\varphi_{\theta_k}$.

\par
To decouple this homeomorphic ambiguity, we reformulate $o_{\theta_k}$ as:
\begin{equation}
\label{eq:o_reform}
	o_{\theta_k} = \tilde o_{\theta_k} \circ \varphi_{\theta_k} \ ,
\end{equation}
where:
\begin{equation}
	\tilde o_{\theta_k} : \varphi_{\theta_k} ((-1, 1)^2) \mapsto [0, 1]
\end{equation}
is an auxiliary probabilistic occupancy field defined on the \textit{maximal surface patch} \linebreak $\varphi_{\theta_k} ((-1, 1)^2) \subset \mathbb{R}^3$. This also requires us to extend the domain and codomain of $\varphi_{\theta_k}$ to the open square and $\mathbb{R}^3$, respectively. Nonetheless, this is just a matter of notation since $\varphi_{\theta_k}$ is modeled using an MLP with a \textit{natural domain} of $\mathbb{R}^2$ and 
codomain of $\mathbb{R}^3$. Under this reformulation that conditions $o_{\theta_k}$ on $\varphi_{\theta_k}$, $\tilde o_{\theta_k}$ can be learned such that it is invariant to ambiguities in $\phi$. Particularly, since the same surface patch is described irrespective of the specific $\phi$, it is sufficient for $\tilde o_{\theta_k}$ to be occupied only within that surface patch and \textit{vacant} elsewhere (\ie ``trim away'' arbitrary \textit{surface patch excess}). This enables the learning of $\varphi_{\theta_k}$ with arbitrary $\phi$ that is independent of $\tilde o_{\theta_k}$.

\subsection{Training}
\label{subsec:method:training}

\par
To learn the minimal neural atlas of a target surface $\mathcal{S}^*$, we only assume that we are given its raw unoriented point cloud during training, which we denote as the set $\mathcal{X}^*$. For training, we uniformly sample a common fixed number of points or \textit{UV samples} in the open square of each chart $k$ to yield the set $\mathcal{V}_k$.

\par
Due to the lack of minimal atlas annotations (\eg target point cloud for each chart of a minimal atlas), a straightforward supervision of the surface parameterization and (auxiliary) probabilistic occupancy field for each chart is not possible. To mitigate this problem, we introduce the reconstruction loss $\mathcal{L}_{rec}$, occupancy loss $\mathcal{L}_{occ}$ and metric distortion loss $\mathcal{L}_{dist}$.
Without a loss of generality, the losses are presented similar to Sec.~\ref{subsec:method:surface_repr} in the context of fitting a single target surface. The total training loss is then given by their weighted sum:
\begin{equation}
	\mathcal{L} = \lambda_{rec} \mathcal{L}_{rec} + \lambda_{occ} \mathcal{L}_{occ} + \lambda_{dist} \mathcal{L}_{dist} \ ,
\end{equation}
where $\lambda_{rec}, \lambda_{occ}$ and $\lambda_{dist}$ are the hyperparameters to balance the loss terms. 

\bfPar{Reconstruction Loss.}
The concern of topology is decoupled from the parameterization in our proposed surface representation. As a result, we can ensure that the geometry of the target surface is accurately represented as long as the \textit{maximal surface} $\tilde{\mathcal{S}}$, given by the collection of all maximal surface patches $\bigcup_{k \in \mathcal{K}} \varphi_{\theta_k}((-1, 1)^2)$, forms a \textit{cover} of the target surface $\mathcal{S}^*$. To this end, we regularize the surface parameterization of each chart with the unidirectional \textit{Chamfer Distance} \cite{fan2017_psgn} that gives the mean squared distance of the target point cloud and its \textit{maximal surface point cloud} $\bigcup_{k \in \mathcal{K}} \varphi_{\theta_k}(\mathcal{V}_k)$ nearest neighbor:
\begin{equation}
	\mathcal{L}_{rec} = \frac{1}{\abs{\mathcal{X}^*}} \sum_{\bm{x}^* \in \mathcal{X}^*} \min_{k \in \mathcal{K}} \min_{\bm{u} \in \mathcal{V}_k} \norm{\bm{x}^* - \varphi_{\theta_k}(\bm{u})}^2_2 \ .
\end{equation}

\bfPar{Occupancy Loss.}
To truly represent the target surface with the correct topology, it is necessary for the auxiliary probabilistic occupancy field of each chart $\tilde o_{\theta_k}$ to ``trim away'' only the \textit{surface excess} given by $\tilde{\mathcal{S}}\setminus \mathcal{S}^*$.

\itPar{\textbf{Naïve Binary Classification Formulation.}}
This 
is achieved by enforcing an occupancy of `1' at the nearest neighbors of the target point cloud, and `0' at other non-nearest neighbor maximal surface points, which effectively casts the learning of $\tilde o_{\theta_k}$ as a binary classification problem. However, this form of annotation 
incorrectly assigns an occupancy of `0' at some maximal points which also form the target surface. Such mislabeling can be attributed to the difference in sampling density as well as distribution between the target and maximal surface, and effects of random sampling on the nearest neighbor operator.

\itPar{\textbf{Positive-Unlabeled Learning Formulation.}}
Instead of the interpretation of 
mislabeling, 
we can take an alternative view of partial labeling. Specifically, 
a maximal point annotated with a label of `1' is considered as a labeled positive (occupied) sample and a maximal point annotated with a label of `0' is considered as an unlabeled sample 
instead of a labeled negative (vacant) sample. This interpretation allows us to cast the learning of $\tilde o_{\theta_k}$ 
as a \textit{Positive and Unlabeled Learning} (PU Learning) problem \cite{elkan2008_pu_learning,bekker2020_pu_learning_survey} (also called \textit{learning from positive and unlabeled examples}).

\par
Our labeling mechanism satisfies the \textit{single-training-set scenario} \cite{elkan2008_pu_learning,bekker2020_pu_learning_survey} since the maximal points are \textit{independent and identically distributed} (i.i.d.) on the maximal surface and are either labeled positive (occupied) or unlabeled to form the ``training set''. Following \cite{elkan2008_pu_learning}, we assume that our labeling mechanism satisfies the \textit{Selected Completely at Random} (SCAR) assumption, which entails that the labeled maximal points are i.i.d. to, or selected completely at random from, the maximal points on the target surface. Under such an assumption, the auxiliary probabilistic occupancy field defined on can be factorized as:
\begin{equation}
\label{eq:o-tilde_factor}
	\tilde o_{\theta_k} (\tilde{\bm{x}}) = \frac{\tilde l_{\theta_k} (\tilde{\bm{x}})}{c}\ ,
\end{equation}
where:
\begin{equation}
	\tilde l_{\theta_k}: \varphi_{\theta_k} ((-1, 1)^2) \mapsto [0, 1]
\end{equation}
returns the probability that the given maximal point $\tilde{\bm{x}}$ is labeled, and $c$ is the constant probability that a maximal point on the target surface is labeled. In the PU learning literature, $\tilde l_{\theta_k}$ and $c$ are referred to as a \textit{non-traditional classifier} and \textit{label frequency} respectively. Note that $c$ is proportional to the relative sampling density between the target and maximal point cloud.

\par
$\tilde l_{\theta_k}$ can 
now be learned in the standard supervised binary classification setting with the \textit{Binary Cross Entropy} (BCE) loss as follows:
\begin{equation}
\begin{split}
	\mathcal{L}_{occ} = -\frac{1}{\sum_{k \in \mathcal{K}} \abs{\mathcal{V}_k}} \sum_{k \in \mathcal{K}} \sum_{\bm{u} \in \mathcal{V}_k} \mathrm{BCE} (\mathbbm{1}_{\mathcal{V}_k^*} (\bm{u}), \ \tilde l_{\theta_k} \circ \varphi_{\theta_k}(\bm{u})) \ ,
\end{split}
\end{equation}
where $\mathcal{V}_k^* \subseteq \mathcal{V}_k$ is the set of UV samples corresponding to the target point cloud nearest neighbors. As a result of the reformulation of the probabilistic occupancy field, it can be observed that the surface parameterization of each chart directly contributes to the occupancy loss. 
In practice, we prevent the backpropagation of the occupancy loss gradients to the surface parameterizations. This enables the parameterizations to converge to a lower reconstruction loss since they are now decoupled from the minimization of the occupancy loss. Furthermore, this also facilitates the separation of concerns between $o_{\theta_k}$ and $\varphi_{\theta_k}$.

\bfPar{Metric Distortion Loss.}
To learn a minimal neural atlas with distortion-minimal surface parameterization, we explicitly regularize the parameterization of each chart to preserve the \textit{metric} of the parametric domain, up to a common scale. 
We briefly introduce some underlying concepts before going into the details of the loss function. 

\par
Let $J_k (\bm{u}) = \begin{bmatrix} \nicefrac{\partial \varphi_{\theta_k}}{\partial u} & \nicefrac{\partial \varphi_{\theta_k}}{\partial v} \end{bmatrix}$, where $\bm{u} = \begin{bmatrix} u & v \end{bmatrix}^\top$, be the Jacobian of the surface parameterization of chart $k$. It describes the tangent space of the surface at the point $\varphi_{\theta_k} (\bm{u})$. The \textit{metric tensor} or \textit{first fundamental form} $g_k (\bm{u}) = J_k (\bm{u})^\top J_k (\bm{u})$ enables the computation of various differential geometric properties, such as length, area, normal, curvature and distortion.

\par
To quantify metric distortion up to a specific common scale of $L$, we adopt a scaled variant of the \textit{Symmetric Dirichlet Energy} (SDE) \cite{schreiner2004_sde_real2,smith2015_sde_real,rabinovich_sde}, which is an isometric distortion energy, given by:
\begin{equation}
	\frac{1}{\sum_{k \in \mathcal{K}} \abs{\mathcal{W}_k}} \sum_{k \in \mathcal{K}} \sum_{\bm{u} \in \mathcal{W}_k} \frac{1}{L^2} \mathrm{trace} (g_k (\bm{u})) + L^2 \mathrm{trace} (g_k (\bm{u})^{-1}) \ ,
\end{equation}
where the distortion is quantified with respect to the set of UV samples denoted as $\mathcal{W}_k$. We refer this metric distortion energy as the \textit{Scaled Symmetric Dirichlet Energy} (SSDE). As the SSDE reduces to the SDE when $L=1$, the SSDE can be alternatively interpreted as the SDE of the derivative surface parameterization, given by post-scaling the parameterization of interest by a factor of $\nicefrac{1}{L}$.

\par
Since we are interested in enforcing metric preservation up to an arbitrary common scale, it is necessary to deduce the optimal scale $L^*$ of the SSDE, for any given atlas (hence given $g_k$). To this end, we determine the $L^*$ by finding the $L$ that minimizes the SSDE. As the SSDE is a convex function of $L$, its unique global minimum can be analytically derived. 
Finally, the metric distortion loss used to learn a minimal neural atlas with distortion-minimal parameterization is simply given by:
\begin{equation}
\label{eq:ssde_Lstar}
	\mathcal{L}_{dist} = 2 \sqrt{ \mathrm{mean}_{\mathcal{V}^*} (\mathrm{trace} \circ g_k) \ \mathrm{mean}_{\mathcal{V}^*} (\mathrm{trace} \circ g_k^{-1}) } \ ,
\end{equation}
where:
\begin{equation}
	\mathrm{mean}_{\mathcal{W}} (f) = \frac{1}{\sum_{k \in \mathcal{K}} \abs{\mathcal{W}_k}} \sum_{k \in \mathcal{K}} \sum_{\bm{u} \in \mathcal{W}_k} f (\bm{u}) \ .
\end{equation}
\par
Note that we only regularize UV samples corresponding to nearest neighbors of the target point cloud, which are labeled as occupied. This provides more flexibility to the parameterization outside of the parametric domain. While \cite{bednarik2020_dsp} has proposed a novel loss to minimize metric distortion, our metric distortion loss is derived from the well-established SDE, which quantifies distortion based on relevant fundamental properties of the metric tensor, rather than its raw structure. We also observe better numerical stability as $\mathcal{L}_{dist}$ is given by the geometric mean of two values roughly inversely proportional to each other.

\subsection{Inference}
\label{subsec:method:inference}

\bfPar{Label Frequency Estimation.}
The label frequency $c$ can be estimated during inference with the \textit{positive subdomain} assumption \cite{bekker2020_pu_learning_survey}. This requires the existence of a subset of the target surface that is uniquely covered by a chart, which we assume to be true. We refer such regions as \textit{chart interiors} since they are generally far from the chart boundary, where overlapping between charts occur. 

\par
Similar to training, we uniformly sample the open square to infer a set of maximal surface points. Given that $\tilde l_{\theta_k}$ is well-calibrated \cite{elkan2008_pu_learning}, maximal points on a chart interior have a $\tilde l_{\theta_k}$ value of $c$ under the positive subdomain assumption. 
In practice, we identify such points by assuming at least $\eta$ percent of maximal points lie on a chart interior, and these points correspond to the maximal points with the highest confidence in $\tilde l_{\theta_k}$. We refer $\eta$ as the \textit{minimum interior rate}. 
$c$ can 
then be estimated by the mean $\tilde l_{\theta_k}$ of the interior maximal points \cite{elkan2008_pu_learning,bekker2020_pu_learning_survey}. As $\tilde l_{\theta_k}$ is not explicitly calibrated and the SCAR assumption does not strictly hold in practice, we adopt the median estimator instead for improved robustness.

\bfPar{Point Cloud and Mesh Extraction.}
After the label frequency has been estimated, the \textit{reconstructed minimal neural atlas} $\{ (U_{\theta_k}, \varphi_{\theta_k}) \mid k \in \mathcal{K} \}$, and hence the \textit{reconstructed surface} $\mathcal{S} = \bigcup_{k \in \mathcal{K}} \varphi_k(U_{\theta_k})$, are then well-defined. 
As a result, we can extract the \textit{reconstructed surface point cloud} $\mathcal{X} = \bigcup_{k \in \mathcal{K}} \varphi_{\theta_k} (\mathcal{V}_k \cap U_{\theta_k})$. 
Furthermore, we can also extract a mesh from the reconstructed minimal neural atlas, similar to \cite{groueix2018_atlasnet}. We refer this as the \textit{reconstructed mesh}. This can be done by first defining a regular mesh in the open square of each chart and then discarding triangles with vertices outside of the reconstructed parametric domain. The mesh is then transferred to the reconstructed surface via the parameterization of each chart.

\par
Nevertheless, it is often useful to extract a point cloud or mesh with a specific number of vertices. We achieve this in an approximate but efficient manner by adopting a two-step batch \textit{rejection sampling} strategy. Firstly, we employ a small batch of UV samples to estimate the \textit{occupancy rate}, which quantifies the extent to which the open square of all charts are occupied. Given such an estimate, we then deduce the number of additional UV samples required to eventually yield a point cloud or mesh with approximately the target size.

\section{Experiments}
\label{sec:exp}

\par
We conduct two standard experiments: surface reconstruction (Sec.~\ref{subsec:exp:surfrec}) and single-view reconstruction (Sec.~\ref{subsec:exp:svr}) to verify that our representation can effectively learn a minimal atlas with distortion-minimal parameterization for surfaces of arbitrary topology. The first experiment considers the basic task of reconstructing the target surface given its point cloud, while the second is concerned with the complex task of surface reconstruction from a single image of the target. In addition to the benchmark experiments, we also perform ablation studies to investigate the significance of various components in our representation (Sec.~\ref{subsec:exp:ablation}).

\bfPar{Datasets.}
We perform all experiments on the widely used \textit{ShapeNet} dataset \cite{chang2015_shapenet}, which is a large-scale dataset of 3D models of common objects. Specifically, we adopt the dataset preprocessed by ONet \cite{mescheder2019_onet}. Instead of the default unit cube normalization on the point clouds, we follow existing atlas-based representations on a unit ball normalization. The ShapeNet dataset serves as a strong benchmark on representing general single-object closed surfaces.

\par
Additionally, we also perform the surface reconstruction experiment on the CLOTH3D++ \cite{madadi2021_cloth3d++} dataset, which contains approximately 13,000 3D models of garments across 6 categories. Following ONet, we preprocess the dataset by uniformly sampling 100,000 points on the mesh of each garment. The point clouds are then similarly normalized to a unit ball. With this dataset, we are able to evaluate the representation power on general single-object open surfaces.

\bfPar{Metrics.}
We adopt a consistent set of metrics to assess the performance of a surface representation on all experiments. To quantify the accuracy of surface reconstruction, we employ the standard bidirectional Chamfer Distance (CD) \cite{fan2017_psgn} as well as the \textit{F-score} at the default distance threshold of 1\% (F@1\%) \cite{knapitsch2017_fscore_original,tatarchenko2019_fscore_propose}, which has been shown to be a more representative metric than CD \cite{tatarchenko2019_fscore_propose}. Following prior works on atlas-based representations, we report these two metrics on the reconstructed surface point cloud, given by regularly sampling the parametric domain of each chart. 
Furthermore, we also report the metrics on the \textit{reconstructed mesh point cloud}, given by uniformly sampling the reconstructed mesh. This is similarly done in \cite{gupta2020_nmf}, as well as in implicit representation works. We 
refer to the first set of metrics as Point Cloud CD and F@1\%, while the second as Mesh CD and F@1\%. The reported reconstruction metrics are computed with a point cloud size of 25,000 for both the reconstruction and the target.

\par
As pointed out by \cite{park2019_deepsdf}, topological errors in the reconstructed surface are better accounted for when evaluating on the reconstructed mesh point cloud. This is due to the non-uniform distribution of the reconstructed surface point cloud, especially at regions of high distortion where sampled points are sparse. Nevertheless, we still report the point cloud metrics to assess the reconstruction accuracy in the related task or setting of point cloud reconstruction.

\par
We employ a set of metrics to quantify the distortion of the chart parameterizations. In particular, we use the SSDE at the optimal scale $L^*$ (Eq.~\ref{eq:ssde_Lstar}) to measure the metric distortion up to a common scale. We also quantify the area distortion up to a common scale using a distortion energy, which is derived in a similar manner from the equi-area distortion energy introduced in \cite{degener2003_equiarea}. Lastly, we measure the \textit{conformal} distortion, or distortion of local angles, using the MIPS energy \cite{hormann2000_mips}. The reported distortion metrics are computed with respect to the UV samples associated with the reconstructed surface point cloud. We 
offset the distortion metrics such that a value of zero implies no distortion.

\begin{table}[t!]
\centering
\ra{0.925}
\scriptsize
\caption{Surface Reconstruction on CLOTH3D++.}
\label{table:surfrec:cloth3d++}

\begin{tabular}{@{}clcclcclccc@{}}
\toprule
 & \multicolumn{1}{c}{} & \multicolumn{2}{c}{Point Cloud} & \phantom{..} & \multicolumn{2}{c}{Mesh} & \phantom{..} & \multicolumn{3}{c}{Distortion} \\ \cmidrule(lr){3-4} \cmidrule(lr){6-7} \cmidrule(lr){9-11} 
\multirow{-3}{*}{\begin{tabular}[c]{@{}c@{}}No. of\\ Charts\end{tabular}} & \multicolumn{1}{c}{\multirow{-3}{*}{\begin{tabular}[c]{@{}c@{}}Surface\\ Representation\end{tabular}}} & CD, $10^{-4} \downarrow$ & F$@1\% \uparrow$ &  & CD, $10^{-4} \downarrow$ & F$@1\% \uparrow$ &  & Metric $\downarrow$ & Conformal $\downarrow$ & Area $\downarrow$ \\ \midrule
 & AtlasNet & \textbf{4.074} & \textbf{88.56} &  & 18.99 & 82.40 &  & 15.54 & 3.933 & 0.8428 \\
 & AtlasNet++ & 4.296 & 87.88 &  & 4.937 & 86.20 &  & 13.22 & 3.368 & 0.6767 \\
 & DSP & 7.222 & 82.41 &  & 21.86 & 78.93 &  & \textbf{1.427} & {\ul 0.5746} & \textbf{0.1032} \\
 & TearingNet & 6.872 & 84.99 &  & 8.321 & 83.40 &  & 17.40 & 3.407 & 1.149 \\
 & \cellcolor[HTML]{EFEFEF}Ours w/o $\mathcal{L}_{dist}$ & \cellcolor[HTML]{EFEFEF}{\ul 4.206} & \cellcolor[HTML]{EFEFEF}{\ul 88.38} & \cellcolor[HTML]{EFEFEF} & \cellcolor[HTML]{EFEFEF}\textbf{4.373} & \cellcolor[HTML]{EFEFEF}\textbf{87.85} & \cellcolor[HTML]{EFEFEF} & \cellcolor[HTML]{EFEFEF}3.263 & \cellcolor[HTML]{EFEFEF}1.328 & \cellcolor[HTML]{EFEFEF}0.1516 \\
\multirow{-6}{*}{1} & \cellcolor[HTML]{EFEFEF}Ours & \cellcolor[HTML]{EFEFEF}4.296 & \cellcolor[HTML]{EFEFEF}88.00 & \cellcolor[HTML]{EFEFEF} & \cellcolor[HTML]{EFEFEF}{\ul 4.476} & \cellcolor[HTML]{EFEFEF}{\ul 87.36} & \cellcolor[HTML]{EFEFEF} & \cellcolor[HTML]{EFEFEF}{\ul 1.600} & \cellcolor[HTML]{EFEFEF}\textbf{0.5688} & \cellcolor[HTML]{EFEFEF}{\ul 0.1652} \\ \midrule
 & AtlasNet & 3.856 & 89.78 &  & 7.075 & 86.71 &  & 6.411 & 1.931 & 0.4031 \\
 & AtlasNet++ & 4.106 & 88.62 &  & 4.734 & 86.93 &  & 20.65 & 5.095 & 1.116 \\
 & DSP & 4.710 & 87.12 &  & 5.536 & 85.91 &  & \textbf{0.2160} & \textbf{0.0771} & \textbf{0.0283} \\
 & \cellcolor[HTML]{EFEFEF}Ours w/o $\mathcal{L}_{dist}$ & \cellcolor[HTML]{EFEFEF}\textbf{3.603} & \cellcolor[HTML]{EFEFEF}\textbf{90.78} & \cellcolor[HTML]{EFEFEF} & \cellcolor[HTML]{EFEFEF}\textbf{3.846} & \cellcolor[HTML]{EFEFEF}\textbf{89.91} & \cellcolor[HTML]{EFEFEF} & \cellcolor[HTML]{EFEFEF}3.654 & \cellcolor[HTML]{EFEFEF}1.328 & \cellcolor[HTML]{EFEFEF}0.2227 \\
\multirow{-5}{*}{2} & \cellcolor[HTML]{EFEFEF}Ours & \cellcolor[HTML]{EFEFEF}{\ul 3.775} & \cellcolor[HTML]{EFEFEF}{\ul 90.08} & \cellcolor[HTML]{EFEFEF} & \cellcolor[HTML]{EFEFEF}{\ul 3.982} & \cellcolor[HTML]{EFEFEF}{\ul 89.50} & \cellcolor[HTML]{EFEFEF} & \cellcolor[HTML]{EFEFEF}{\ul 0.9227} & \cellcolor[HTML]{EFEFEF}{\ul 0.3582} & \cellcolor[HTML]{EFEFEF}{\ul 0.0847} \\ \midrule
 & AtlasNet & 3.396 & 91.47 &  & 6.269 & 88.74 &  & 8.028 & 2.485 & 0.4144 \\
 & AtlasNet++ & 3.368 & 91.62 &  & 3.652 & 90.67 &  & 12.89 & 3.615 & 1.073 \\
 & DSP & \textbf{3.227} & \textbf{92.06} &  & \textbf{3.501} & \textbf{91.26} &  & \textbf{0.4284} & \textbf{0.1252} & \textbf{0.0439} \\
 & \cellcolor[HTML]{EFEFEF}Ours w/o $\mathcal{L}_{dist}$ & \cellcolor[HTML]{EFEFEF}3.300 & \cellcolor[HTML]{EFEFEF}91.87 & \cellcolor[HTML]{EFEFEF} & \cellcolor[HTML]{EFEFEF}3.684 & \cellcolor[HTML]{EFEFEF}90.72 & \cellcolor[HTML]{EFEFEF} & \cellcolor[HTML]{EFEFEF}4.603 & \cellcolor[HTML]{EFEFEF}1.654 & \cellcolor[HTML]{EFEFEF}0.3047 \\
\multirow{-5}{*}{25} & \cellcolor[HTML]{EFEFEF}Ours & \cellcolor[HTML]{EFEFEF}{\ul 3.299} & \cellcolor[HTML]{EFEFEF}{\ul 91.90} & \cellcolor[HTML]{EFEFEF} & \cellcolor[HTML]{EFEFEF}{\ul 3.554} & \cellcolor[HTML]{EFEFEF}{\ul 91.07} & \cellcolor[HTML]{EFEFEF} & \cellcolor[HTML]{EFEFEF}{\ul 0.5637} & \cellcolor[HTML]{EFEFEF}{\ul 0.1770} & \cellcolor[HTML]{EFEFEF}{\ul 0.0940} \\ \bottomrule
\end{tabular}
\end{table}

\begin{table}[t!]
\centering
\ra{0.925}
\scriptsize
\caption{Surface Reconstruction on ShapeNet.}
\label{table:surfrec:shapenet}

\begin{tabular}{@{}clcclcclccc@{}}
\toprule
 & \multicolumn{1}{c}{} & \multicolumn{2}{c}{Point Cloud} & \phantom{..} & \multicolumn{2}{c}{Mesh} & \phantom{..} & \multicolumn{3}{c}{Distortion} \\ \cmidrule(lr){3-4} \cmidrule(lr){6-7} \cmidrule(lr){9-11} 
\multirow{-3}{*}{\begin{tabular}[c]{@{}c@{}}No. of\\ Charts\end{tabular}} & \multicolumn{1}{c}{\multirow{-3}{*}{\begin{tabular}[c]{@{}c@{}}Surface\\ Representation\end{tabular}}} & CD, $10^{-4} \downarrow$ & F$@1\% \uparrow$ &  & CD, $10^{-4} \downarrow$ & F$@1\% \uparrow$ &  & Metric $\downarrow$ & Conformal $\downarrow$ & Area $\downarrow$ \\ \midrule
 & AtlasNet & 8.131 & 79.74 &  & 13.37 & 74.60 &  & 21.23 & 4.151 & 1.574 \\
 & AtlasNet++ & 8.467 & 78.46 &  & 10.82 & 75.69 &  & 30.40 & 5.687 & 2.017 \\
 & DSP & 14.22 & 70.03 &  & 16.29 & 68.58 &  & \textbf{0.4684} & \textbf{0.1618} & \textbf{0.0580} \\
 & TearingNet & 11.64 & 75.96 &  & 20.01 & 70.86 &  & 21.96 & 5.092 & 1.882 \\
 & \cellcolor[HTML]{F3F3F3}Ours w/o $\mathcal{L}_{dist}$ & \cellcolor[HTML]{F3F3F3}\textbf{6.684} & \cellcolor[HTML]{F3F3F3}\textbf{83.05} & \cellcolor[HTML]{EFEFEF} & \cellcolor[HTML]{F3F3F3}\textbf{7.133} & \cellcolor[HTML]{F3F3F3}\textbf{81.76} & \cellcolor[HTML]{EFEFEF} & \cellcolor[HTML]{F3F3F3}8.264 & \cellcolor[HTML]{F3F3F3}2.654 & \cellcolor[HTML]{F3F3F3}0.4130 \\
\multirow{-6}{*}{1} & \cellcolor[HTML]{F3F3F3}Ours & \cellcolor[HTML]{F3F3F3}{\ul 7.559} & \cellcolor[HTML]{F3F3F3}{\ul 80.45} & \cellcolor[HTML]{EFEFEF} & \cellcolor[HTML]{F3F3F3}{\ul 7.959} & \cellcolor[HTML]{F3F3F3}{\ul 79.39} & \cellcolor[HTML]{EFEFEF} & \cellcolor[HTML]{F3F3F3}{\ul 2.546} & \cellcolor[HTML]{F3F3F3}{\ul 0.8929} & \cellcolor[HTML]{F3F3F3}{\ul 0.2246} \\ \midrule
 & AtlasNet & 7.071 & 81.98 &  & 10.96 & 77.68 &  & 16.27 & 4.037 & 1.041 \\
 & AtlasNet++ & 7.516 & 80.59 &  & 9.280 & 78.17 &  & 32.39 & 6.252 & 2.626 \\
 & DSP & 10.79 & 76.39 &  & 11.98 & 74.85 &  & \textbf{0.4130} & \textbf{0.1571} & \textbf{0.0380} \\
 & \cellcolor[HTML]{EFEFEF}Ours w/o $\mathcal{L}_{dist}$ & \cellcolor[HTML]{EFEFEF}\textbf{6.266} & \cellcolor[HTML]{EFEFEF}\textbf{84.04} & \cellcolor[HTML]{EFEFEF} & \cellcolor[HTML]{EFEFEF}{\ul 6.875} & \cellcolor[HTML]{EFEFEF}{\ul 82.22} & \cellcolor[HTML]{EFEFEF} & \cellcolor[HTML]{EFEFEF}10.23 & \cellcolor[HTML]{EFEFEF}3.303 & \cellcolor[HTML]{EFEFEF}0.5155 \\
\multirow{-5}{*}{3} & \cellcolor[HTML]{EFEFEF}Ours & \cellcolor[HTML]{EFEFEF}{\ul 6.311} & \cellcolor[HTML]{EFEFEF}{\ul 83.63} & \cellcolor[HTML]{EFEFEF} & \cellcolor[HTML]{EFEFEF}\textbf{6.761} & \cellcolor[HTML]{EFEFEF}\textbf{82.23} & \cellcolor[HTML]{EFEFEF} & \cellcolor[HTML]{EFEFEF}{\ul 2.189} & \cellcolor[HTML]{EFEFEF}{\ul 0.7094} & \cellcolor[HTML]{EFEFEF}{\ul 0.2521} \\ \midrule
 & AtlasNet & 6.285 & 83.98 &  & 7.855 & 81.25 &  & 14.48 & 4.522 & 0.9469 \\
 & AtlasNet++ & 6.451 & 83.50 &  & 7.333 & 81.87 &  & 20.34 & 5.643 & 1.981 \\
 & DSP & 7.995 & 81.54 &  & 8.609 & 80.08 &  & \textbf{0.9477} & \textbf{0.2900} & \textbf{0.1040} \\
 & \cellcolor[HTML]{EFEFEF}Ours w/o $\mathcal{L}_{dist}$ & \cellcolor[HTML]{EFEFEF}{\ul 5.844} & \cellcolor[HTML]{EFEFEF}{\ul 85.11} & \cellcolor[HTML]{EFEFEF} & \cellcolor[HTML]{EFEFEF}\textbf{6.646} & \cellcolor[HTML]{EFEFEF}{\ul 83.52} & \cellcolor[HTML]{EFEFEF} & \cellcolor[HTML]{EFEFEF}7.639 & \cellcolor[HTML]{EFEFEF}2.595 & \cellcolor[HTML]{EFEFEF}0.4464 \\
\multirow{-5}{*}{25} & \cellcolor[HTML]{EFEFEF}Ours & \cellcolor[HTML]{EFEFEF}\textbf{5.780} & \cellcolor[HTML]{EFEFEF}\textbf{85.28} & \cellcolor[HTML]{EFEFEF} & \cellcolor[HTML]{EFEFEF}{\ul 6.726} & \cellcolor[HTML]{EFEFEF}\textbf{83.86} & \cellcolor[HTML]{EFEFEF} & \cellcolor[HTML]{EFEFEF}{\ul 1.178} & \cellcolor[HTML]{EFEFEF}{\ul 0.3760} & \cellcolor[HTML]{EFEFEF}{\ul 0.1576} \\ \bottomrule
\end{tabular}
\end{table}

\bfPar{Baselines.}
We benchmark minimal neural atlas against state-of-the-art explicit neural representations that can be learned given raw unoriented target point clouds for training. Specifically, we compare with AtlasNet \cite{groueix2018_atlasnet}, DSP \cite{bednarik2020_dsp} and TearingNet \cite{pang2021_tearingnet}. Furthermore, a variant of AtlasNet that is trained with the Mesh CD and SSDE at the optimal scale, 
in addition to the original Point Cloud CD loss, is also adopted as an additional baseline, which we refer to as AtlasNet++. It serves as a strong baseline since topological errors in the reconstructions are explicitly regularized with the Mesh CD, unlike other baselines.
The remaining losses help to minimize the excessive distortion caused by optimizing the Mesh CD, as similarly mentioned in \cite{gupta2020_nmf}.

\subsection{Surface Reconstruction}
\label{subsec:exp:surfrec}

\begin{figure}[t]
\centering 
		\begin{subfigure}{0.135\linewidth}
    		\centering
            \captionsetup{font=scriptsize,labelfont=scriptsize}
			\caption*{AtlasNet}
			\includegraphics[width=1\linewidth,trim={1cm 1cm 1cm 1cm},clip]{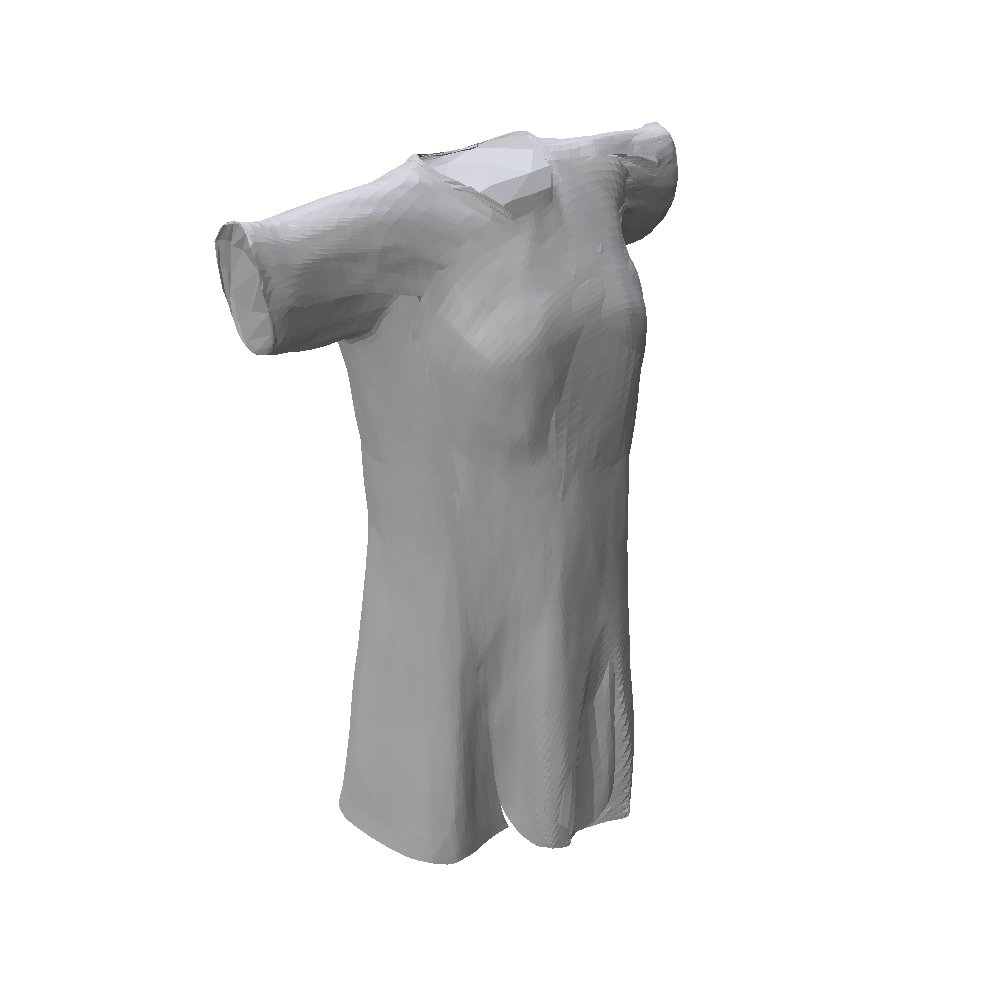}
            \includegraphics[width=1\linewidth,trim={1cm 5cm 1cm 1cm},clip]{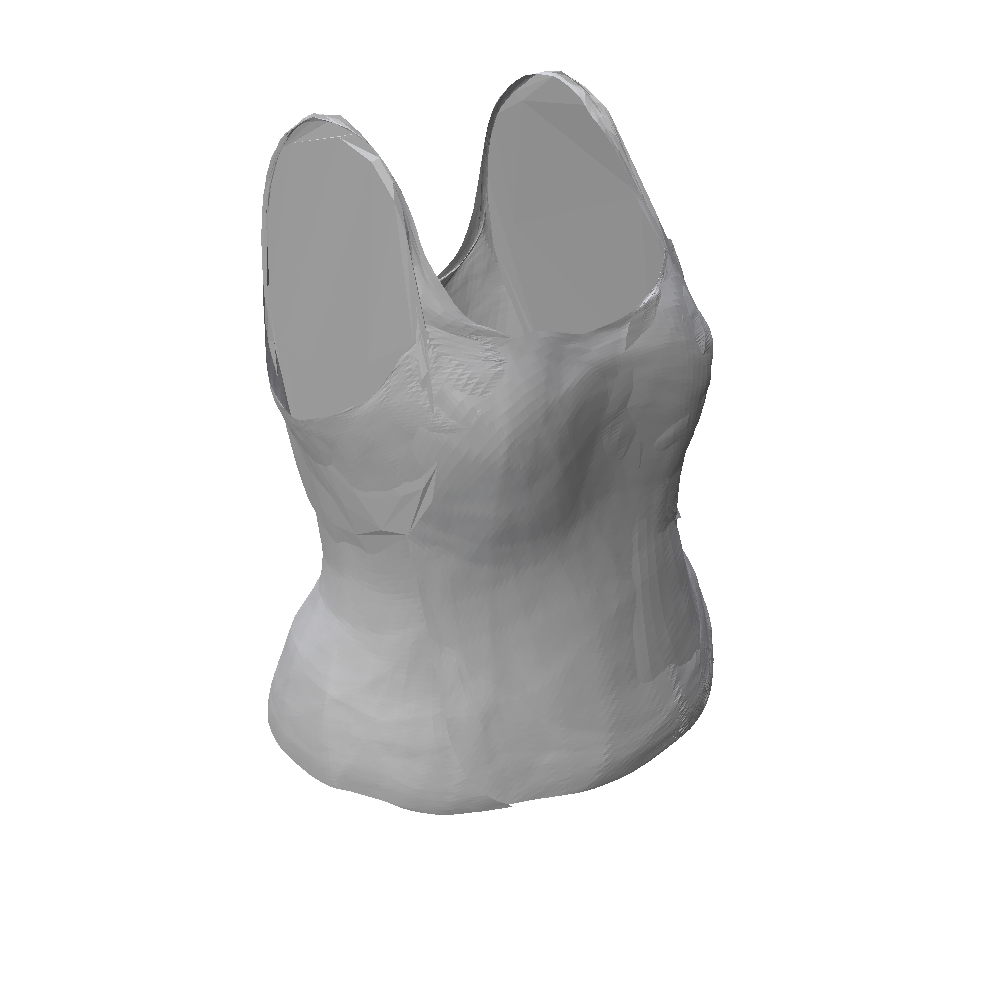}
            \includegraphics[width=1\linewidth,trim={5cm 3cm 5cm 5cm},clip]{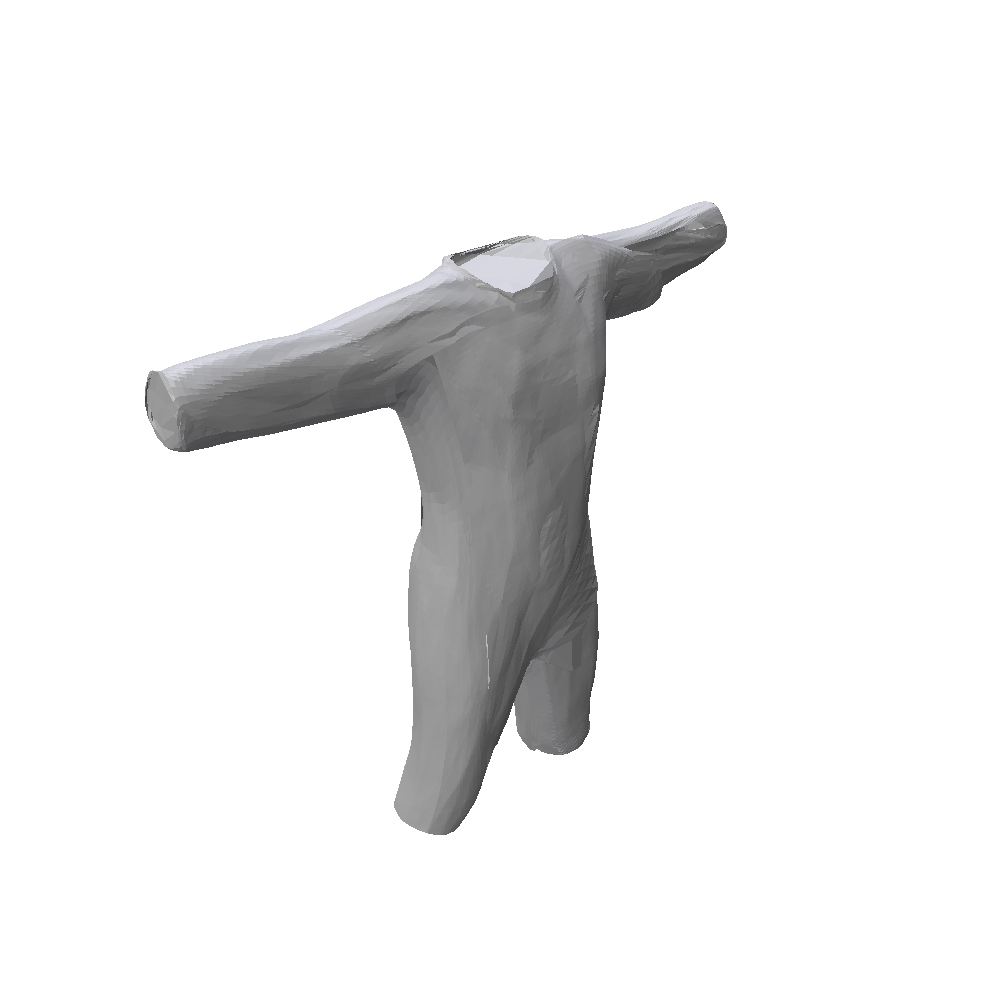}
            
			\includegraphics[width=1\linewidth,trim={3cm 2cm 1cm 1cm},clip]{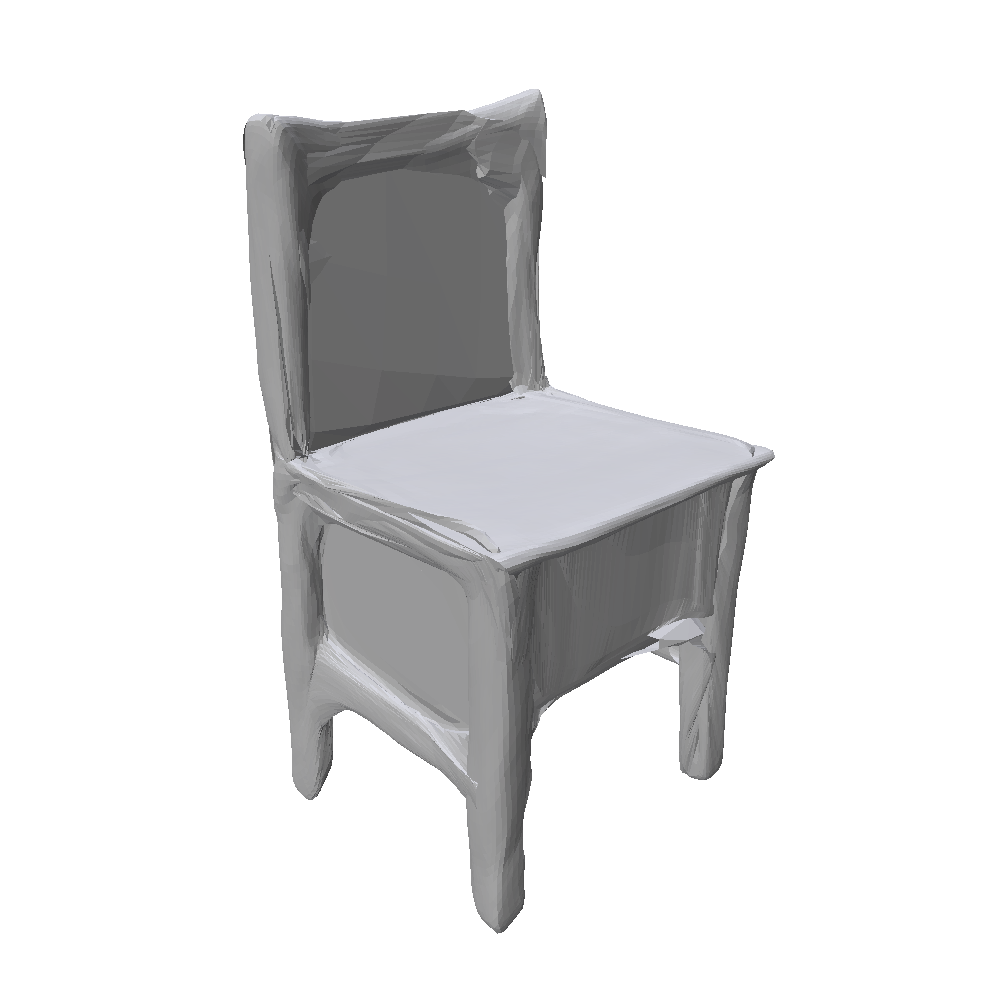}
            \includegraphics[width=1\linewidth,trim={1cm 3cm 2cm 4cm},clip]{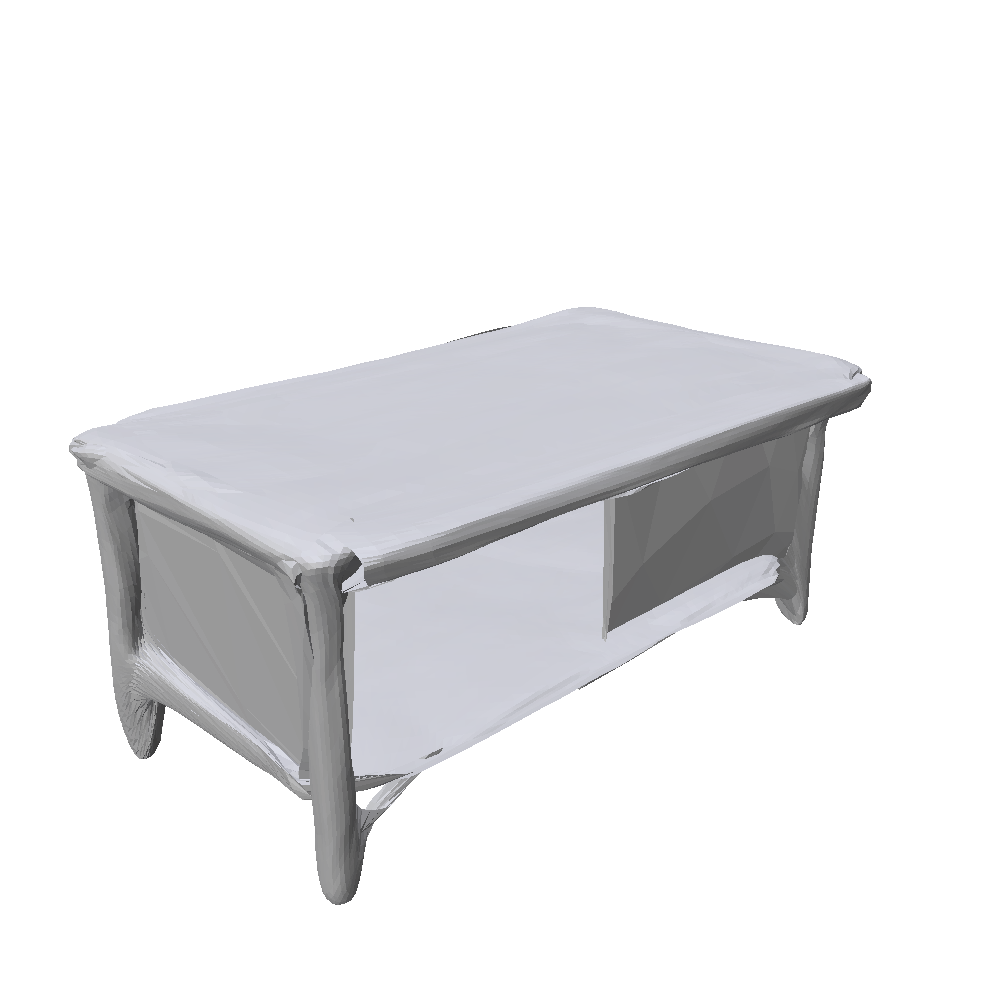}
            \includegraphics[width=1\linewidth,trim={1cm 3cm 2cm 4cm},clip]{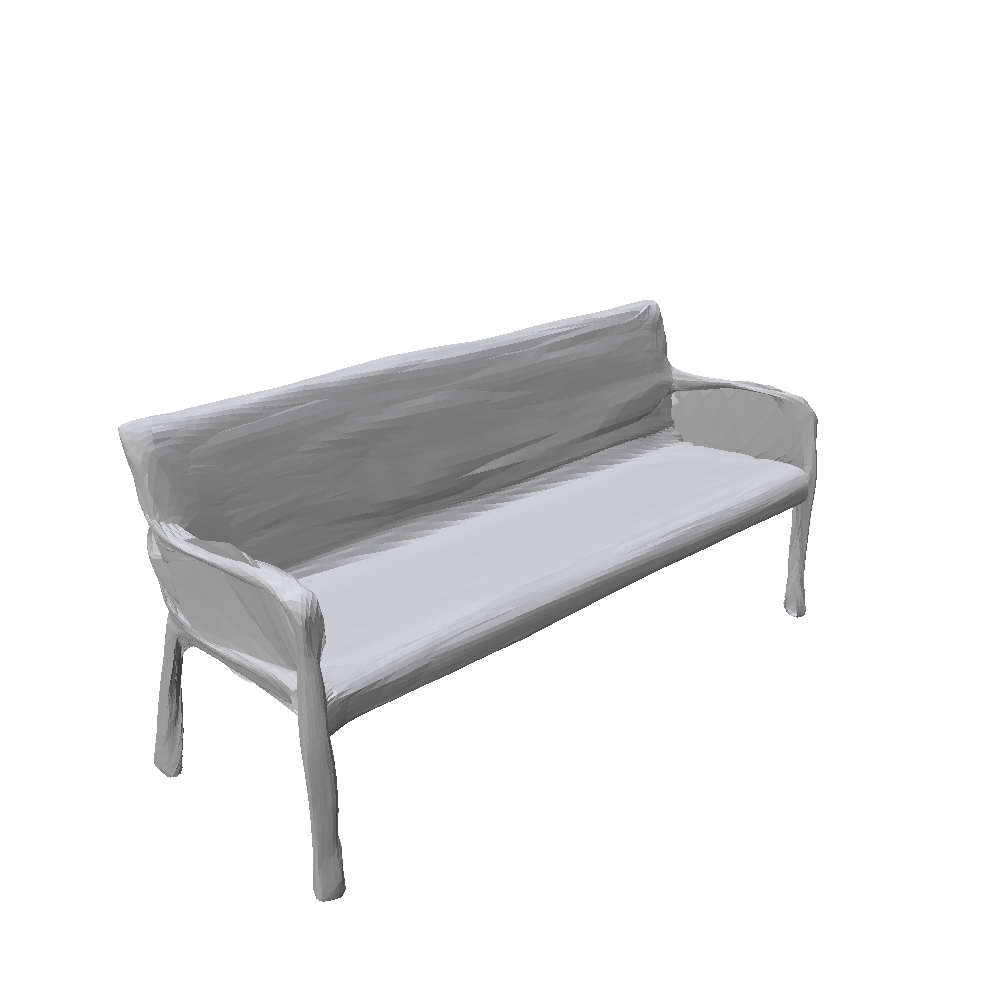}
		\end{subfigure}
        \begin{subfigure}{0.135\linewidth}
    		\centering
            \captionsetup{font=scriptsize,labelfont=scriptsize}
			\caption*{AtlasNet++}
			\includegraphics[width=1\linewidth,trim={1cm 1cm 1cm 1cm},clip]{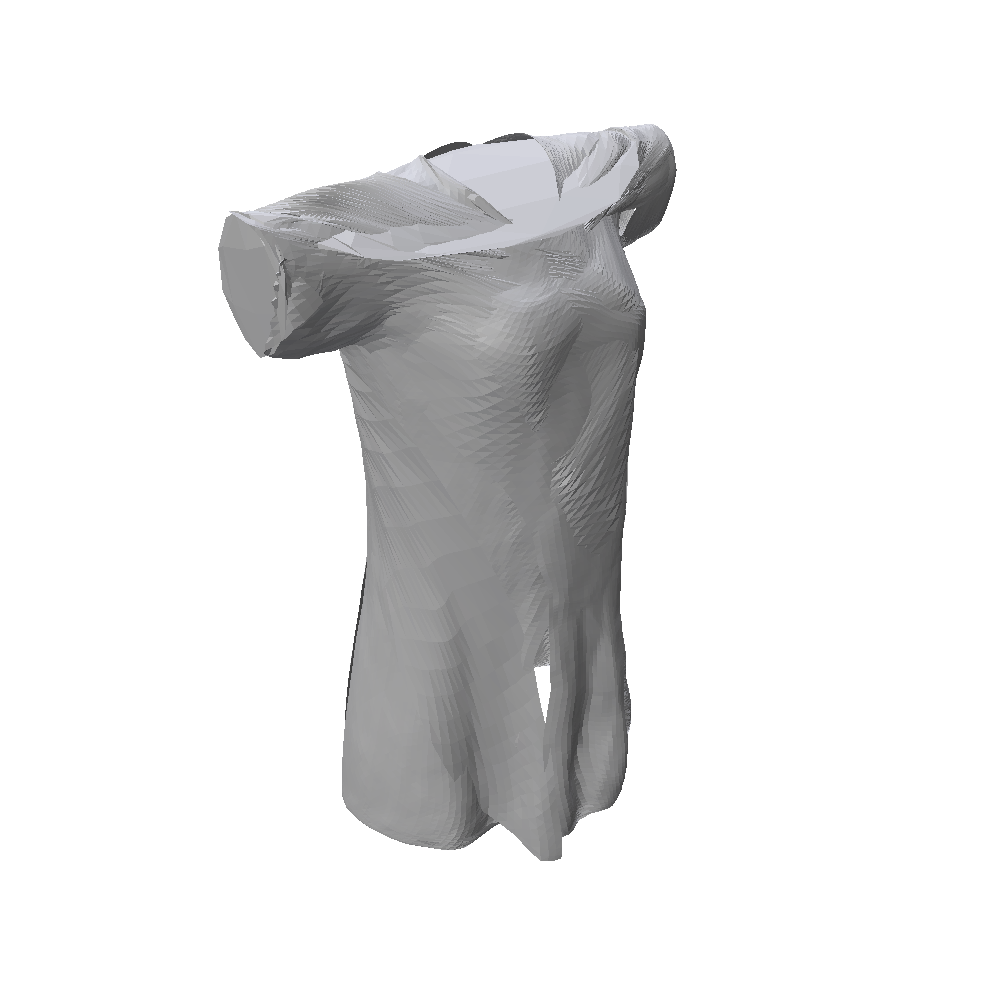}
            \includegraphics[width=1\linewidth,trim={1cm 5cm 1cm 1cm},clip]{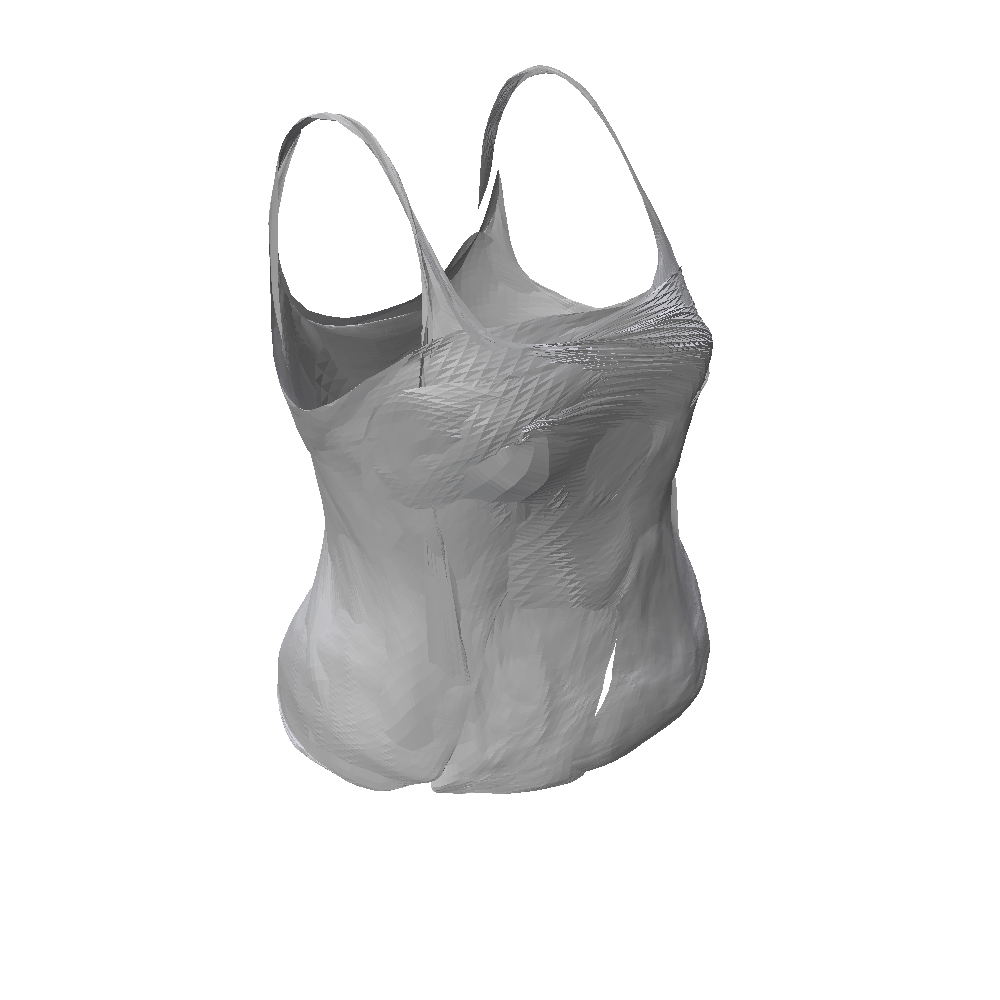}
            \includegraphics[width=1\linewidth,trim={5cm 3cm 5cm 5cm},clip]{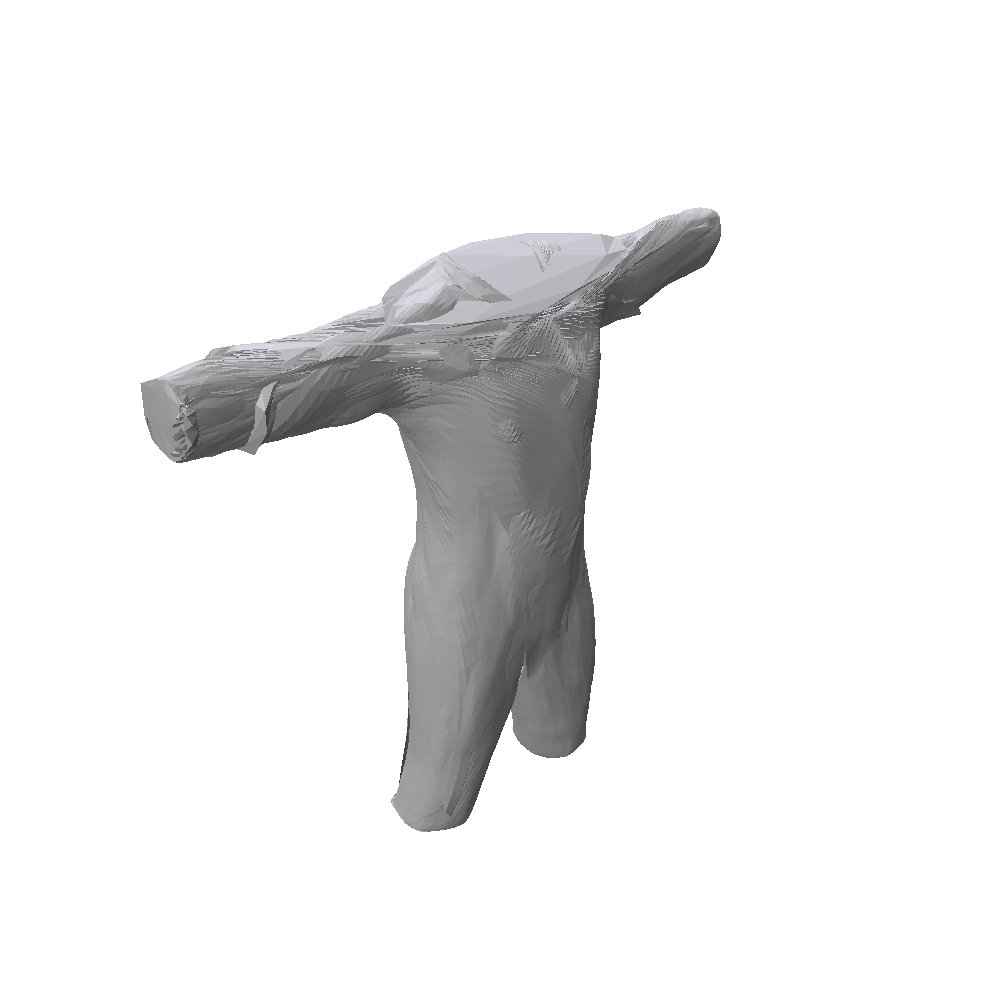}
            
			\includegraphics[width=1\linewidth,trim={3cm 2cm 1cm 1cm},clip]{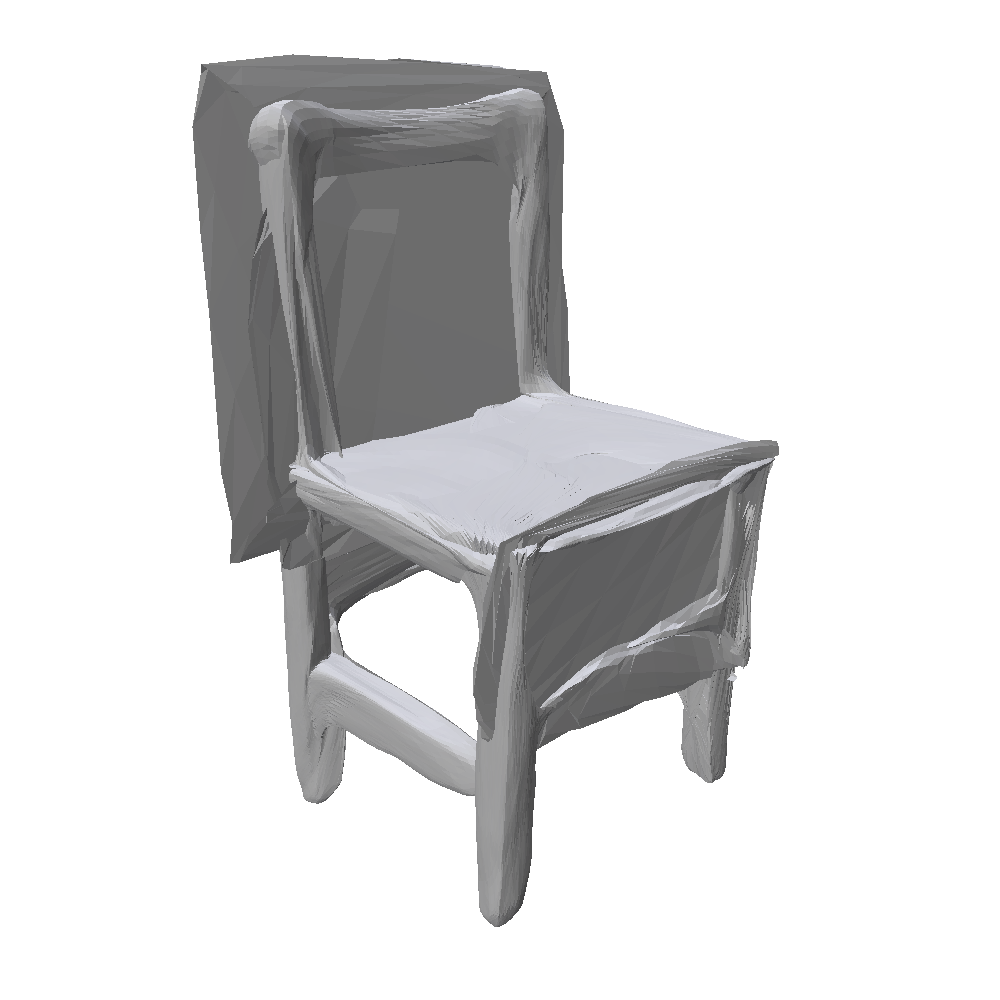}
            \includegraphics[width=1\linewidth,trim={1cm 3cm 2cm 4cm},clip]{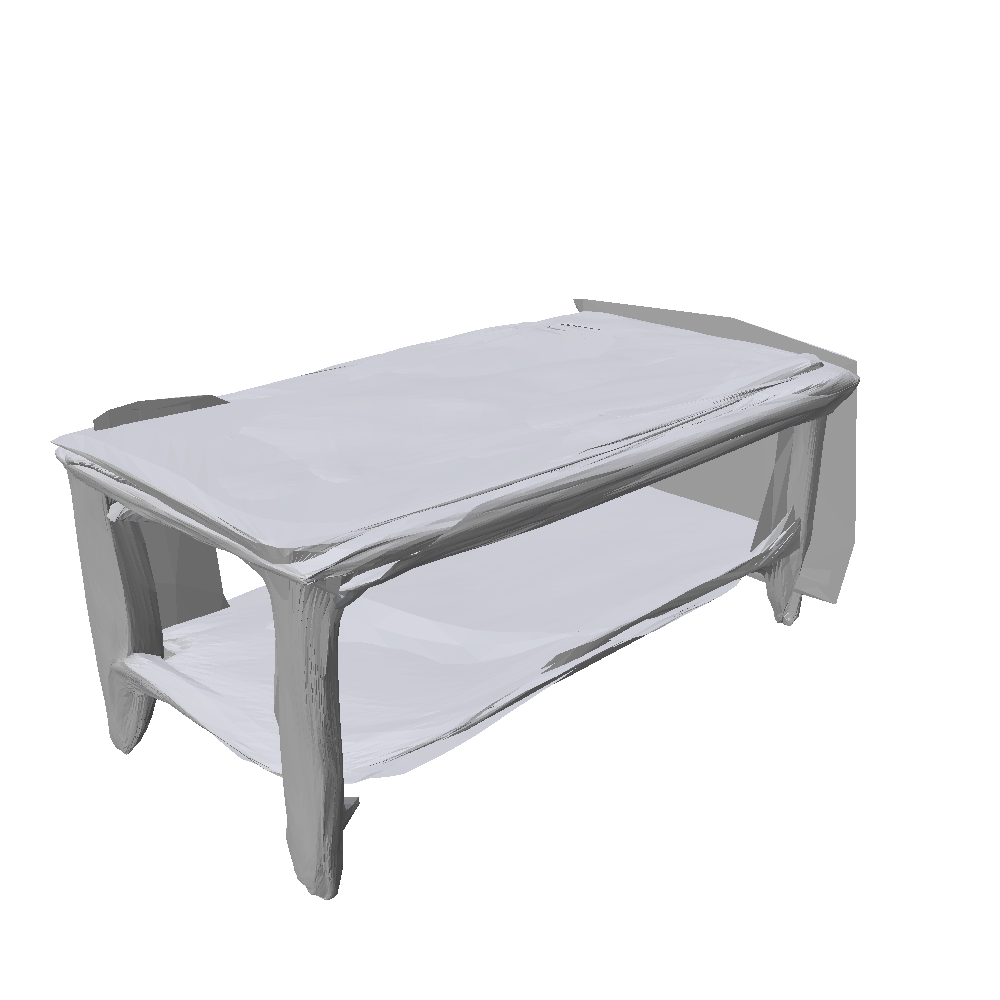}
            \includegraphics[width=1\linewidth,trim={1cm 3cm 2cm 4cm},clip]{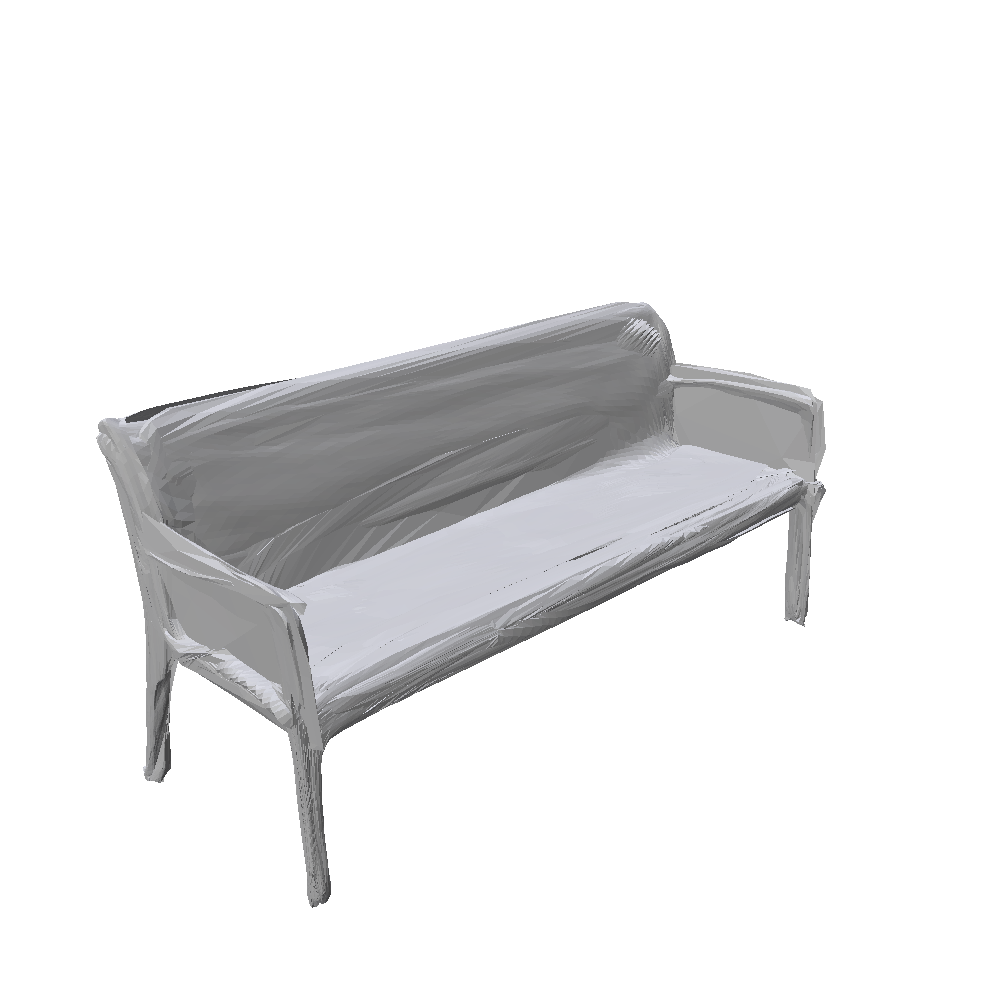}
		\end{subfigure}
        \begin{subfigure}{0.135\linewidth}
    		\centering
            \captionsetup{font=scriptsize,labelfont=scriptsize}
			\caption*{DSP}
			\includegraphics[width=1\linewidth,trim={1cm 1cm 1cm 1cm},clip]{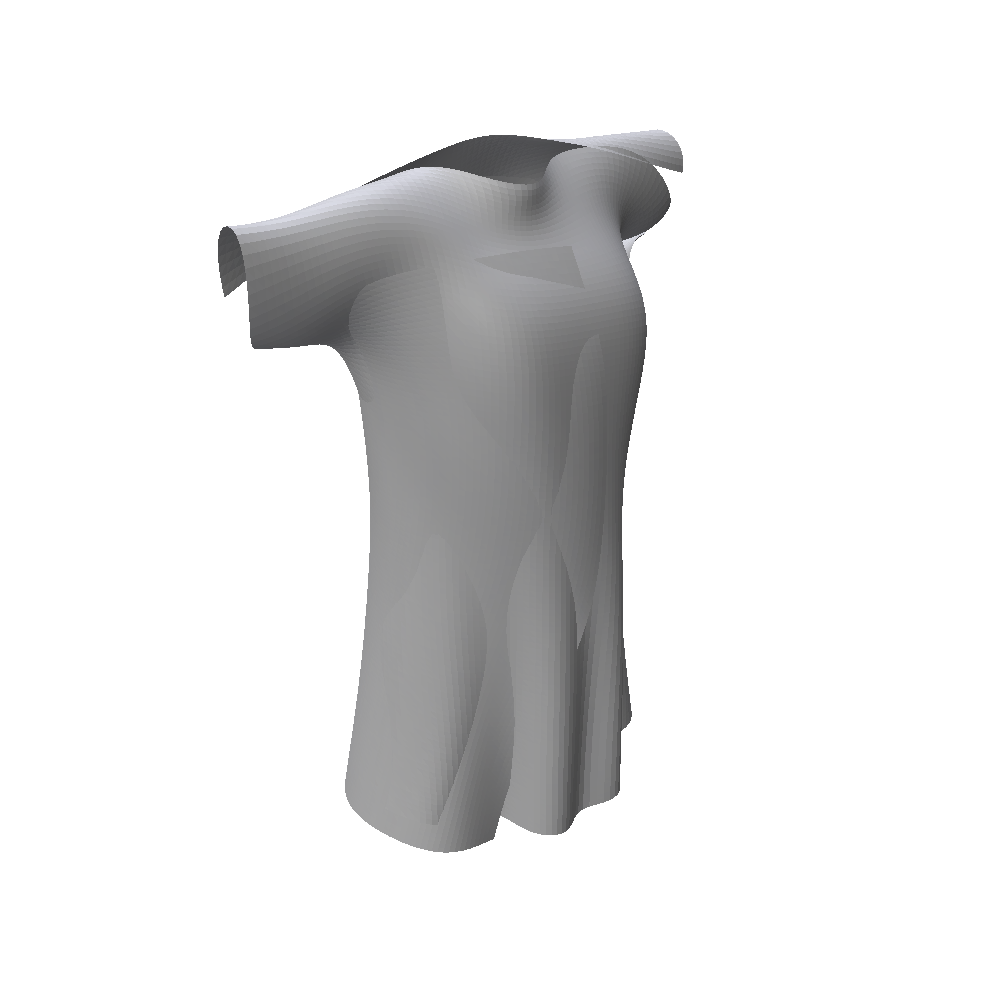}
            \includegraphics[width=1\linewidth,trim={1cm 5cm 1cm 1cm},clip]{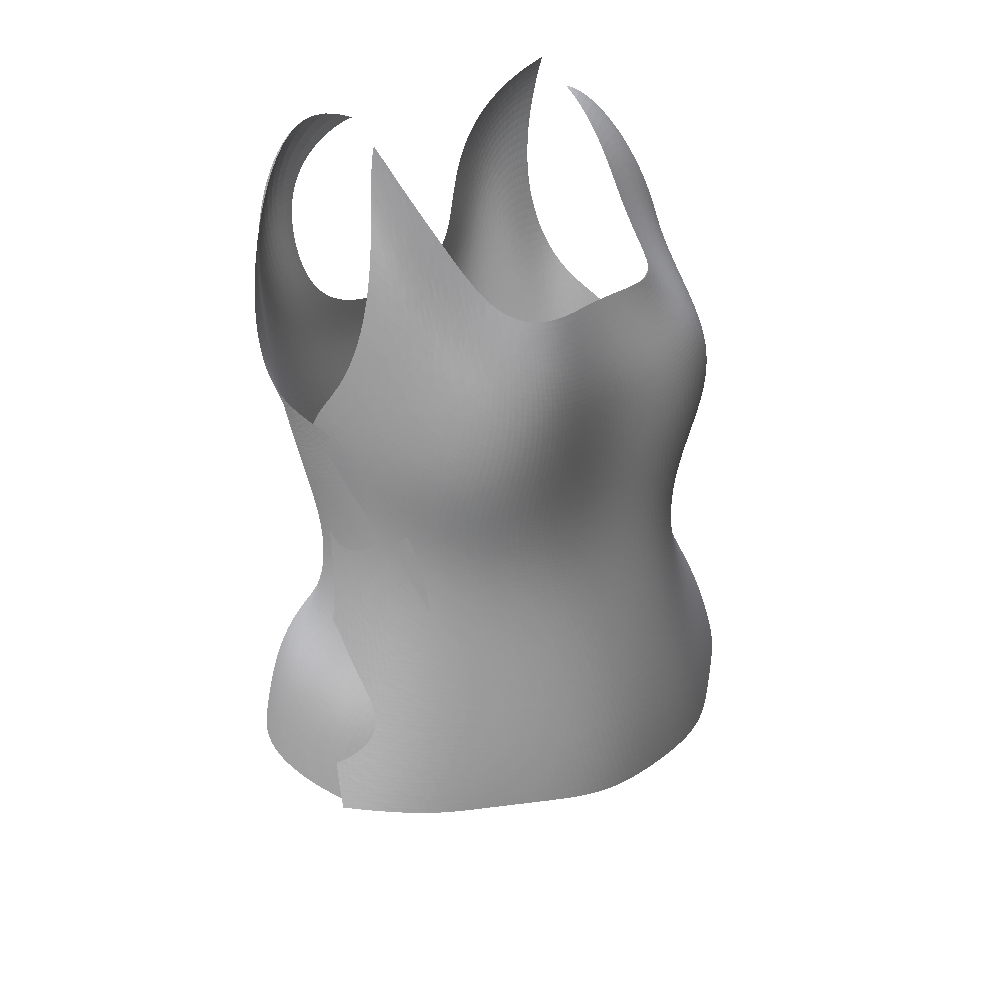}
            \includegraphics[width=1\linewidth,trim={5cm 3cm 5cm 5cm},clip]{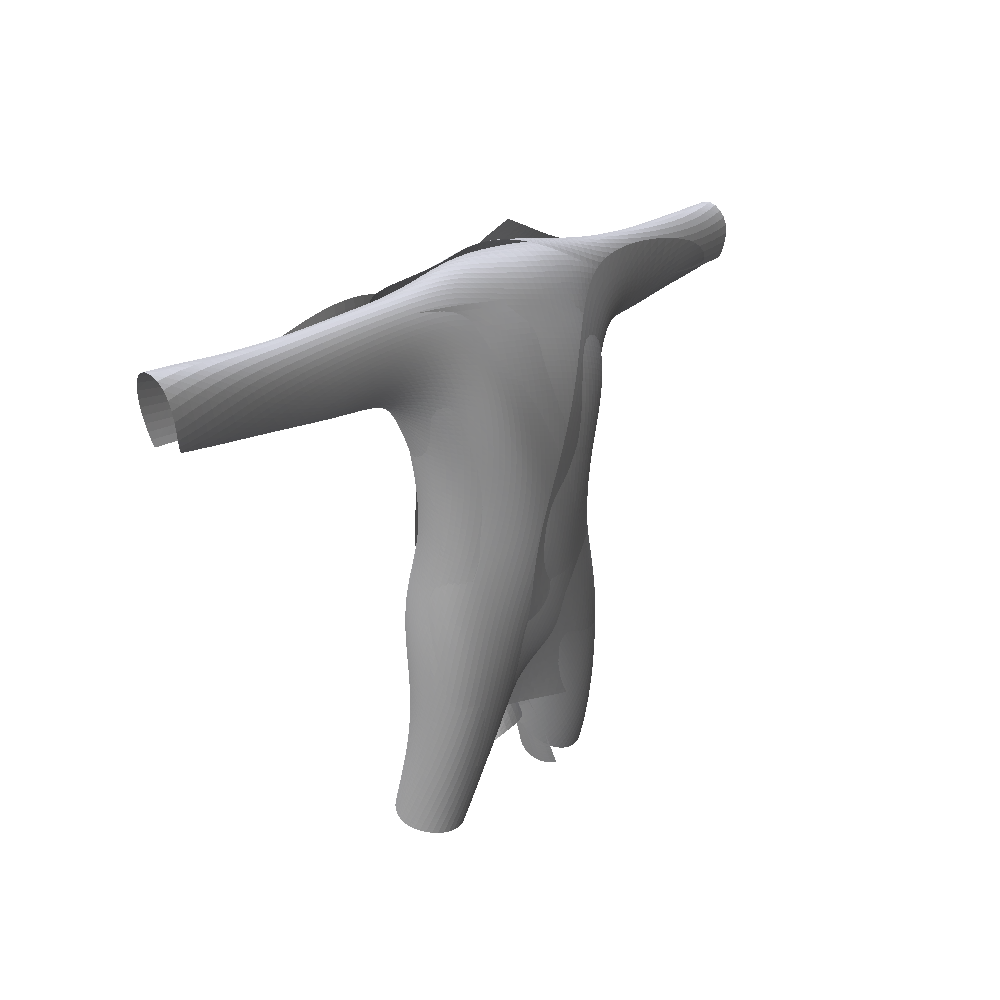}
            
			\includegraphics[width=1\linewidth,trim={3cm 2cm 1cm 1cm},clip]{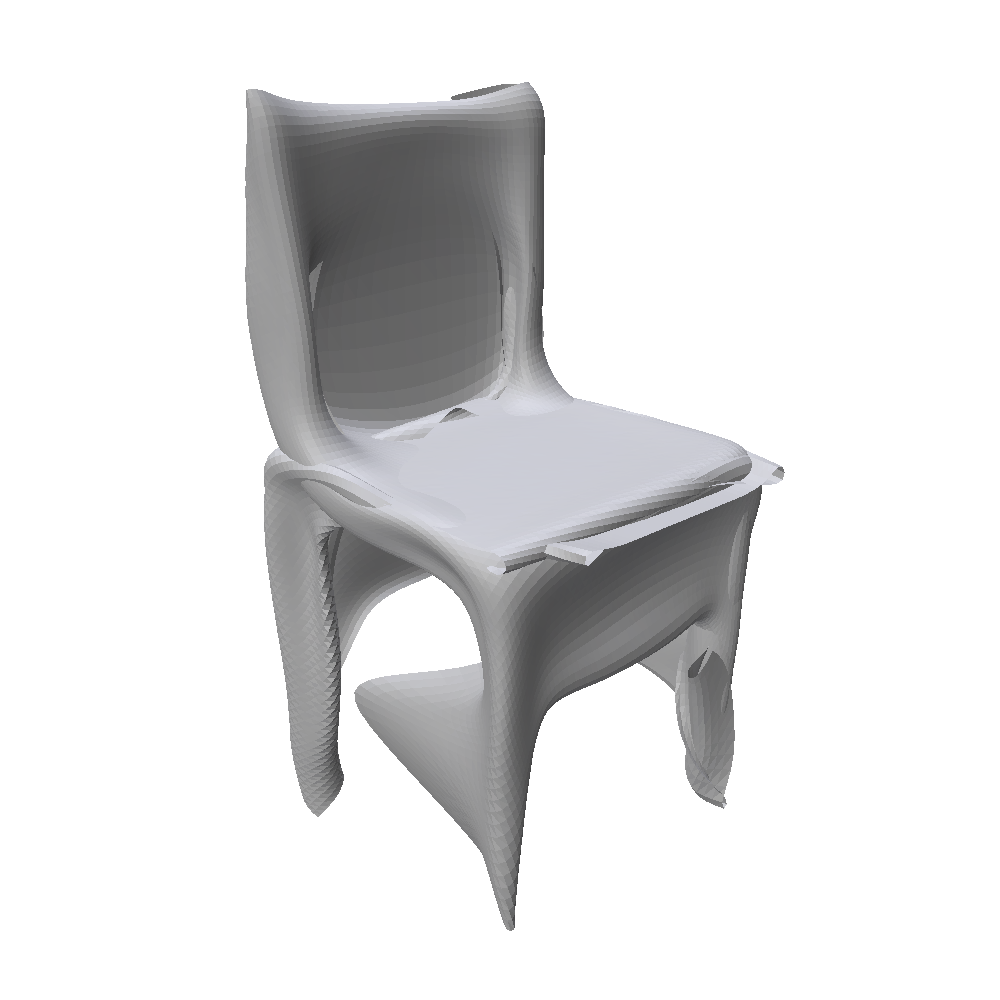}
            \includegraphics[width=1\linewidth,trim={1cm 3cm 2cm 4cm},clip]{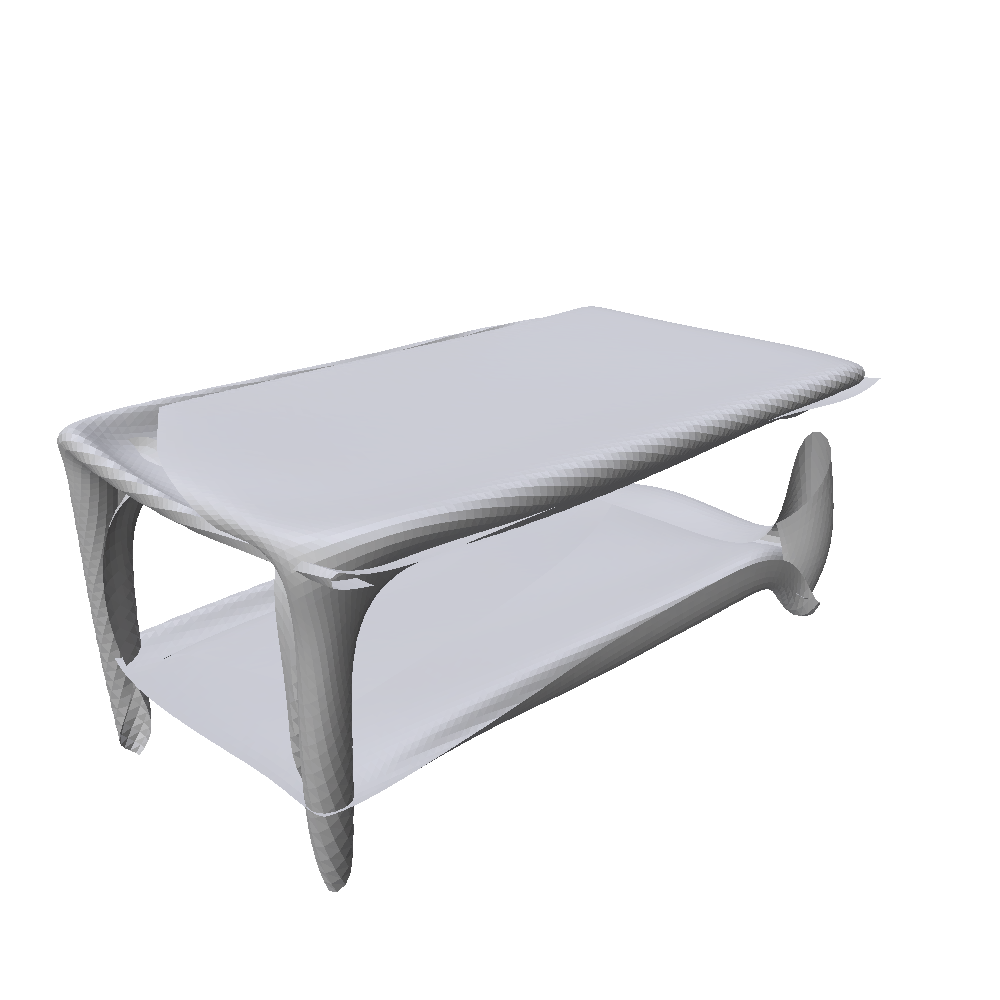}
            \includegraphics[width=1\linewidth,trim={1cm 3cm 2cm 4cm},clip]{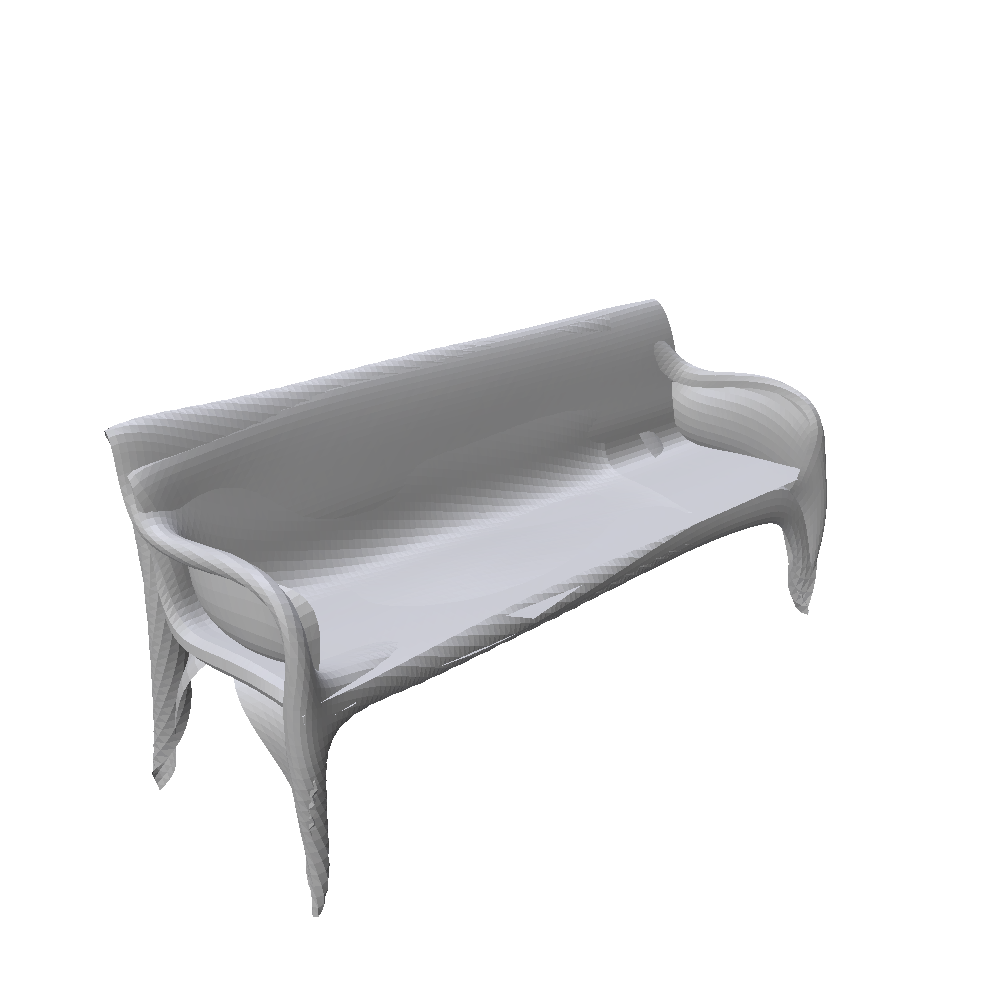}
		\end{subfigure}
        \begin{subfigure}{0.135\linewidth}
    		\centering
            \captionsetup{font=scriptsize,labelfont=scriptsize}
			\caption*{TearingNet}
			\includegraphics[width=1\linewidth,trim={1cm 1cm 1cm 1cm},clip]{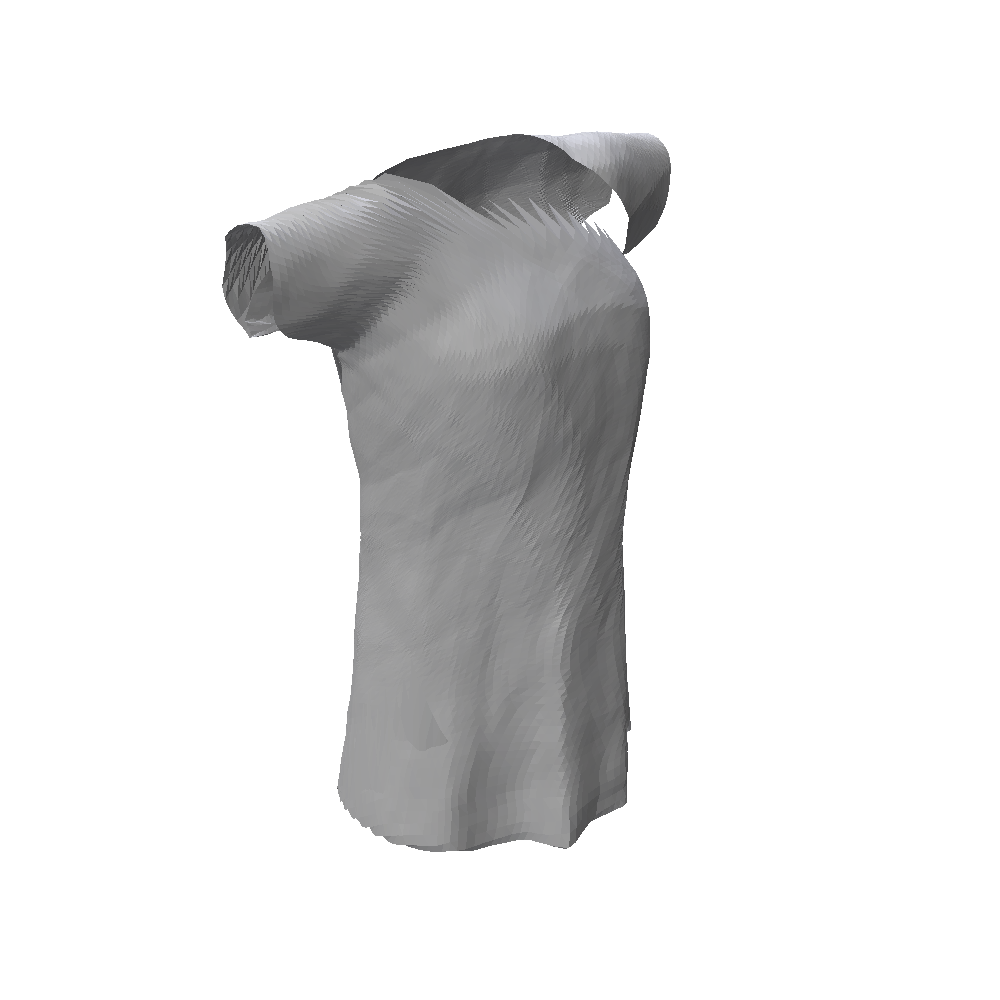}
            \includegraphics[width=1\linewidth,trim={1cm 5cm 1cm 1cm},clip]{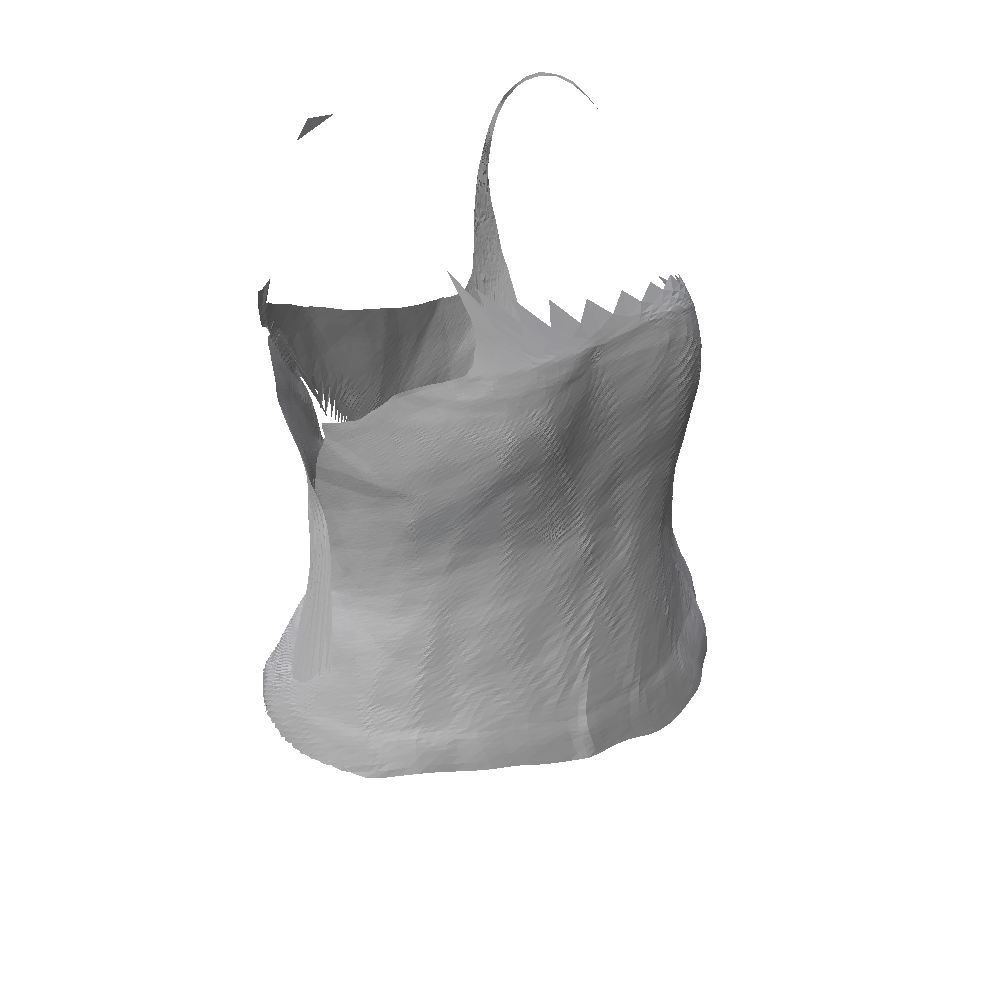}
            \includegraphics[width=1\linewidth,trim={5cm 3cm 5cm 5cm},clip]{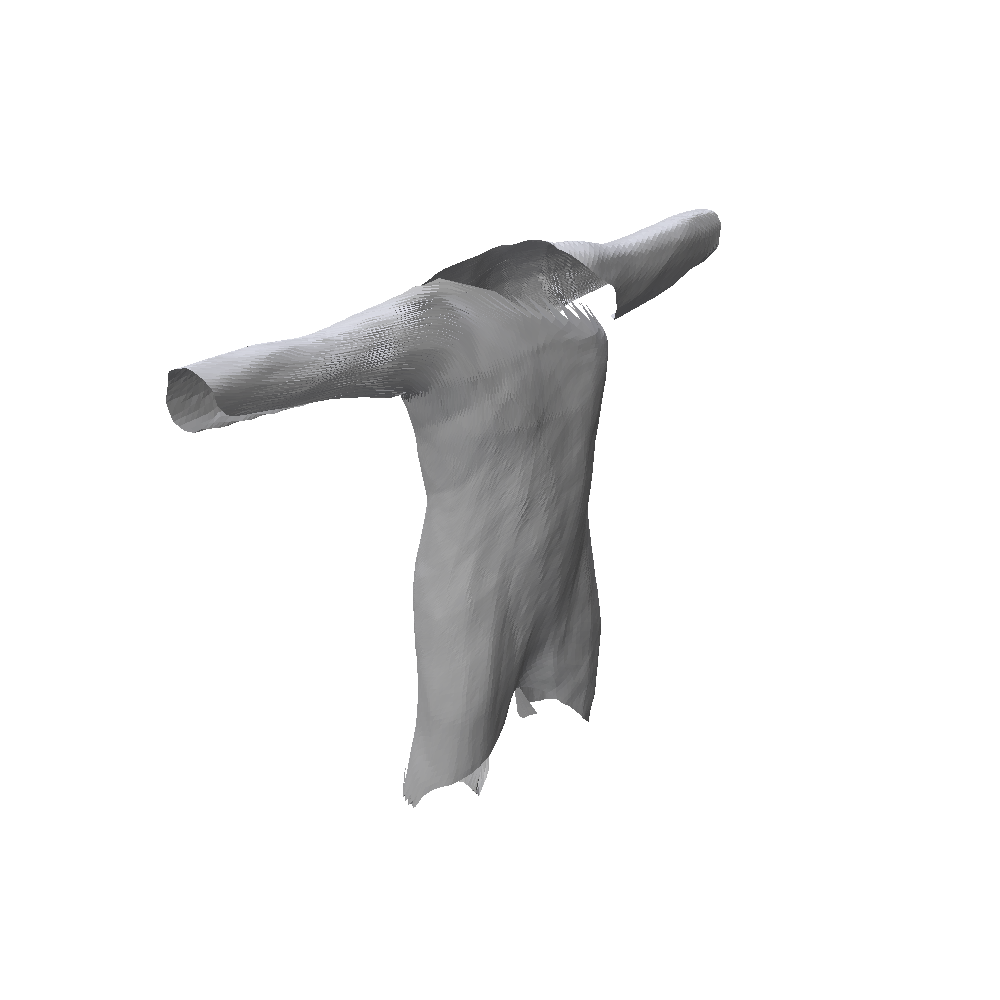}
            
			\includegraphics[width=1\linewidth,trim={3cm 2cm 1cm 1cm},clip]{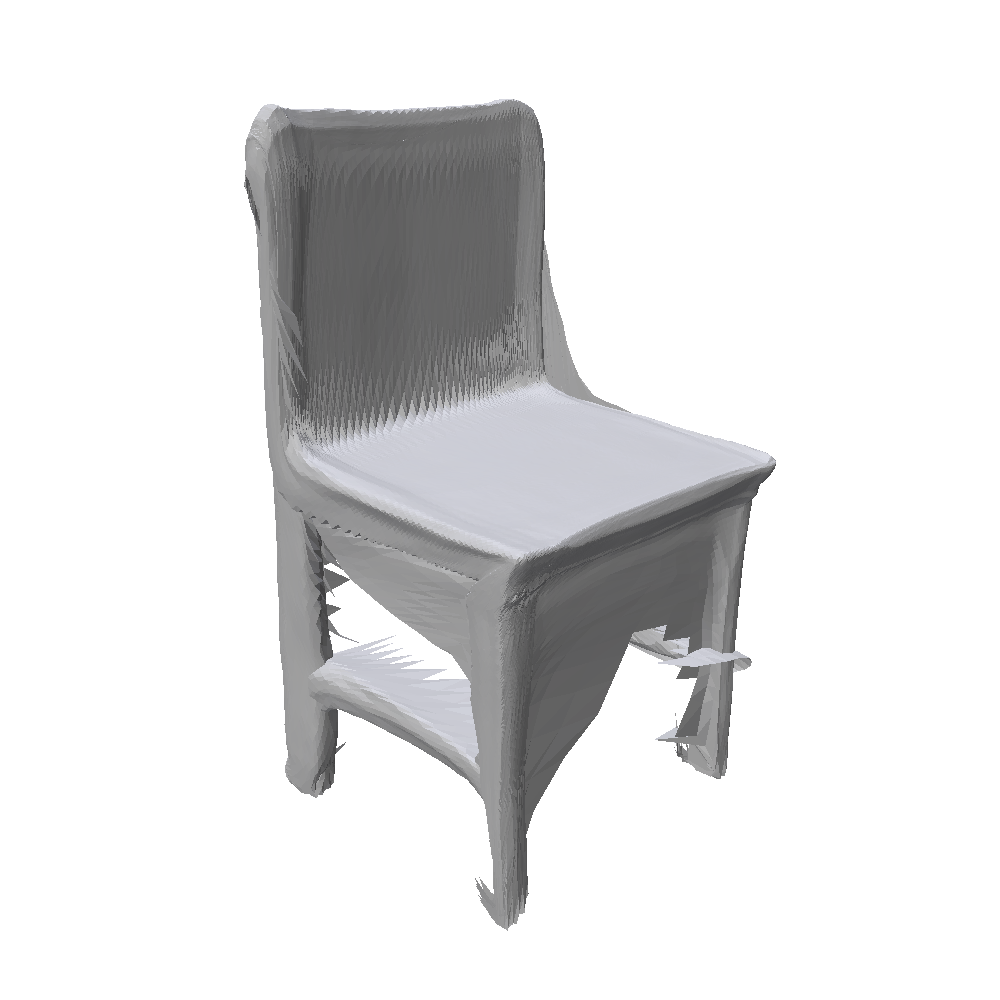}
            \includegraphics[width=1\linewidth,trim={1cm 3cm 2cm 4cm},clip]{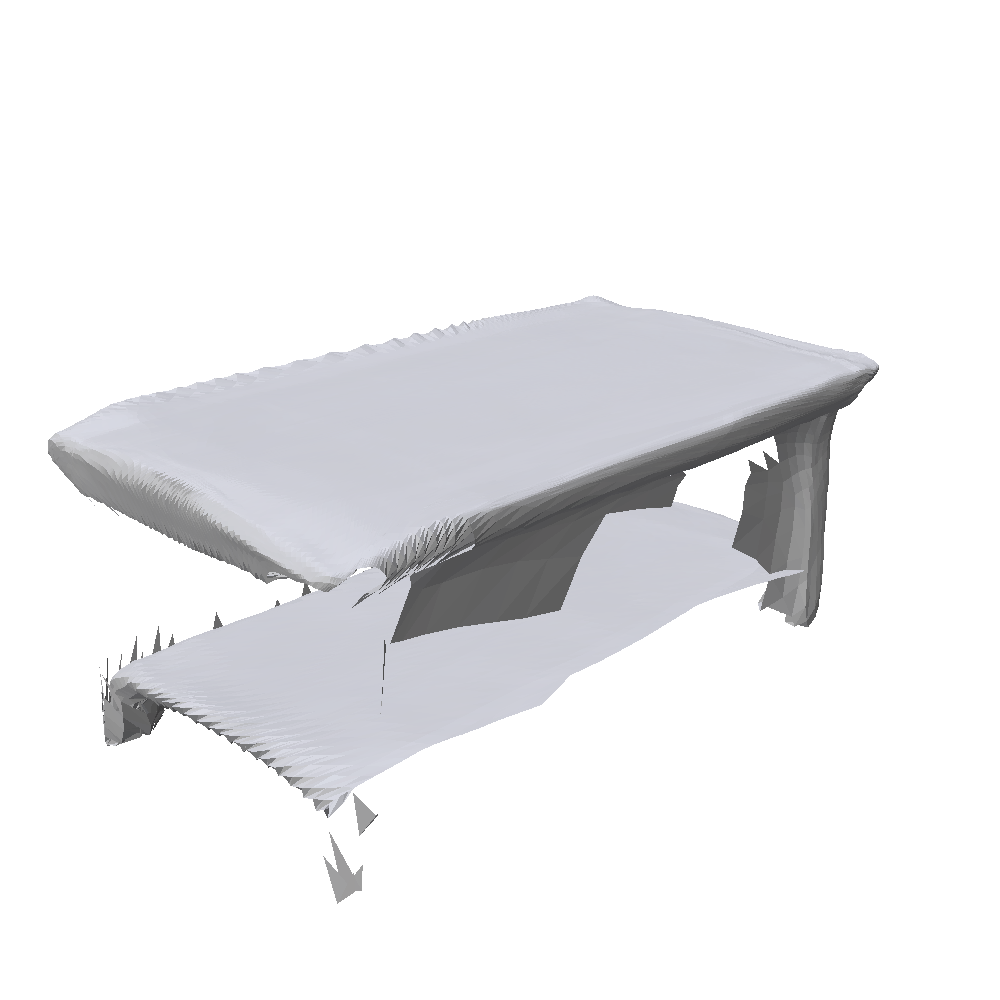}
            \includegraphics[width=1\linewidth,trim={1cm 3cm 2cm 4cm},clip]{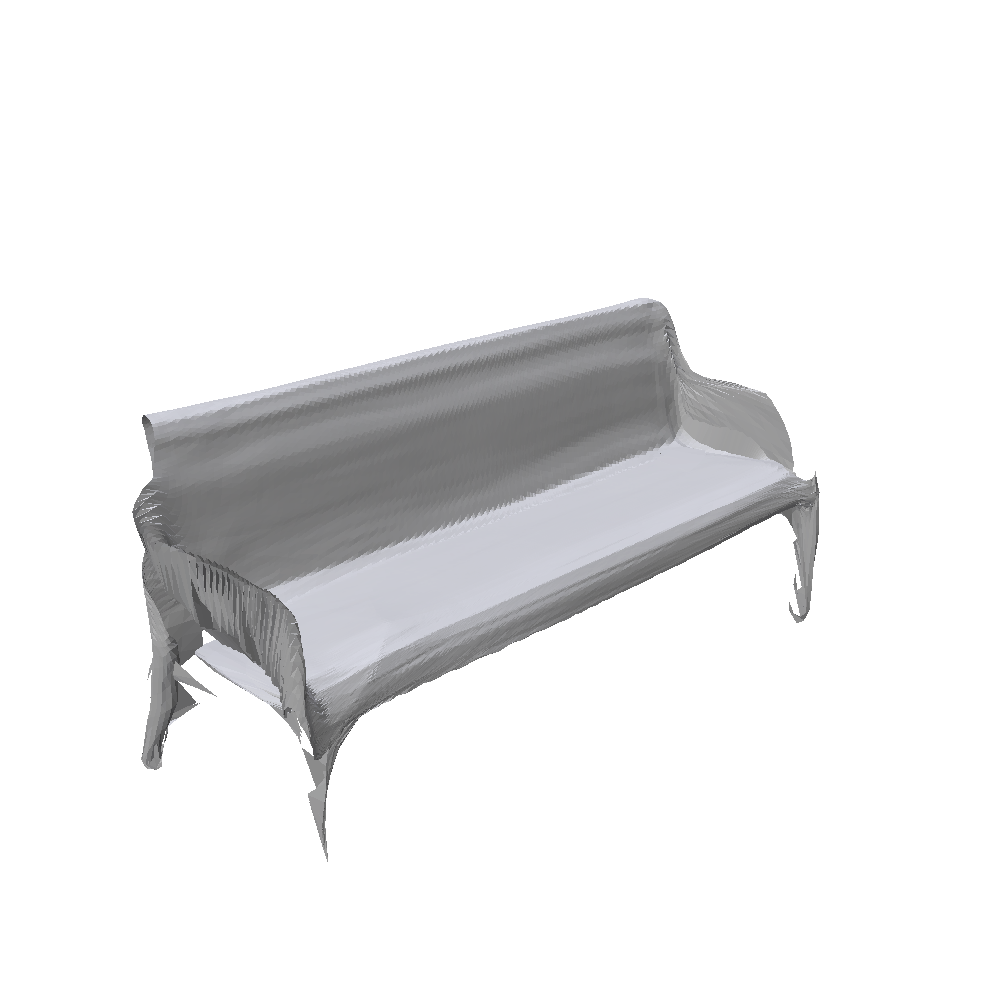}
		\end{subfigure}
        \begin{subfigure}{0.135\linewidth}
    		\centering
            \captionsetup{font=scriptsize,labelfont=scriptsize}
			\caption*{Ours}
			\includegraphics[width=1\linewidth,trim={1cm 1cm 1cm 1cm},clip]{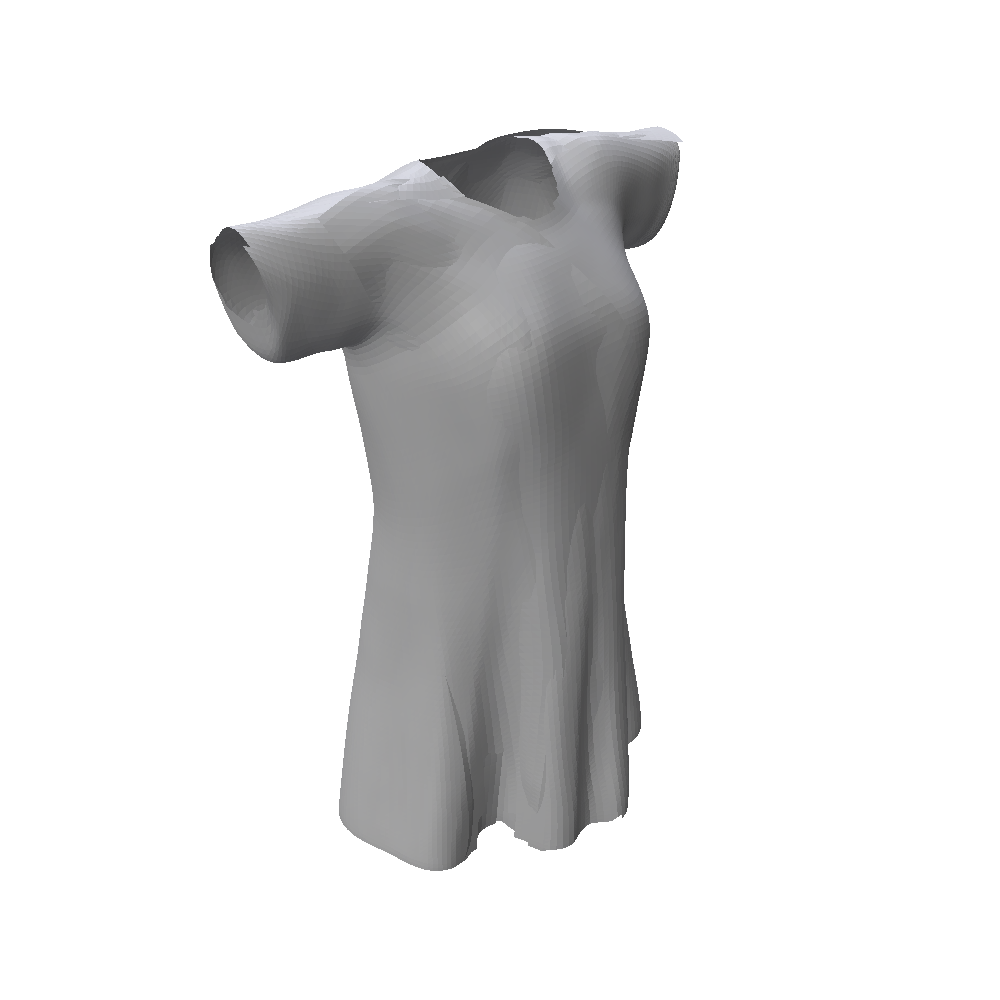}
            \includegraphics[width=1\linewidth,trim={1cm 5cm 1cm 1cm},clip]{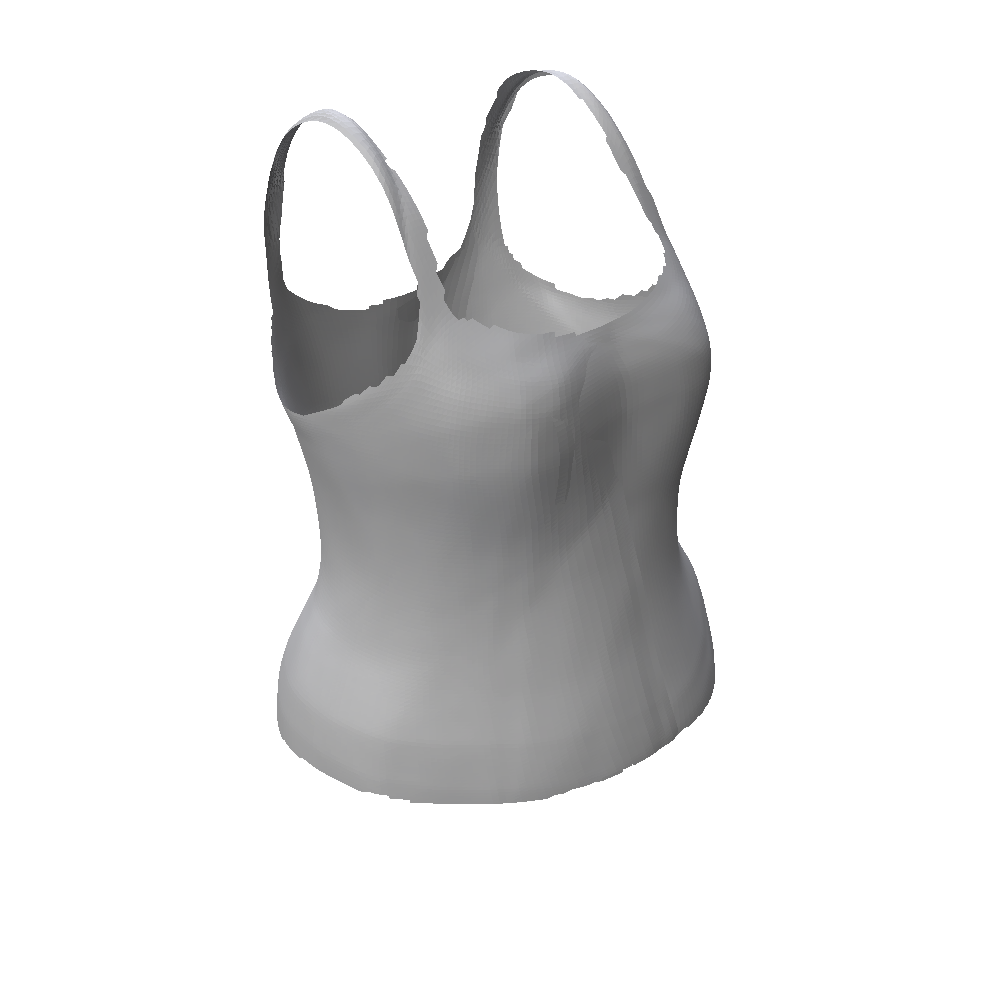}
            \includegraphics[width=1\linewidth,trim={5cm 3cm 5cm 5cm},clip]{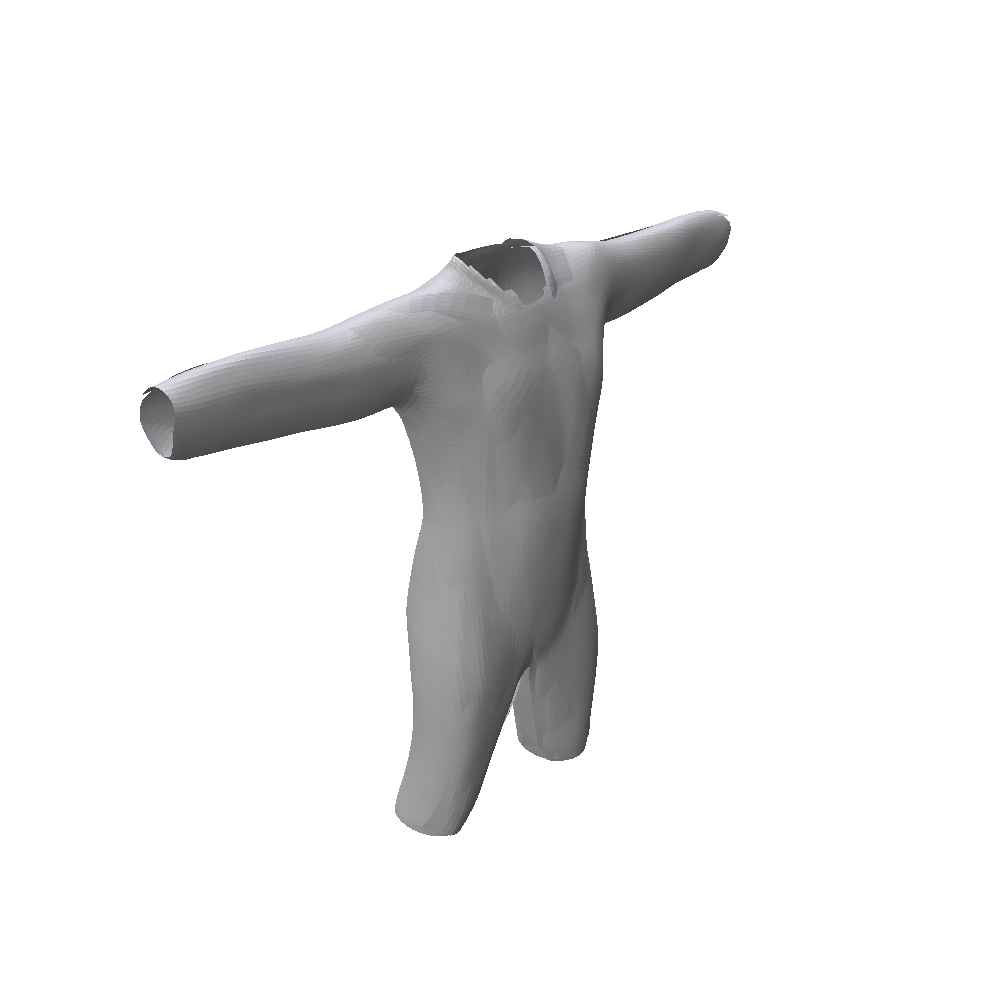}
            
			\includegraphics[width=1\linewidth,trim={3cm 2cm 1cm 1cm},clip]{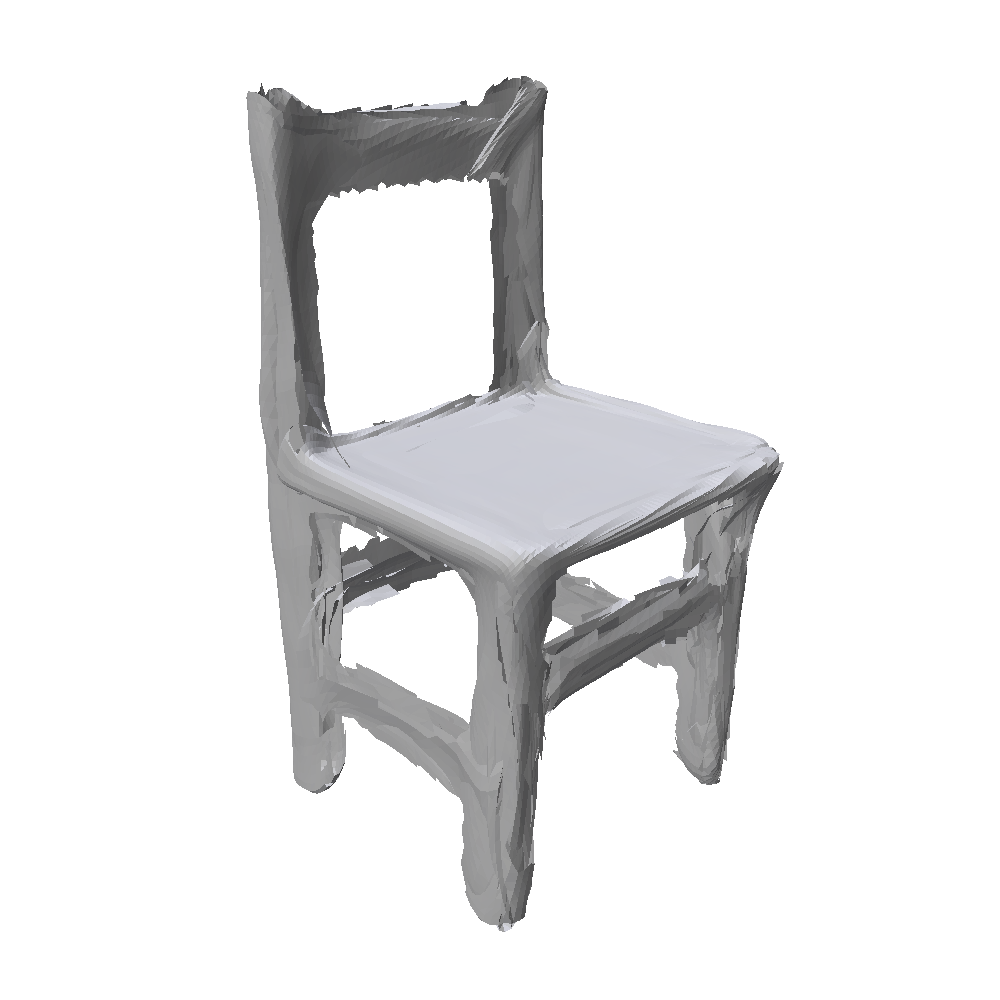}
            \includegraphics[width=1\linewidth,trim={1cm 3cm 2cm 4cm},clip]{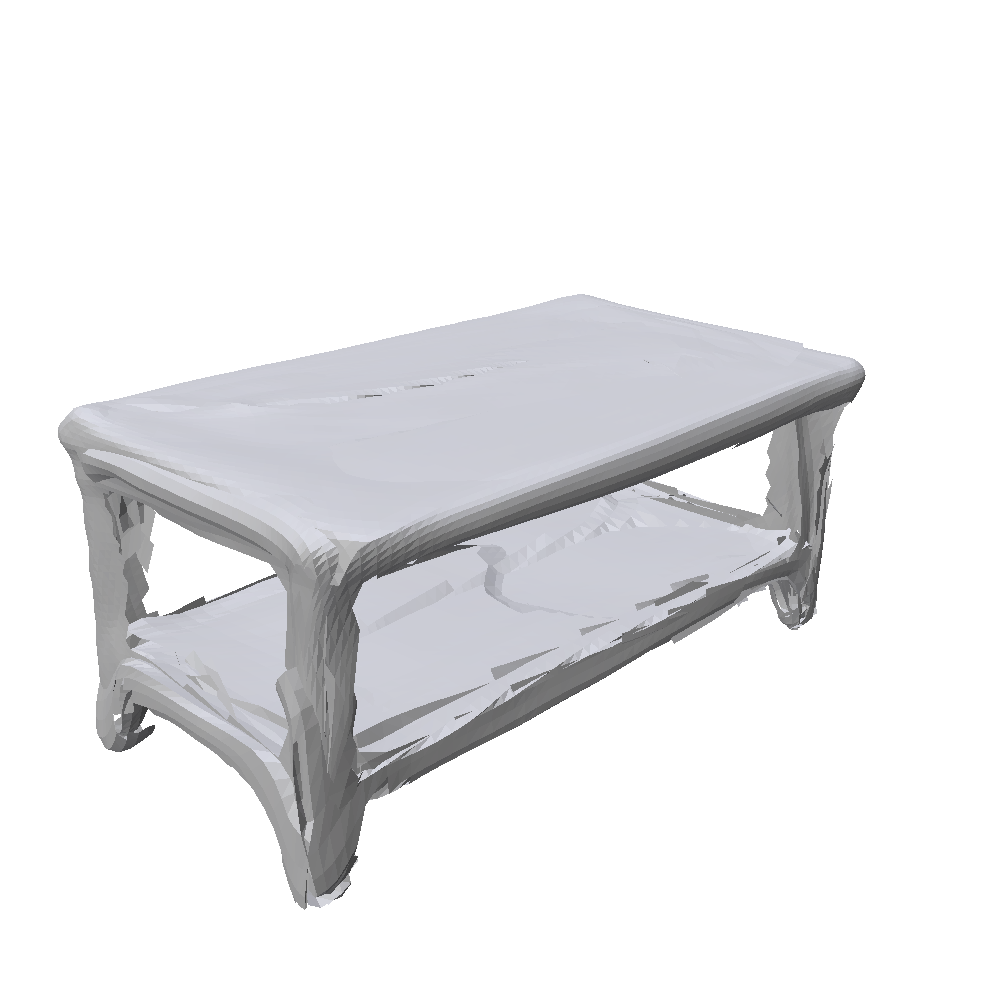}
            \includegraphics[width=1\linewidth,trim={1cm 3cm 2cm 4cm},clip]{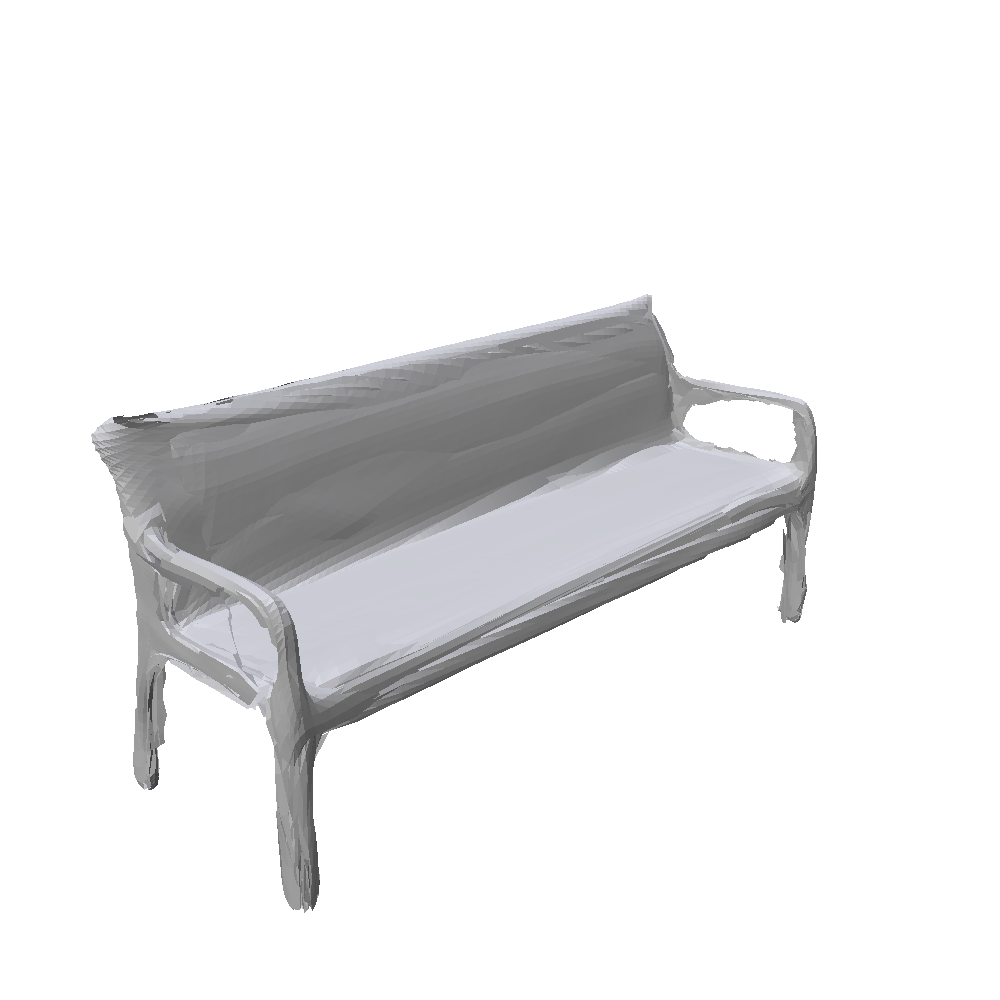}
		\end{subfigure}
        \begin{subfigure}{0.135\linewidth}
    		\centering
            \captionsetup{font=scriptsize,labelfont=scriptsize}
			\caption*{Target}
			\includegraphics[width=1\linewidth,trim={1.5cm 0.5cm 1.5cm 3cm},clip]{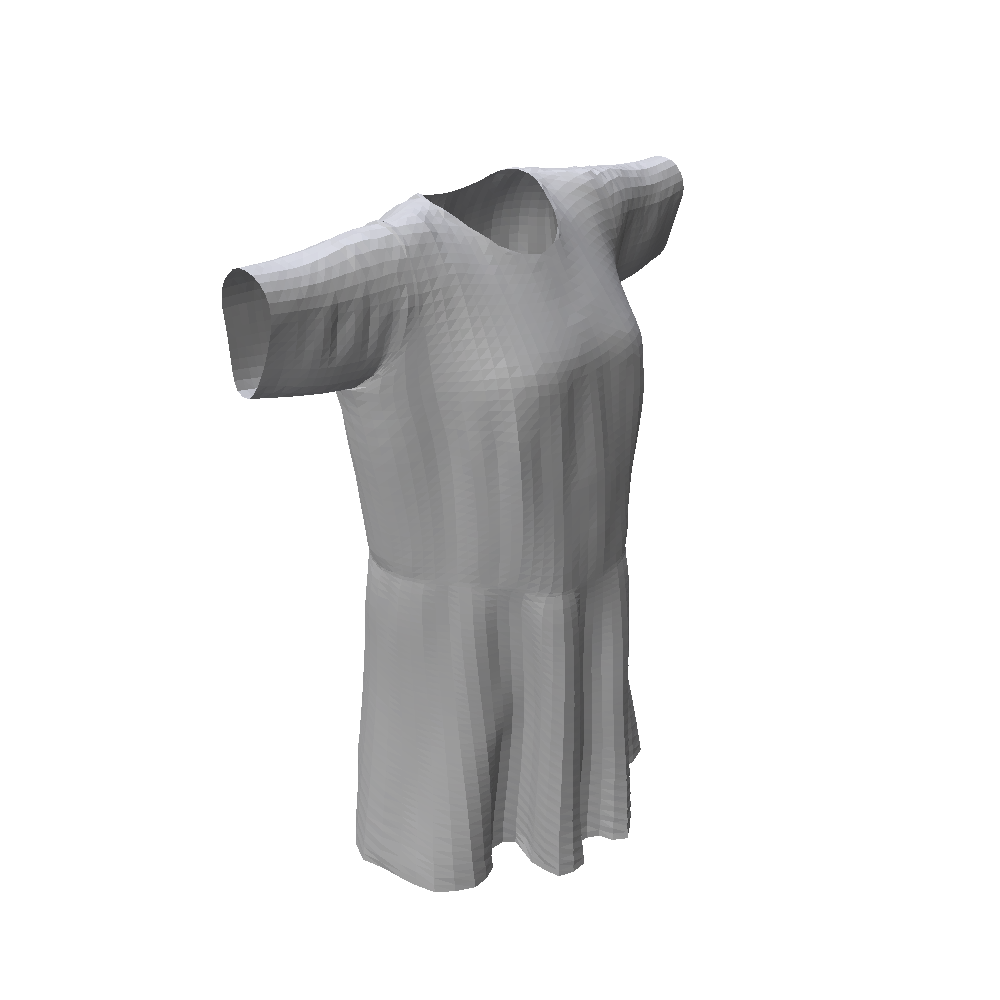}
            \includegraphics[width=1\linewidth,trim={1.5cm 5cm 1.5cm 1cm},clip]{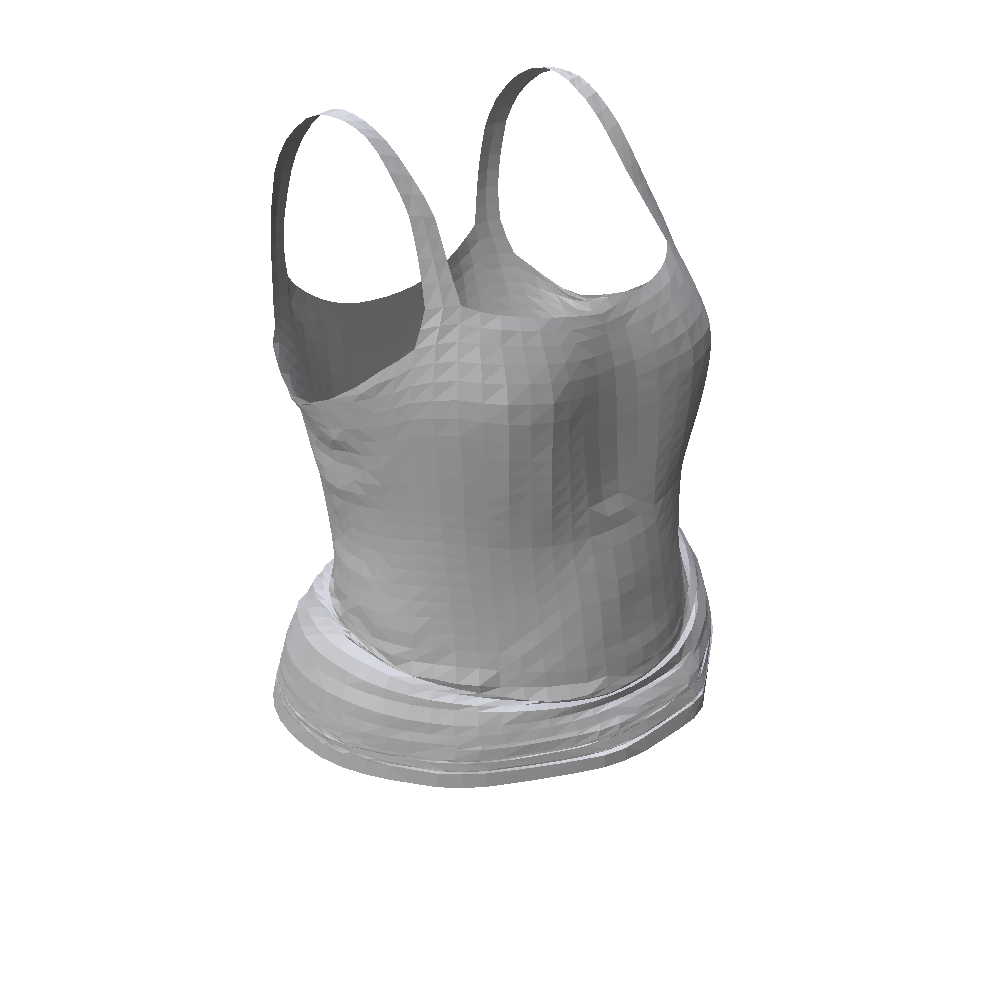}
            \includegraphics[width=1\linewidth,trim={5.5cm 4.5cm 5.5cm 5cm},clip]{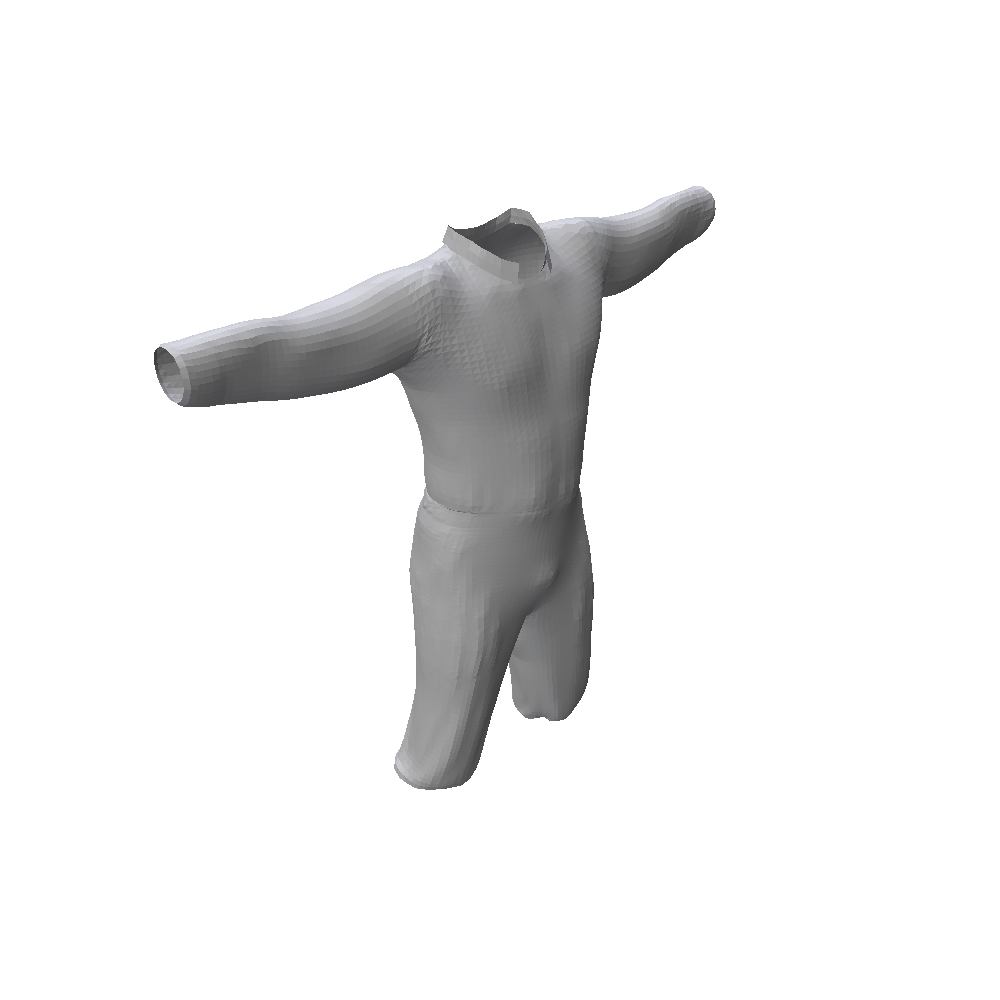}
            
			\includegraphics[width=1\linewidth,trim={6cm 2cm 4cm 8cm},clip]{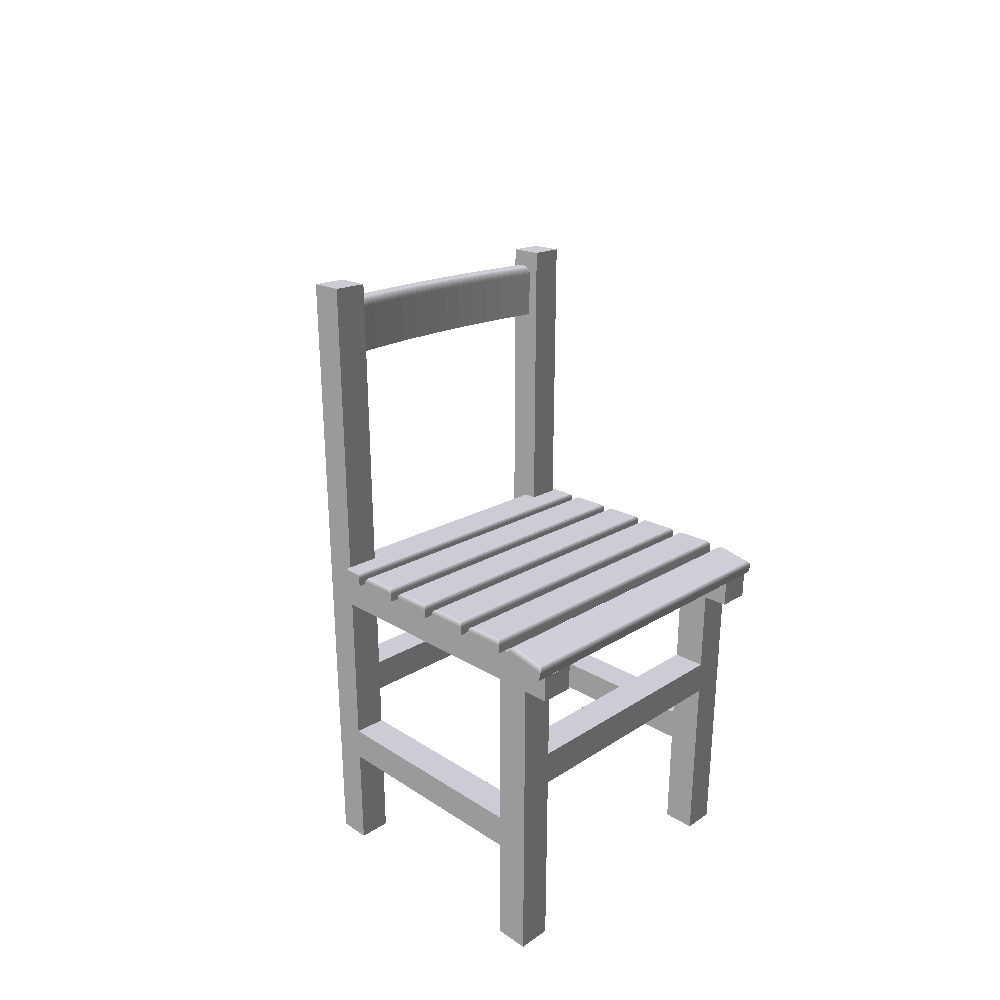}
            \includegraphics[width=1\linewidth,trim={3cm 3cm 4cm 8cm},clip]{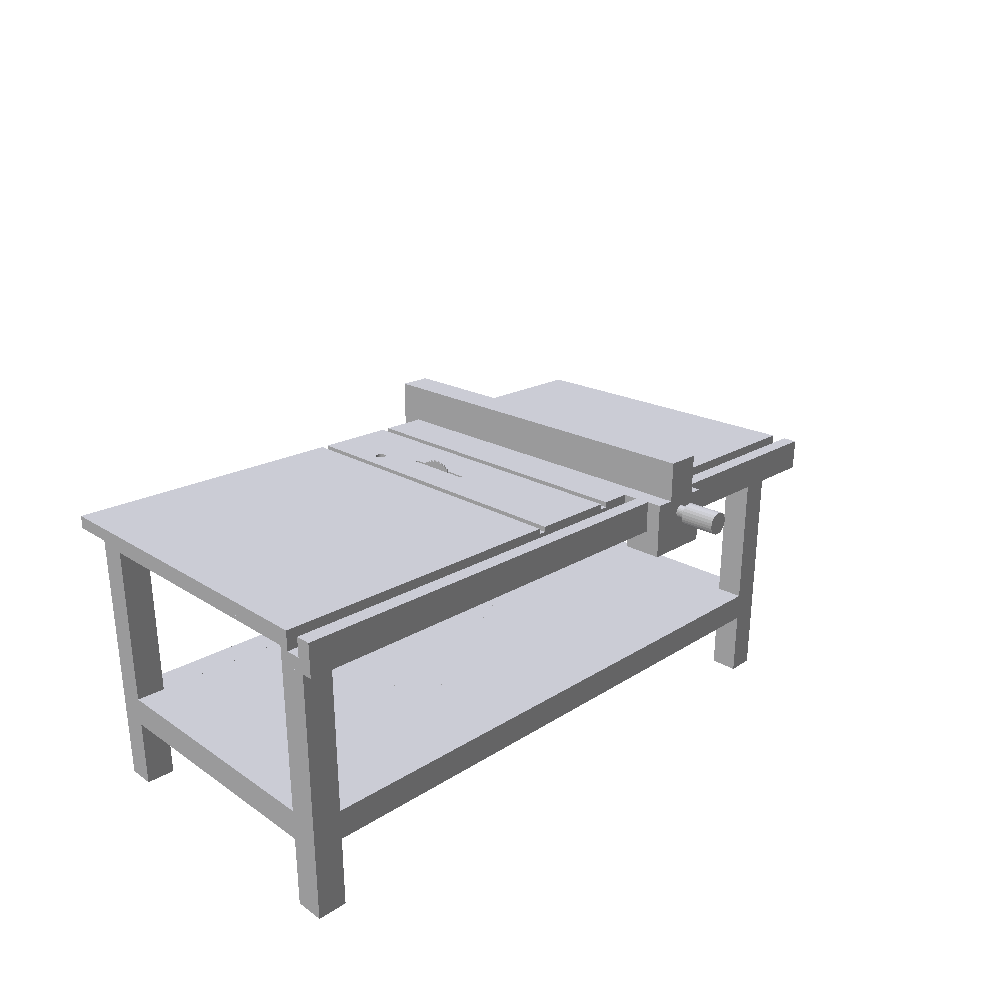}
            \includegraphics[width=1\linewidth,trim={1cm 3cm 2cm 4cm},clip]{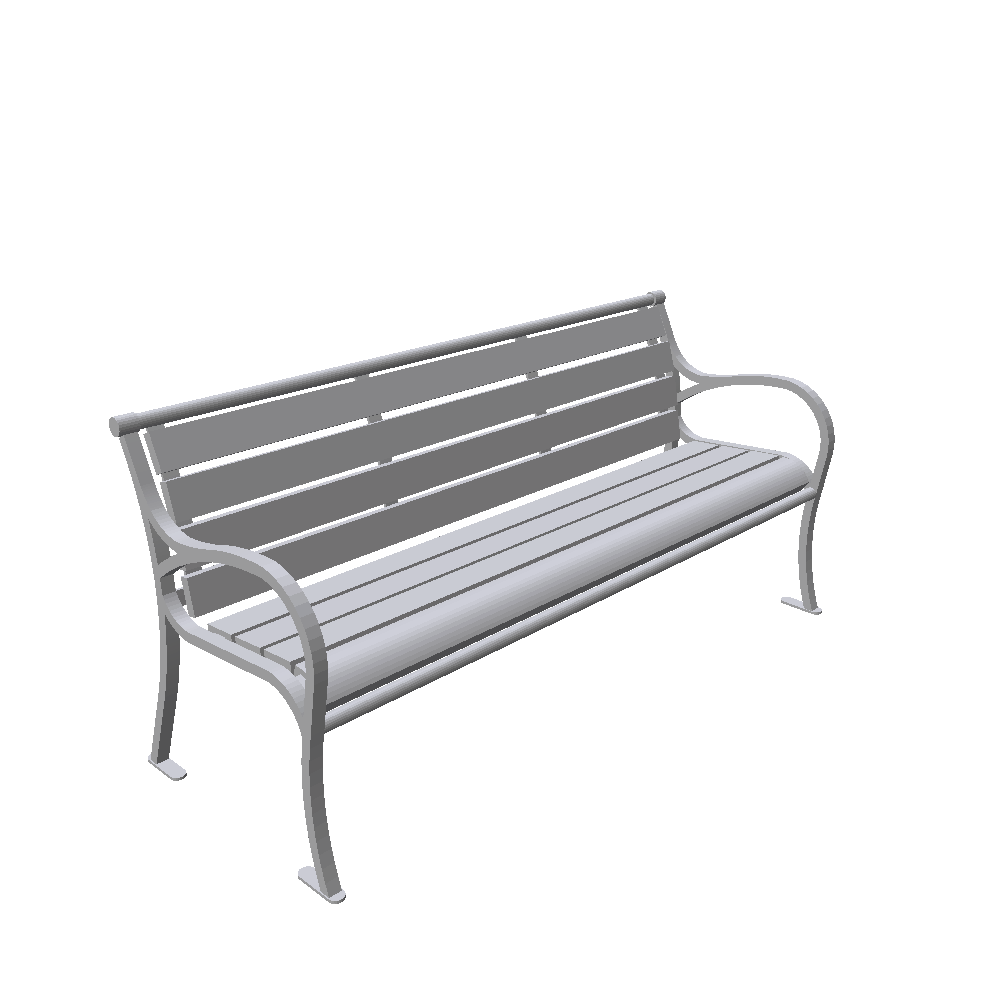}
		\end{subfigure}
        \begin{subfigure}{0.135\linewidth}
    		\centering
            \captionsetup{font=scriptsize,labelfont=scriptsize}
			\caption*{Input}
			\includegraphics[width=1\linewidth,trim={1cm 1cm 1cm 1cm},clip]{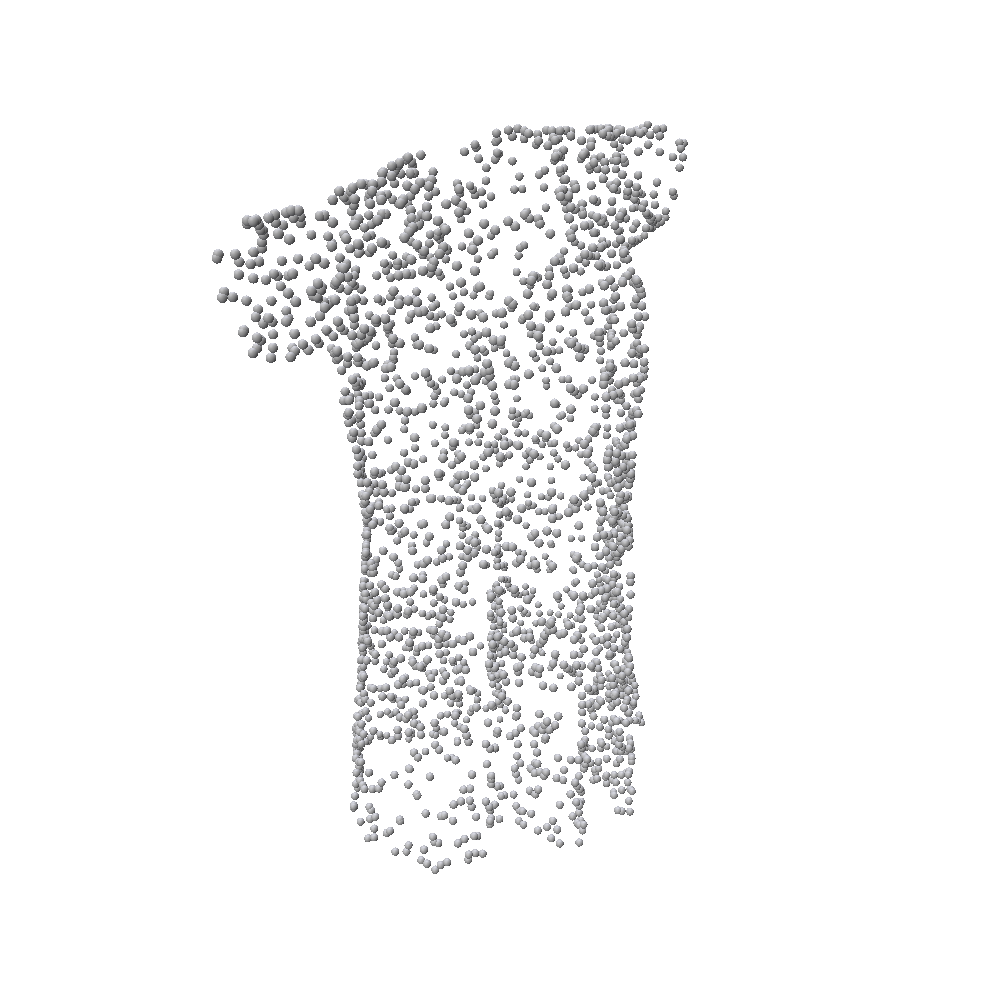}
            \includegraphics[width=1\linewidth,trim={1cm 5cm 1cm 1cm},clip]{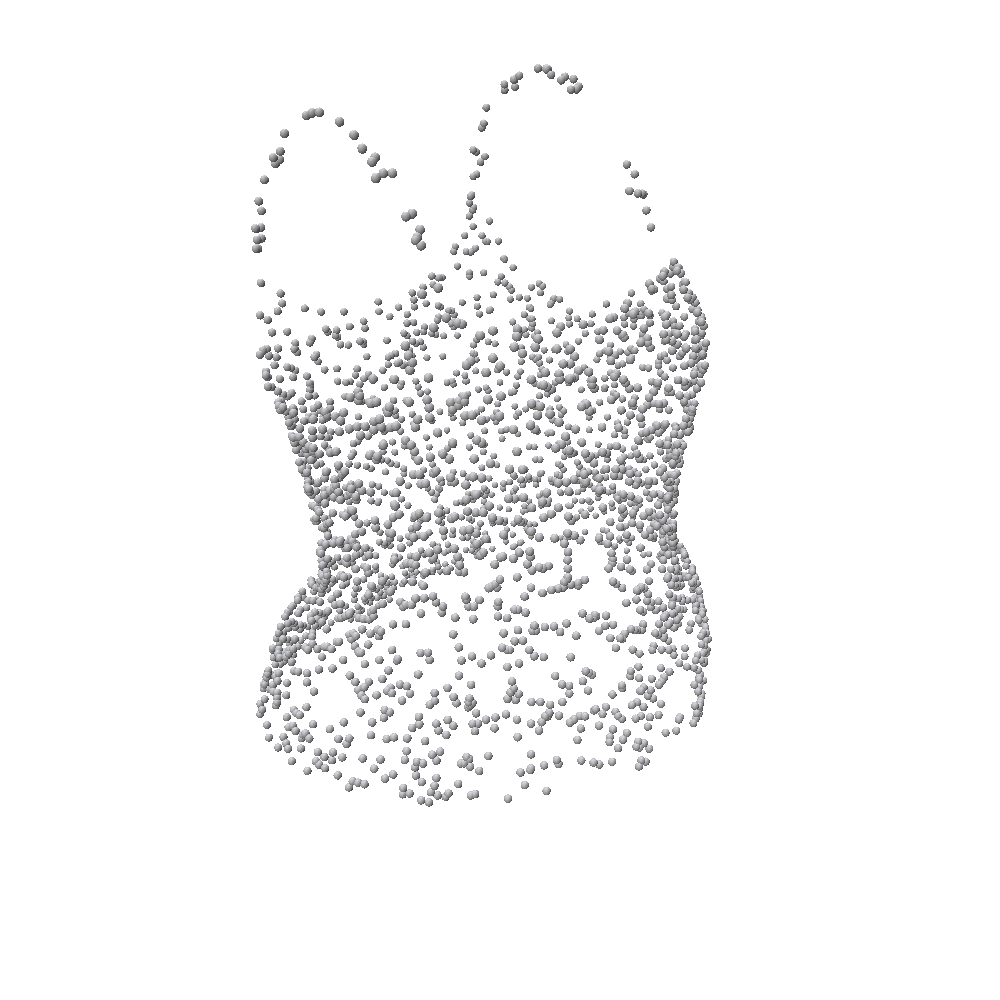}
            \includegraphics[width=1\linewidth,trim={5cm 3cm 5cm 5cm},clip]{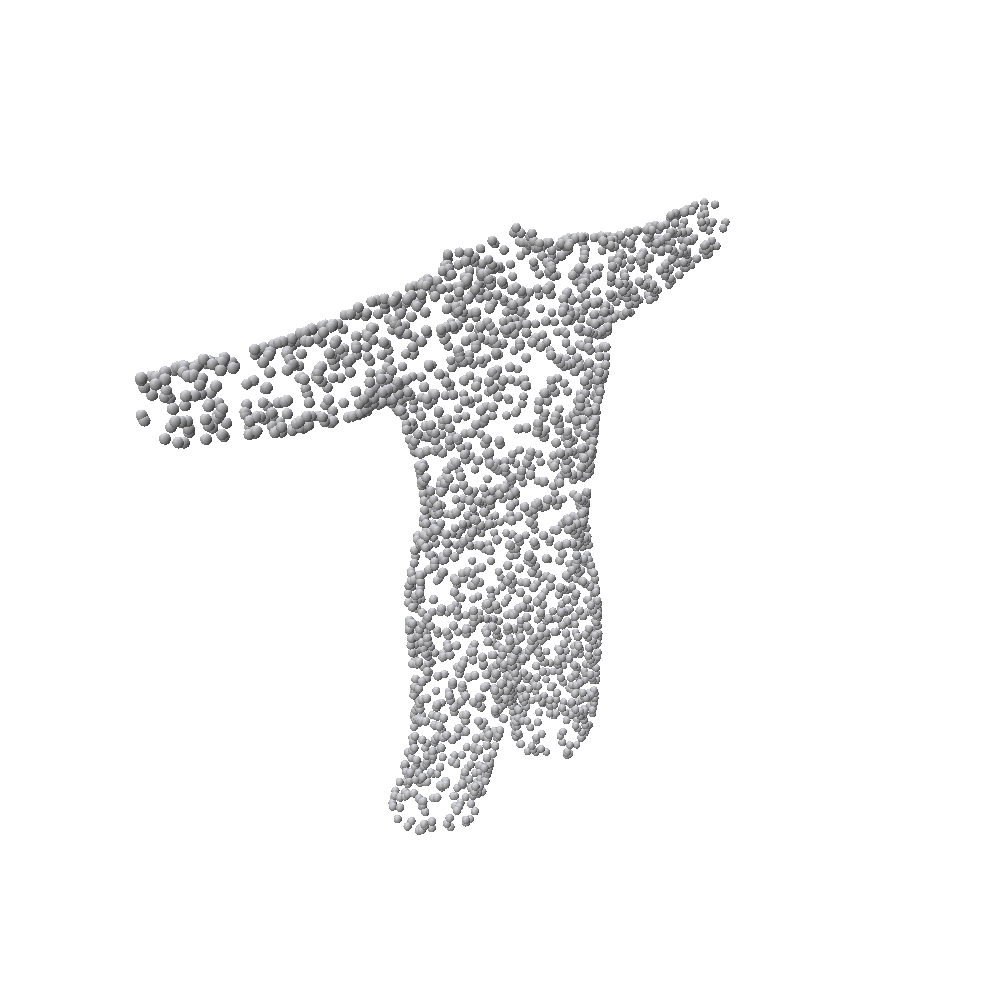}
            
			\includegraphics[width=1\linewidth,trim={3cm 2cm 1cm 1cm},clip]{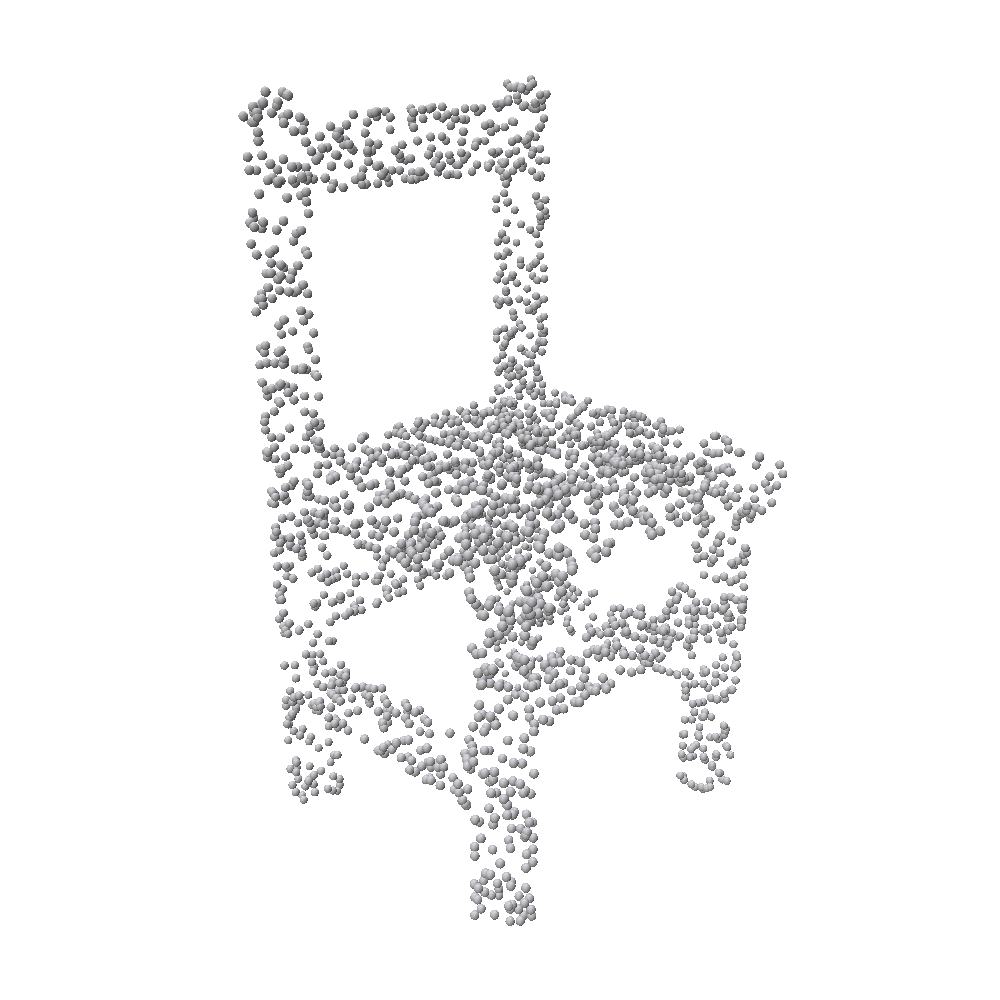}
            \includegraphics[width=1\linewidth,trim={1cm 3cm 2cm 4cm},clip]{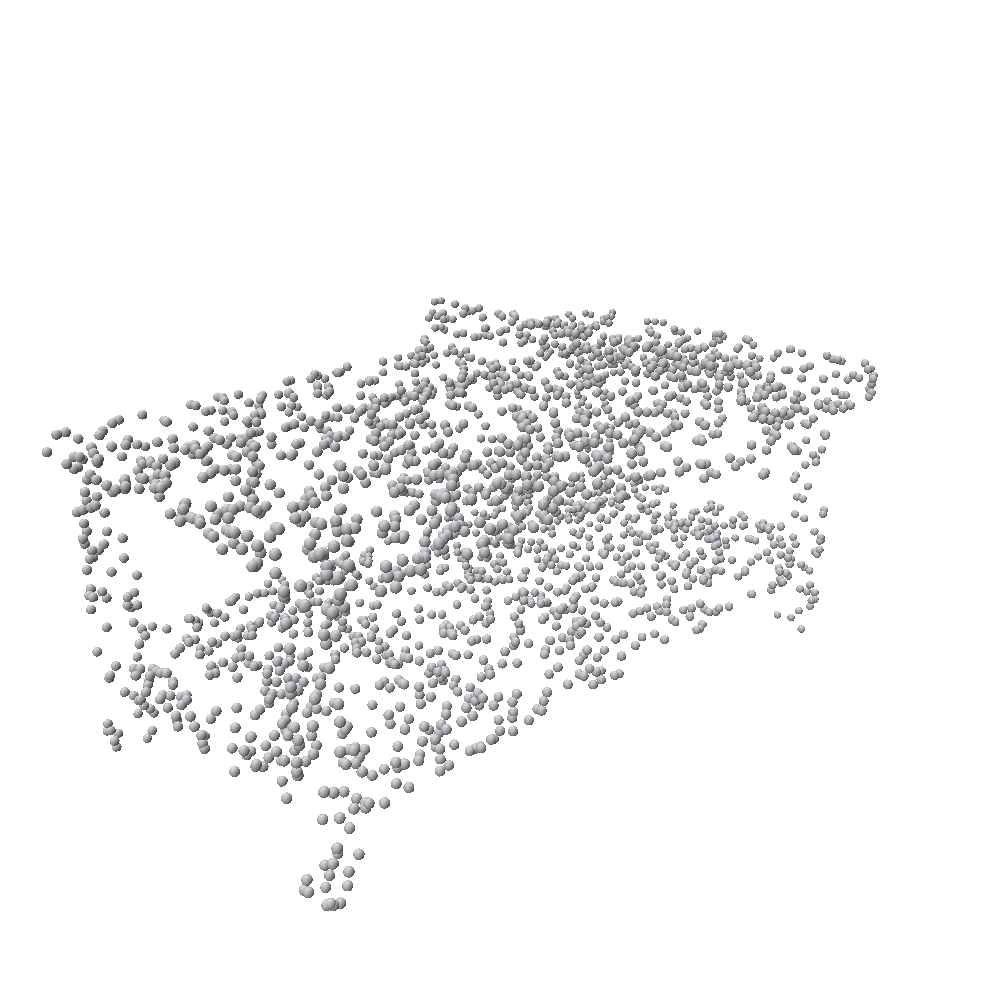}
            \includegraphics[width=1\linewidth,trim={1cm 3cm 2cm 4cm},clip]{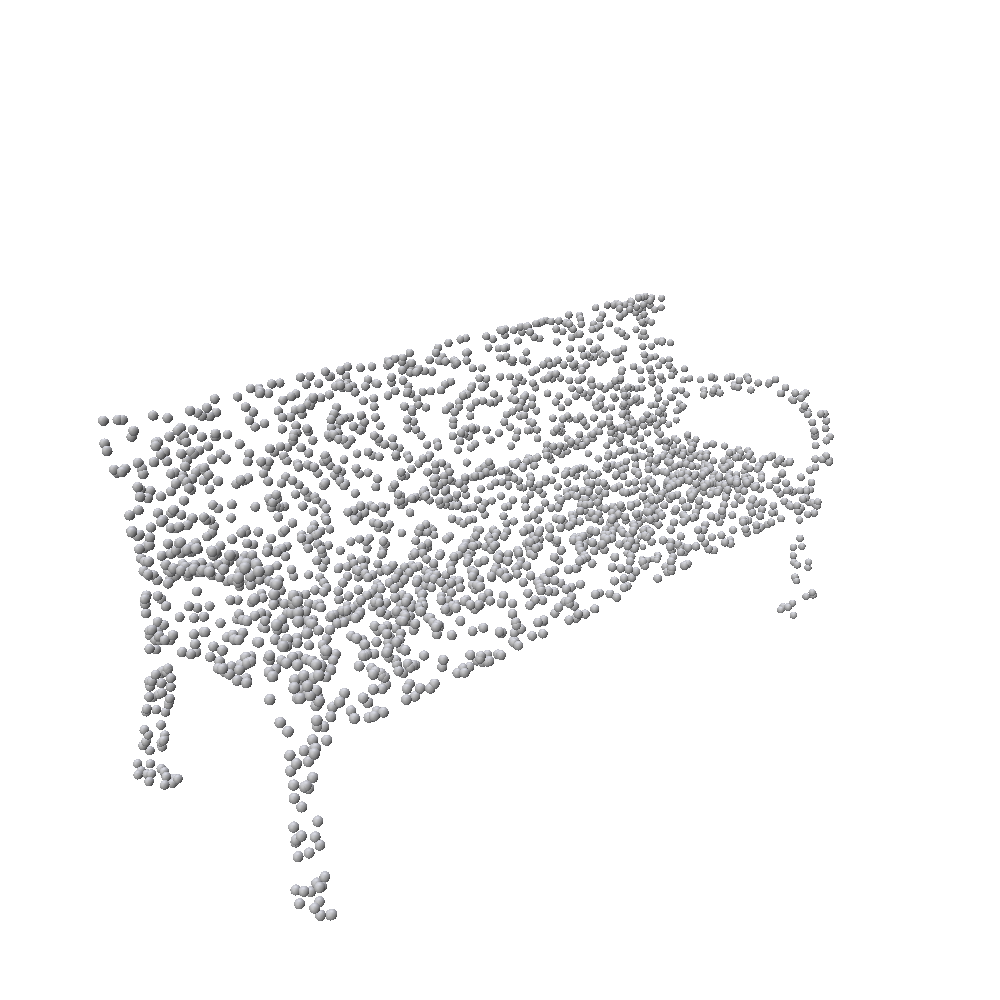}
		\end{subfigure}
        \caption{Surface Reconstruction on CLOTH3D++ and ShapeNet.}
        \label{fig:surfrec}
\end{figure}

\par
In this experiment, we consider the specific setting of reconstructing the target surface given an input point cloud of size 2,500. The input point cloud also serves as the target point cloud for training. This is consistent with previous works such as \cite{groueix2018_atlasnet,bednarik2020_dsp,deng2020_bps,gupta2020_nmf}. Nevertheless, we adopt a larger UV sample size of 5,000 for training in all works.

\par
To evaluate whether a method can faithfully learn a minimal representation, we benchmark all surface representations at 3 different number of charts, 
except for TearingNet. In particular, we evaluate at 1, 2 and 25 charts on CLOTH3D++ and 1, 3 and 25 charts on ShapeNet. The inconsistency of 2 and 3 charts between both datasets is attributed to the fact that these baselines theoretically admit a minimal atlas of 2 and 3 charts for general single-object open and closed surfaces respectively. Benchmarking at 1 and 25 charts, which is the default for the baselines, also allows us to assess the limiting performance of a surface representation as the number of charts decreases or increases, respectively. Furthermore, we also evaluate at 1 chart because our proposed representation admits a minimal atlas of 1 chart for general single-object open surfaces.

\par
The quantitative results for surface reconstruction on CLOTH3D++ and ShapeNet are reported in Table~\ref{table:surfrec:cloth3d++} and~\ref{table:surfrec:shapenet}, respectively. In general, our surface representation achieves higher point cloud reconstruction performance at any given number of charts, especially on the more complex ShapeNet dataset. This indicates that the overall surface geometry is more accurately reconstructed by our representation, which 
can be attributed to the separation of concerns between $o_{\theta_k}$ and $\varphi_{\theta_k}$.  Furthermore, minimal neural atlas significantly outperforms the baselines in terms of the mesh reconstruction accuracy at any given number of charts, which is particularly true for lower number of charts and on ShapeNet. 
Together with the observation that our point cloud and mesh reconstruction metrics are relatively on par with each other, this suggests that minimal neural atlas can also reconstruct the topology of the target surface more accurately. These conclusions are also supported qualitatively in Fig.~\ref{fig:surfrec}, where we show the reconstructed meshes of all surface representations at 2 and 3 charts, except for TearingNet, on CLOTH3D++ and ShapeNet.

\par
The reconstruction metrics of our representation are also substantially more consistent across a wide range of charts. It is also worth noting that despite using fewer charts, minimal neural atlas often outperforms the baselines in terms of reconstruction accuracy. This affirms the ability of our representation to learn a minimal atlas for general surfaces. Moreover, the chart parameterizations of minimal neural atlas exhibit inherently lower distortion as it achieves lower metric values compared to AtlasNet, TearingNet and even AtlasNet++ without explicit regularization of distortion. Unlike DSP, the reported results also indicate that our representation is able to 
significantly reduce distortion without sacrificing reconstruction accuracy by additionally minimizing the metric distortion loss.

\subsection{Single-View Reconstruction}
\label{subsec:exp:svr}

\begin{table}[t]
\centering
\ra{0.925}
\scriptsize
\caption{Single-View Reconstruction on ShapeNet.}
\label{table:svr}

\begin{tabular}{@{}clcclcclccc@{}}
\toprule
 & \multicolumn{1}{c}{} & \multicolumn{2}{c}{Point Cloud} & \phantom{..} & \multicolumn{2}{c}{Mesh} & \phantom{..} & \multicolumn{3}{c}{Distortion} \\ \cmidrule(lr){3-4} \cmidrule(lr){6-7} \cmidrule(lr){9-11} 
\multirow{-3}{*}{\begin{tabular}[c]{@{}c@{}}No. of\\ Charts\end{tabular}} & \multicolumn{1}{c}{\multirow{-3}{*}{\begin{tabular}[c]{@{}c@{}}Surface\\ Representation\end{tabular}}} & CD, $10^{-3} \downarrow$ & F$@1\% \uparrow$ &  & CD, $10^{-3} \downarrow$ & F$@1\% \uparrow$ &  & Metric $\downarrow$ & Conformal $\downarrow$ & Area $\downarrow$ \\ \midrule
 & AtlasNet & \textbf{3.100} & 57.78 &  & 4.512 & 52.85 &  & 34.60 & 4.771 & 3.189 \\
 & AtlasNet++ & {\ul 3.254} & 55.74 &  & 4.049 & 54.13 &  & 33.48 & 6.579 & 3.657 \\
 & DSP & 5.210 & 46.28 &  & 6.048 & 44.55 &  & \textbf{1.113} & \textbf{0.3718} & \textbf{0.1294} \\
 & TearingNet & 3.633 & 55.27 &  & 5.250 & 51.65 &  & 30.52 & 5.610 & 2.778 \\
 & \cellcolor[HTML]{EFEFEF}Ours w/o $\mathcal{L}_{dist}$ & \cellcolor[HTML]{EFEFEF}3.754 & \cellcolor[HTML]{EFEFEF}\textbf{61.63} & \cellcolor[HTML]{EFEFEF} & \cellcolor[HTML]{EFEFEF}\textbf{3.891} & \cellcolor[HTML]{EFEFEF}\textbf{60.38} & \cellcolor[HTML]{EFEFEF} & \cellcolor[HTML]{EFEFEF}8.148 & \cellcolor[HTML]{EFEFEF}2.510 & \cellcolor[HTML]{EFEFEF}0.4778 \\
\multirow{-6}{*}{1} & \cellcolor[HTML]{EFEFEF}Ours & \cellcolor[HTML]{EFEFEF}3.840 & \cellcolor[HTML]{EFEFEF}{\ul 59.90} & \cellcolor[HTML]{EFEFEF} & \cellcolor[HTML]{EFEFEF}{\ul 4.002} & \cellcolor[HTML]{EFEFEF}{\ul 58.71} & \cellcolor[HTML]{EFEFEF} & \cellcolor[HTML]{EFEFEF}{\ul 2.674} & \cellcolor[HTML]{EFEFEF}{\ul 0.9069} & \cellcolor[HTML]{EFEFEF}{\ul 0.2420} \\ \midrule
 & AtlasNet & \textbf{2.992} & 59.08 &  & 4.125 & 54.92 &  & 25.00 & 4.759 & 2.059 \\
 & AtlasNet++ & {\ul 3.077} & 57.84 &  & {\ul 3.706} & 56.26 &  & 38.07 & 6.263 & 3.419 \\
 & DSP & 4.447 & 50.12 &  & 5.096 & 48.11 &  & \textbf{0.5316} & \textbf{0.1505} & \textbf{0.0901} \\
 & \cellcolor[HTML]{EFEFEF}Ours w/o $\mathcal{L}_{dist}$ & \cellcolor[HTML]{EFEFEF}3.582 & \cellcolor[HTML]{EFEFEF}\textbf{62.11} & \cellcolor[HTML]{EFEFEF} & \cellcolor[HTML]{EFEFEF}\textbf{3.704} & \cellcolor[HTML]{EFEFEF}\textbf{60.68} & \cellcolor[HTML]{EFEFEF} & \cellcolor[HTML]{EFEFEF}9.972 & \cellcolor[HTML]{EFEFEF}3.177 & \cellcolor[HTML]{EFEFEF}0.5906 \\
\multirow{-5}{*}{3} & \cellcolor[HTML]{EFEFEF}Ours & \cellcolor[HTML]{EFEFEF}3.621 & \cellcolor[HTML]{EFEFEF}{\ul 61.90} & \cellcolor[HTML]{EFEFEF} & \cellcolor[HTML]{EFEFEF}3.744 & \cellcolor[HTML]{EFEFEF}{\ul 60.56} & \cellcolor[HTML]{EFEFEF} & \cellcolor[HTML]{EFEFEF}{\ul 1.776} & \cellcolor[HTML]{EFEFEF}{\ul 0.5537} & \cellcolor[HTML]{EFEFEF}{\ul 0.2304} \\ \midrule
 & AtlasNet & \textbf{2.883} & 60.68 &  & 3.655 & 57.21 &  & 18.77 & 4.853 & 1.516 \\
 & AtlasNet++ & {\ul 2.961} & 59.40 &  & {\ul 3.469} & 57.88 &  & 25.76 & 6.201 & 2.522 \\
 & DSP & 3.582 & 55.60 &  & 4.336 & 52.94 &  & {\ul 1.492} & {\ul 0.3712} & {\ul 0.2254} \\
 & \cellcolor[HTML]{EFEFEF}Ours w/o $\mathcal{L}_{dist}$ & \cellcolor[HTML]{EFEFEF}3.413 & \cellcolor[HTML]{EFEFEF}{\ul 62.71} & \cellcolor[HTML]{EFEFEF} & \cellcolor[HTML]{EFEFEF}\textbf{3.464} & \cellcolor[HTML]{EFEFEF}{\ul 61.42} & \cellcolor[HTML]{EFEFEF} & \cellcolor[HTML]{EFEFEF}6.620 & \cellcolor[HTML]{EFEFEF}2.206 & \cellcolor[HTML]{EFEFEF}0.4532 \\
\multirow{-5}{*}{25} & \cellcolor[HTML]{EFEFEF}Ours & \cellcolor[HTML]{EFEFEF}3.437 & \cellcolor[HTML]{EFEFEF}\textbf{63.05} & \cellcolor[HTML]{EFEFEF} & \cellcolor[HTML]{EFEFEF}3.514 & \cellcolor[HTML]{EFEFEF}\textbf{61.93} & \cellcolor[HTML]{EFEFEF} & \cellcolor[HTML]{EFEFEF}\textbf{1.037} & \cellcolor[HTML]{EFEFEF}\textbf{0.3321} & \cellcolor[HTML]{EFEFEF}\textbf{0.1497} \\ \bottomrule
\end{tabular}
\end{table}

\par
This experiment adopts the exact same setting as the surface reconstruction experiment, except the input is an image of the target surface. The quantitative results on ShapeNet reported in Table~\ref{table:svr} remains largely similar to the previous experiment. While our representation incurs a relatively higher point cloud CD, it consistently achieves a higher reconstruction performance on the more representative F@1\% metric \cite{tatarchenko2019_fscore_propose}. 
We can thus reach to the same conclusions, as per the previous experiment.

\subsection{Ablation Studies}
\label{subsec:exp:ablation}

\begin{table}[t]
\centering
\ra{0.925}
\scriptsize
\caption{Ablation Study of Minimal Neural Atlas with $\mathcal{L}_{dist}$.}
\label{table:ablation:reform_factor_pe}

\begin{tabular}{@{}lllcclcclclc@{}}
\toprule
\multicolumn{2}{c}{\multirow{3}{*}{Variant}} & \phantom{..} & \multicolumn{2}{c}{Point Cloud} & \phantom{...} & \multicolumn{2}{c}{Mesh} & \phantom{..} & \multirow{3}{*}{\begin{tabular}[c]{@{}c@{}}Metric\\ Distortion\end{tabular}} \multirow{3}{*}{\begin{tabular}[c]{@{}c@{}}$\downarrow$\end{tabular}} & \phantom{} & \multirow{3}{*}{\begin{tabular}[c]{@{}c@{}}Occupancy\\ Rate\end{tabular}} \multirow{3}{*}{\begin{tabular}[c]{@{}c@{}}$\uparrow$\end{tabular}} \\ \cmidrule(lr){4-5} \cmidrule(lr){7-8}
\multicolumn{2}{c}{} &  & CD, $10^{-4} \downarrow$ & F$@1\% \uparrow$ &  & CD, $10^{-4} \downarrow$ & F$@1\% \uparrow$ &  & \multicolumn{1}{l}{} &  &  \\ \midrule
\multicolumn{2}{l}{No $o_{\theta_k}$ reformulation} &  & 15.60 & 69.12 &  & 16.59 & 67.31 &  & 2.348 &  & 80.49 \\
\multicolumn{2}{l}{No $\tilde o_{\theta_k}$ factorization} &  & 327.8 & 29.51 &  & 386.3 & 27.59 &  & 4.951 &  & 6.762 \\
\multicolumn{2}{l}{No $\tilde l_{\theta_k}$ pos. encoding} &  & 7.419 & 82.04 &  & 8.050 & 80.23 &  & \textbf{1.848} &  & \textbf{83.45} \\
\multicolumn{2}{l}{Full Model} &  & \textbf{6.311} & \textbf{83.63} &  & \textbf{6.761} & \textbf{82.23} &  & 2.189 &  & 77.48 \\ \bottomrule
\end{tabular}
\end{table}

\par
The ablation studies are conducted in the same setting as surface reconstruction on ShapeNet using 3 charts. The results reported in Table~\ref{table:ablation:reform_factor_pe} verifies the immense importance of decoupling the homeomorphic ambiguity by reformulating $o_{\theta_k}$ with Eq.~\ref{eq:o_reform}, as well as casting the learning of $\tilde o_{\theta_k}$ as a PU Learning problem, which can be easily solved given the factorization of $\tilde o_{\theta_k}$ in Eq.~\ref{eq:o-tilde_factor}, instead of a naïve binary classification problem. It also shows that it is crucial to apply positional embedding \cite{mildenhall2020_nerf,vaswani2017_transformer} on the input maximal point coordinates of $\tilde l_{\theta_k}$ 
to learn a more detailed occupancy field for better reconstructions, albeit at a minor cost of distortion.

\section{Conclusion}
\label{sec:conclusion}

\par
In this paper, we propose \textit{Minimal Neural Atlas}, a novel explicit neural surface representation that can effectively learn a minimal atlas with distortion-minimal parameterization for general surfaces of arbitrary topology, which is enabled by a fully learnable parametric domain. Despite its achievements, our representation remains prone to artifacts common in atlas-based representations, such as intersections and seams between charts. Severe violation of the SCAR assumption, due to imperfect modeling of the target surface, non-matching sampling distribution \textit{etc.}, also leads to unintended holes on the reconstructed surface, which we leave for future work. Although we motivated this work in the context of representing surfaces, our representation naturally extends to general $n$-manifolds. It would thus be interesting to explore its applications in other domains.

\bfPar{Acknowledgements.}
This research/project is supported by the National Research Foundation, Singapore under its AI Singapore Programme (AISG Award No: AISG2-RP-2021-024), and the Tier 2 grant MOE-T2EP20120-0011 from the Singapore Ministry
of Education.

%
%
\bibliographystyle{splncs04}
\bibliography{references}

\begin{thebibliography}{10}
\providecommand{\url}[1]{\texttt{#1}}
\providecommand{\urlprefix}{URL }
\providecommand{\doi}[1]{https://doi.org/#1}

\bibitem{atzmon2020_sal}
Atzmon, M., Lipman, Y.: {SAL: Sign agnostic learning of shapes from raw data}.
  In: Proceedings of the IEEE Computer Society Conference on Computer Vision
  and Pattern Recognition (2020)

\bibitem{atzmon2021_sald}
Atzmon, M., Lipman, Y.: {SALD}: Sign agnostic learning with derivatives. In:
  International Conference on Learning Representations (2021)

\bibitem{badki2020_meshlet}
Badki, A., Gallo, O., Kautz, J., Sen, P.: Meshlet priors for 3d mesh
  reconstruction. In: 2020 IEEE/CVF Conference on Computer Vision and Pattern
  Recognition (CVPR) (2020)

\bibitem{ma2021_neural_pull}
Baorui, M., Zhizhong, H., Yu-shen, L., Matthias, Z.: {Neural-Pull: Learning
  Signed Distance Functions from Point Clouds by Learning to Pull Space onto
  Surfaces}. In: International Conference on Machine Learning (ICML) (2021)

\bibitem{bednarik2020_dsp}
Bednarik, J., Parashar, S., Gundogdu, E., Salzmann, M., Fua, P.: {Shape
  Reconstruction by Learning Differentiable Surface Representations}. In:
  Proceedings of the IEEE Computer Society Conference on Computer Vision and
  Pattern Recognition (2020)

\bibitem{bekker2020_pu_learning_survey}
Bekker, J., Davis, J.: {Learning from Positive and Unlabeled Data: A Survey}.
  Machine Learning  (2020)

\bibitem{boulch2021_needrop}
Boulch, A., Langlois, P., Puy, G., Marlet, R.: Needrop: Self-supervised shape
  representation from sparse point clouds using needle dropping. In: 2021
  International Conference on 3D Vision (3DV) (2021)

\bibitem{chang2015_shapenet}
Chang, A.X., Funkhouser, T., Guibas, L., Hanrahan, P., Huang, Q., Li, Z.,
  Savarese, S., Savva, M., Song, S., Su, H., Xiao, J., Yi, L., Yu, F.:
  {ShapeNet: An Information-Rich 3D Model Repository}. Tech. Rep.
  arXiv:1512.03012 [cs.GR], Stanford University --- Princeton University ---
  Toyota Technological Institute at Chicago (2015)

\bibitem{charles2017_pointnet}
Charles, R.Q., Su, H., Kaichun, M., Guibas, L.J.: Pointnet: Deep learning on
  point sets for 3d classification and segmentation. In: 2017 IEEE Conference
  on Computer Vision and Pattern Recognition (CVPR) (2017)

\bibitem{chen2019_im-net}
{Chen}, Z., {Zhang}, H.: Learning implicit fields for generative shape
  modeling. In: 2019 IEEE/CVF Conference on Computer Vision and Pattern
  Recognition (CVPR) (2019)

\bibitem{chibane2020_ndf}
Chibane, J., Mir, A., Pons-Moll, G.: {Neural Unsigned Distance Fields for
  Implicit Function Learning}. In: Advances in Neural Information Processing
  Systems (2020)

\bibitem{cornea2003_ls_cat3}
Cornea, O., Lupton, G., Oprea, J., Tanr{\'e}, D.: Lusternik-Schnirelmann
  category. American Mathematical Soc. (2003)

\bibitem{degener2003_equiarea}
Degener, P., Meseth, J., Klein, R.: An adaptable surface parameterization
  method. In: IMR (2003)

\bibitem{deng2009_imagenet}
Deng, J., Dong, W., Socher, R., Li, L.J., Li, K., Fei-Fei, L.: Imagenet: A
  large-scale hierarchical image database. In: 2009 IEEE Conference on Computer
  Vision and Pattern Recognition (2009)

\bibitem{deng2020_bps}
Deng, Z., Bednarik, J., Salzmann, M., Fua, P.: {Better Patch Stitching for
  Parametric Surface Reconstruction}. In: Proceedings - 2020 International
  Conference on 3D Vision, 3DV 2020 (2020)

\bibitem{elkan2008_pu_learning}
Elkan, C., Noto, K.: Learning classifiers from only positive and unlabeled
  data. In: Proceedings of the 14th ACM SIGKDD International Conference on
  Knowledge Discovery and Data Mining (2008)

\bibitem{fan2017_psgn}
Fan, H., Su, H., Guibas, L.: A point set generation network for 3d object
  reconstruction from a single image. In: 2017 IEEE Conference on Computer
  Vision and Pattern Recognition (CVPR) (2017)

\bibitem{fox1941_ls_cat1}
Fox, R.H.: On the lusternik-schnirelmann category. Annals of Mathematics
  (1941)

\bibitem{gropp2020_igr}
Gropp, A., Yariv, L., Haim, N., Atzmon, M., Lipman, Y.: {Implicit Geometric
  Regularization for Learning Shapes}. In: International Conference on Machine
  Learning (2020)

\bibitem{groueix2018_atlasnet}
Groueix, T., Fisher, M., Kim, V.G., Russell, B.C., Aubry, M.: {A
  Papier-M{\^a}ch{\'e} Approach to Learning 3D Surface Generation}. In:
  Proceedings of the IEEE Computer Society Conference on Computer Vision and
  Pattern Recognition (2018)

\bibitem{gupta2020_nmf}
Gupta, K., Chandraker, M.: {Neural Mesh Flow: 3D Manifold Mesh Generation via
  Diffeomorphic Flows}. In: Advances in Neural Information Processing Systems
  (2020)

\bibitem{kaiming2016_resnet}
He, K., Zhang, X., Ren, S., Sun, J.: Deep residual learning for image
  recognition. In: 2016 IEEE Conference on Computer Vision and Pattern
  Recognition (CVPR) (2016)

\bibitem{hormann2000_mips}
Hormann, K., Greiner, G.: Mips: An efficient global parametrization method.
  Tech. rep., Erlangen-Nuernberg Univ (Germany) Computer Graphics Group (2000)

\bibitem{ioffe2015_bn}
Ioffe, S., Szegedy, C.: Batch normalization: Accelerating deep network training
  by reducing internal covariate shift. In: International conference on machine
  learning (2015)

\bibitem{james1978_ls_cat2}
James, I.: On category, in the sense of lusternik-schnirelmann. Topology
  (1978)

\bibitem{kingma2015_adam}
Kingma, D.P., Ba, J.: Adam: {A} method for stochastic optimization. In: 3rd
  International Conference on Learning Representations, {ICLR} 2015, San Diego,
  CA, USA, May 7-9, 2015, Conference Track Proceedings (2015)

\bibitem{knapitsch2017_fscore_original}
Knapitsch, A., Park, J., Zhou, Q.Y., Koltun, V.: Tanks and temples:
  Benchmarking large-scale scene reconstruction. ACM Transactions on Graphics
  (2017)

\bibitem{madadi2021_cloth3d++}
Madadi, M., Bertiche, H., Bouzouita, W., Guyon, I., Escalera, S.: Learning
  cloth dynamics: 3d + texture garment reconstruction benchmark. In:
  Proceedings of the NeurIPS 2020 Competition and Demonstration Track, PMLR
  (2021)

\bibitem{mescheder2019_onet}
{Mescheder}, L., {Oechsle}, M., {Niemeyer}, M., {Nowozin}, S., {Geiger}, A.:
  Occupancy networks: Learning 3d reconstruction in function space. In: 2019
  IEEE/CVF Conference on Computer Vision and Pattern Recognition (CVPR) (2019)

\bibitem{mildenhall2020_nerf}
Mildenhall, B., Srinivasan, P.P., Tancik, M., Barron, J.T., Ramamoorthi, R.,
  Ng, R.: {NeRF: Representing Scenes as Neural Radiance Fields for View
  Synthesis}. In: European Conference on Computer Vision (ECCV) 2020 (2020)

\bibitem{morreale2021_neural_surface_maps}
Morreale, L., Aigerman, N., Kim, V., Mitra, N.J.: Neural surface maps. In: 2021
  IEEE/CVF Conference on Computer Vision and Pattern Recognition (CVPR) (2021)

\bibitem{pang2021_tearingnet}
Pang, J., Li, D., Tian, D.: {TearingNet: Point Cloud Autoencoder to Learn
  Topology-Friendly Representations}. In: 2021 IEEE/CVF Conference on Computer
  Vision and Pattern Recognition (CVPR) (2021)

\bibitem{park2019_deepsdf}
{Park}, J.J., {Florence}, P., {Straub}, J., {Newcombe}, R., {Lovegrove}, S.:
  Deepsdf: Learning continuous signed distance functions for shape
  representation. In: 2019 IEEE/CVF Conference on Computer Vision and Pattern
  Recognition (CVPR) (2019)

\bibitem{paszke2019_pytorch}
Paszke, A., Gross, S., Massa, F., Lerer, A., Bradbury, J., Chanan, G., Killeen,
  T., Lin, Z., Gimelshein, N., Antiga, L., Desmaison, A., K\"{o}pf, A., Yang,
  E., DeVito, Z., Raison, M., Tejani, A., Chilamkurthy, S., Steiner, B., Fang,
  L., Bai, J., Chintala, S.: Pytorch: An imperative style, high-performance
  deep learning library. In: Proceedings of the 33rd International Conference
  on Neural Information Processing Systems (2019)

\bibitem{rabinovich_sde}
Rabinovich, M., Poranne, R., Panozzo, D., Sorkine-Hornung, O.: Scalable locally
  injective mappings. ACM Transactions on Graphics  (2017)

\bibitem{salimans2016_weight_norm}
Salimans, T., Kingma, D.P.: Weight normalization: A simple reparameterization
  to accelerate training of deep neural networks. In: Proceedings of the 30th
  International Conference on Neural Information Processing Systems (2016)

\bibitem{schreiner2004_sde_real2}
Schreiner, J., Asirvatham, A., Praun, E., Hoppe, H.: Inter-surface mapping. ACM
  Transactions on Graphics  (2004)

\bibitem{smith2015_sde_real}
Smith, J., Schaefer, S.: Bijective parameterization with free boundaries. ACM
  Trans. Graph.  (2015)

\bibitem{tatarchenko2019_fscore_propose}
Tatarchenko, M., Richter, S.R., Ranftl, R., Li, Z., Koltun, V., Brox, T.: {What
  Do Single-View 3D Reconstruction Networks Learn?} In: 2019 IEEE/CVF
  Conference on Computer Vision and Pattern Recognition (CVPR) (2019)

\bibitem{vaswani2017_transformer}
Vaswani, A., Shazeer, N., Parmar, N., Uszkoreit, J., Jones, L., Gomez, A.N.,
  Kaiser, {\L}., Polosukhin, I.: {Attention is All you Need}. In: Advances in
  Neural Information Processing Systems (2017)

\bibitem{williams2019_dgp}
Williams, F., Schneider, T., Silva, C., Zorin, D., Bruna, J., Panozzo, D.:
  {Deep Geometric Prior for Surface Reconstruction}. In: Proceedings of the
  IEEE Computer Society Conference on Computer Vision and Pattern Recognition
  (2019)

\bibitem{yang2018_foldingnet}
Yang, Y., Feng, C., Shen, Y., Tian, D.: Foldingnet: Point cloud auto-encoder
  via deep grid deformation. In: 2018 IEEE/CVF Conference on Computer Vision
  and Pattern Recognition (CVPR) (2018)

\bibitem{yariv2020_idr}
Yariv, L., Kasten, Y., Moran, D., Galun, M., Atzmon, M., Ronen, B., Lipman, Y.:
  {Multiview Neural Surface Reconstruction by Disentangling Geometry and
  Appearance}. In: Advances in Neural Information Processing Systems (2020)

\end{thebibliography}

\clearpage
\title{Supplementary Material for\\Minimal Neural Atlas: Parameterizing Complex\\Surfaces with Minimal Charts and Distortion} 

\titlerunning{Supplementary Material for Minimal Neural Atlas}
%
\author{Weng Fei Low\orcidlink{0000-0001-7022-5713}\index{Low, Weng Fei} \and
Gim Hee Lee\orcidlink{0000-0002-1583-0475}}\index{Lee, Gim Hee}

\authorrunning{W. F. Low and G. H. Lee}
%
\institute{
Institute of Data Science (IDS), National University of Singapore\\
NUS Graduate School's Integrative Sciences and Engineering Programme (ISEP)\\
Department of Computer Science, National University of Singapore\\
\email{\{wengfei.low, gimhee.lee\}@comp.nus.edu.sg}\\
\url{https://github.com/low5545/minimal-neural-atlas}
}
\maketitle
\appendix
\setcounter{equation}{14}
\setcounter{table}{4}
\setcounter{figure}{2}

\par
In this supplementary document, we first provide details on the derivation of the \textit{Scaled Symmetric Dirichlet Energy}, 
and how we improve its numerical stability in practice (Sec.~\ref{sec:ssde}). Next, we present implementation details of the proposed surface representation and baselines used in the experiments (Sec.~\ref{sec:details}). Lastly, we provide additional quantitative and qualitative results on the surface reconstruction experiment (Sec.~\ref{sec:results}).

\section{Scaled Symmetric Dirichlet Energy}
\label{sec:ssde}

\subsection{Detailed Derivation}
\label{subsec:ssde:derivation}

\par
When the surface parameterization $\varphi_{\theta_k}$ of each chart preserves the metric of the parametric domain up to a specific common scale of $L$, the two singular values of its Jacobian $J_k$, $\sigma_{k, 1}$ and $\sigma_{k, 2}$ 
are equal to $L$ at every point $\bm{u}$ in the parametric domain, \ie: 
\begin{equation}
	\sigma_{k, 1} (\bm{u}) = \sigma_{k, 2} (\bm{u}) = L \ .
\end{equation}
By the definition of the metric tensor $g_k$, its two singular values and eigenvalues 
are also equal to $L^2$ everywhere.

\par
Consequently, it is clear that the \textit{Symmetric Dirichlet Energy} (SDE) \cite{schreiner2004_sde_real2,smith2015_sde_real,rabinovich_sde} given by:
\begin{equation}
\begin{split}
  &\ \frac{1}{\sum_{k \in \mathcal{K}} \abs{\mathcal{W}_k}} \sum_{k \in \mathcal{K}} \sum_{\bm{u} \in \mathcal{W}_k} \sigma_{k, 1} (\bm{u})^2 + \sigma_{k, 2} (\bm{u})^2 + \frac{1}{\sigma_{k, 1} (\bm{u})^2} + \frac{1}{\sigma_{k, 2} (\bm{u})^2} \\
  = &\ \frac{1}{\sum_{k \in \mathcal{K}} \abs{\mathcal{W}_k}} \sum_{k \in \mathcal{K}} \sum_{\bm{u} \in \mathcal{W}_k}  \mathrm{trace} (g_k (\bm{u})) + \mathrm{trace} (g_k (\bm{u})^{-1}) \\
  = &\ \mathrm{mean}_\mathcal{W} (\mathrm{trace} \circ g_k) + \mathrm{mean}_\mathcal{W} (\mathrm{trace} \circ g_k^{-1})
\end{split}
\end{equation}
quantifies the isometric distortion, since a global minimum value of 4 is attained if and only if both $\sigma_{k, 1}$ and $\sigma_{k, 2}$ equal to $1$ everywhere. The former and latter terms of the SDE correspond to the \textit{Dirichlet} energy of $\varphi_{\theta_k}$ and $\varphi_{\theta_k}^{-1}$, respectively. 

\par
The proposed \textit{Scaled Symmetric Dirichlet Energy} (SSDE) generalizes the SDE to quantify metric distortion up to a specific common scale of $L$. This is simply done by incorporating a scale factor of $L$ as follows:
\begin{equation}
\begin{split}
  &\ \frac{1}{\sum_{k \in \mathcal{K}} \abs{\mathcal{W}_k}} \sum_{k \in \mathcal{K}} \sum_{\bm{u} \in \mathcal{W}_k} \frac{\sigma_{k, 1} (\bm{u})^2}{L^2} + \frac{\sigma_{k, 2} (\bm{u})^2}{L^2} + \frac{L^2}{\sigma_{k, 1} (\bm{u})^2} + \frac{L^2}{\sigma_{k, 2} (\bm{u})^2} \\
  = &\ \frac{1}{\sum_{k \in \mathcal{K}} \abs{\mathcal{W}_k}} \sum_{k \in \mathcal{K}} \sum_{\bm{u} \in \mathcal{W}_k}  \frac{1}{L^2} \mathrm{trace} (g_k (\bm{u})) + L^2 \mathrm{trace} (g_k (\bm{u})^{-1}) \\
  = &\ \frac{1}{L^2} \mathrm{mean}_\mathcal{W} (\mathrm{trace} \circ g_k) + L^2 \mathrm{mean}_\mathcal{W} (\mathrm{trace} \circ g_k^{-1}) \ ,
  \label{eq:ssde}
\end{split}
\end{equation}
such that a global minimum value of 4 is attained if and only if both $\sigma_{k, 1}$ and $\sigma_{k, 2}$ equal to $L$ everywhere.

\par
Furthermore, the SSDE can be employed to quantify metric distortion up to an arbitrary common scale by finding an optimal scale $L^*$ that minimizes it. Since SSDE is a convex function of $L$, $L^*$ is simply given by the critical point:
\begin{equation}
	\frac{\partial \ \mathrm{SSDE}}{\partial L} \bigg\rvert_{L=L^*} = 0 \ ,
\end{equation}
which evaluates to:
\begin{equation}
	{L^*}^2 = \sqrt{\frac{\mathrm{mean}_\mathcal{W} (\mathrm{trace} \circ g_k)}{\mathrm{mean}_\mathcal{W} (\mathrm{trace} \circ g_k^{-1})}} \ .
    \label{eq:L*2}
\end{equation}
By substituting Eq.~\ref{eq:L*2} into Eq.~\ref{eq:ssde} with $L = L^*$, we can simplify the SSDE at the optimal scale $L^*$ as:
\begin{equation}
	2 \sqrt{ \mathrm{mean}_{\mathcal{W}} (\mathrm{trace} \circ g_k) \ \mathrm{mean}_{\mathcal{W}} (\mathrm{trace} \circ g_k^{-1}) } \ .
\end{equation}

\subsection{Improving Numerical Stability}
\label{subsec:ssde:stability}

\par
In general, the SSDE at the optimal scale $L^*$ is numerically stable since it is given by the geometric mean of the Dirichlet energies of $\varphi_{\theta_k}$ and $\varphi_{\theta_k}^{-1}$, which are roughly inversely proportional to each other. Although it is rare in practice, the existence of a singular $g_k$ leads to instability in the Dirichlet energy of $\varphi_{\theta_k}^{-1}$, and hence the SSDE at the optimal scale $L^*$ as well as the SSDE in general.

\par
To improve numerical stability of the SSDE under such a scenario, we augment it with a small positive $\epsilon$ value as follows:
\begin{equation}
\begin{split}
  &\ \frac{1}{\sum_{k \in \mathcal{K}} \abs{\mathcal{W}_k}} \sum_{k \in \mathcal{K}} \sum_{\bm{u} \in \mathcal{W}_k} \frac{\sigma_{k, 1} (\bm{u})^2 + \epsilon}{L^2  + \epsilon} + \frac{\sigma_{k, 2} (\bm{u})^2 + \epsilon}{L^2 + \epsilon} + \frac{L^2 + \epsilon}{\sigma_{k, 1} (\bm{u})^2 + \epsilon} + \frac{L^2 + \epsilon}{\sigma_{k, 2} (\bm{u})^2 + \epsilon} \\
  = &\ \frac{1}{\sum_{k \in \mathcal{K}} \abs{\mathcal{W}_k}} \sum_{k \in \mathcal{K}} \sum_{\bm{u} \in \mathcal{W}_k}  \frac{1}{L^2 + \epsilon} \mathrm{trace} (g_k (\bm{u}) + \epsilon I) + (L^2 + \epsilon) \ \mathrm{trace} ((g_k (\bm{u}) + \epsilon I)^{-1}) \\
  = &\ \frac{1}{L^2 + \epsilon} \mathrm{mean}_\mathcal{W} (\mathrm{trace} \circ \hat g_k) + (L^2 + \epsilon) \ \mathrm{mean}_\mathcal{W} (\mathrm{trace} \circ \hat g_k^{-1}) \ ,
\end{split}
\end{equation}
where:
\begin{equation}
	\hat g_k (\bm{u}) = g_k (\bm{u}) + \epsilon I
\end{equation}
is the \textit{$\epsilon$-conditioned} metric tensor with singular values and eigenvalues $\sigma_{k, i} (\bm{u})^2 + \epsilon$. This introduces a lower and upper bound on the $\epsilon$-conditioned Dirichlet energy of $\varphi_{\theta_k}$ and $\varphi_{\theta_k}^{-1}$, respectively, given by:
\begin{equation}
	\mathrm{mean}_\mathcal{W} (\mathrm{trace} \circ \hat g_k) \geq 2 \epsilon \ , \qquad
    \mathrm{mean}_\mathcal{W} (\mathrm{trace} \circ \hat g_k^{-1}) \leq \frac{2}{\epsilon} \ .
\end{equation}

\par
The $\epsilon$-conditioned SSDE at $L$ preserves the global minimum value of 4 when both $\sigma_{k, 1}$ and $\sigma_{k, 2}$ equal to $L$ everywhere. Following the exact same steps in Sec.~\ref{subsec:ssde:derivation}, it can also be shown that the $\epsilon$-conditioned optimal scale $\hat L^*$ is given by:
\begin{equation}
  {\hat L^*}{}^2 = \sqrt{\frac{\mathrm{mean}_\mathcal{W} (\mathrm{trace} \circ \hat g_k)}{\mathrm{mean}_\mathcal{W} (\mathrm{trace} \circ \hat g_k^{-1})}} - \epsilon \geq 0
\end{equation}
and the $\epsilon$-conditioned SSDE at $\hat L^*$ is similarly given by:
\begin{equation}
	2 \sqrt{ \mathrm{mean}_{\mathcal{W}} (\mathrm{trace} \circ \hat g_k) \ \mathrm{mean}_{\mathcal{W}} (\mathrm{trace} \circ \hat g_k^{-1}) } \ .
\end{equation}
In practice, we find $\epsilon=1 \times 10^{-4}$ to be sufficient for quantifying metric distortion up to an arbitrary common scale with the $\epsilon$-conditioned SSDE at $\hat L^*$.

\section{Implementation Details}
\label{sec:details}

\subsection{Minimal Neural Atlas}
\label{subsec:details:mna}

\subsubsection{Architecture.}

\par
For all experiments, we employ a minimal neural atlas conditioned on an 1024-dimensional latent code $\bm{z} \in \mathbb{R}^{1024}$ to reconstruct the family of surfaces described in the CLOTH3D++ and ShapeNet datasets. The encoder architecture adopted for extracting the latent code of a surface depends on the task, or more precisely the form of input. For surface reconstruction where the input is a point cloud, we employ PointNet \cite{charles2017_pointnet} with all \textit{Batch Normalization} \cite{ioffe2015_bn} layers removed for better training stability and convergence. On the other hand, we adopt an ImageNet\cite{deng2009_imagenet}-pretrained ResNet-18 \cite{kaiming2016_resnet} for single-view reconstruction, where the input is an image.

\par
In contrast to the encoder, we employ the same architecture for the conditional minimal neural atlas in all experiments. The architectures adopted for the conditional $\varphi_{\theta_k}$ and $\tilde l_{\theta_k}$ of each chart $k$ are almost identical and are heavily based on the architecture of the IDR neural scene representation \cite{yariv2020_idr}, which in turn is based on the DeepSDF \cite{park2019_deepsdf} implicit neural surface representation.

\par
Particularly, the conditional $\varphi_{\theta_k}$ maps the concatenated inputs of $\bm{z}$ and $\bm{u}$ to a 3D point on the reconstructed surface via a 4-layer Multi-Layer Perceptron (MLP), where each layer comprises 512 hidden units applied with \textit{Weight Normalization} \cite{salimans2016_weight_norm}. 
Each hidden layer except for the last is followed by a \textit{SoftPlus} activation with hyperparameter $\beta=100$. 
In contrast to \textit{ReLU}, SoftPlus is infinitely-differentiable everywhere. This enables the computation of differentiable geometric properties \cite{bednarik2020_dsp} and facilitates slightly lower distortion, as observed empirically. The inputs are also concatenated with the activations of the second hidden layer to form the next hidden layer inputs. In terms of the model size, this architecture is comparable to that of AtlasNet and DSP, but more lightweight than that of TearingNet.

\par
The conditional $\tilde l_{\theta_k}$ adopts the same architecture as the conditional $\varphi_{\theta_k}$, except for some subtle differences. Specifically, a positional encoding with 6 octaves is applied on maximal point coordinates $\tilde{\bm{x}}$ before being concatenated with $\bm{z}$ to form the input of the network. As shown in the ablation studies, this is important for $\tilde l_{\theta_k}$ to capture high frequency details for more accurate reconstructions. Moreover, ReLU and sigmoid are also used for the intermediate and output activations respectively.


\subsubsection{Training.}

\par
The training loss weights used in all experiments are given by $\lambda_{rec}=1.0, \lambda_{occ}=1.0$ and $\lambda_{dist}=0.00001$. Apart from the encoder, note that
training with equal importance on $\mathcal{L}_{rec}$ and $\mathcal{L}_{occ}$ 
works because they solely supervise $\varphi_{\theta_k}$ and $\tilde l_{\theta_k}$, respectively. For surface reconstruction, we adopt the exact same training procedure as AtlasNet. In particular, we adopt the Adam optimizer \cite{kingma2015_adam} with a learning rate of $0.001$ and \textit{PyTorch}\cite{paszke2019_pytorch}-default hyperparameters. The network is trained for 150 epochs with a learning rate decay of $0.1$ at 120, 140 and 145 epochs. The same training procedure is also employed for single-view reconstruction, except that we also adopt a surface reconstruction-pretrained conditional $\varphi_{\theta_k}$ and $\tilde l_{\theta_k}$.

\subsubsection{Inference.}

\par
For the evaluation of all experiments, the label frequency $c$ is estimated with a minimum interior rate $\eta=40\%$. We also adopt the default occupancy probability threshold $\tau=0.5$ to define the parametric domain of each chart. Furthermore, a reconstructed surface point cloud with approximately 25,000 points is extracted with the proposed two-step batch rejection sampling strategy using an initial UV sample size of 16,667 (\ie \nicefrac{2}{3} of 25,000).

\subsection{Baselines}
\label{subsec:details:baselines}

\par
To provide a fair comparison, we train all baselines on the exact same datasets using their official implementations. The AtlasNet training procedure is used for AtlasNet++ and DSP 
in all experiments. In contrary to all other works, we train DSP without the overlap loss since it requires access to target surface areas. Furthermore, a relatively larger epsilon value of $0.01$ is added to the denominator of the deformation loss to significantly improve its numerical stability. Similar to \cite{gupta2020_nmf}, we train AtlasNet++ with Point Cloud CD, Mesh CD and SSDE at the optimal scale weighted by $1.0, 1.0$ and $0.00001$, respectively.

\par
TearingNet is trained according to its two-step strategy in both experiments, which requires approximately 6 to 7 times the number of epochs compared to other works. Moreover, we omit the optional graph filter as it unnecessarily constrains the UV sampling density during inference to that of training. This is attributed to its dependence on tearing or graph weight hyperparameters $\epsilon$ and $r$ on the UV sampling density. 
For single-view reconstruction with TearingNet, we adopt the same ResNet-18 encoder 
and surface reconstruction-pretrained F-Net in the first step of the training. The optimal graph weight hyperparameter $\epsilon$, which defines the parametric domain of the chart and hence the topology of the reconstructed surface, is tuned with respect to the Mesh CD and F-score @ 1\% on the validation split of ShapeNet since it contains a wide range of surfaces with complex topologies. 
Consequently, $\epsilon=0.025$ is used throughout the experiments for evaluation.

\section{Additional Results}
\label{sec:results}

\par
In this section, we first present a qualitative analysis of distortion (Sec.~\ref{subsec:results:distortion}) and visualizations of SCAR violation artifacts (Sec.~\ref{subsec:results:artifacts}) before looking into the occupancy rates of minimal neural atlas (Sec.~\ref{subsec:results:occupancy}). Next, we investigate the effect of training with different UV sample sizes (Sec.~\ref{subsec:results:sample_size}). Lastly, we examine the sensitivity of our representation on hyperparameters such as the number of positional encoding octaves in $\tilde l_{\theta_k}$ and minimum interior rate $\eta$ (Sec.~\ref{subsec:results:sensitivity}). Unless otherwise stated, results involving only minimal neural atlas are obtained using 3 charts with metric distortion loss from the surface reconstruction experiment on ShapeNet.

\subsection{Qualitative Distortion Analysis}
\label{subsec:results:distortion}

\begin{figure}[tp]
\centering 
		\begin{subfigure}{1\linewidth}
        	\centering
            \rotatebox[origin=c]{90}{AtlasNet}
            \raisebox{-0.5\height}{\includegraphics[width=0.199\linewidth,trim={4cm 3cm 4cm 3cm},clip]{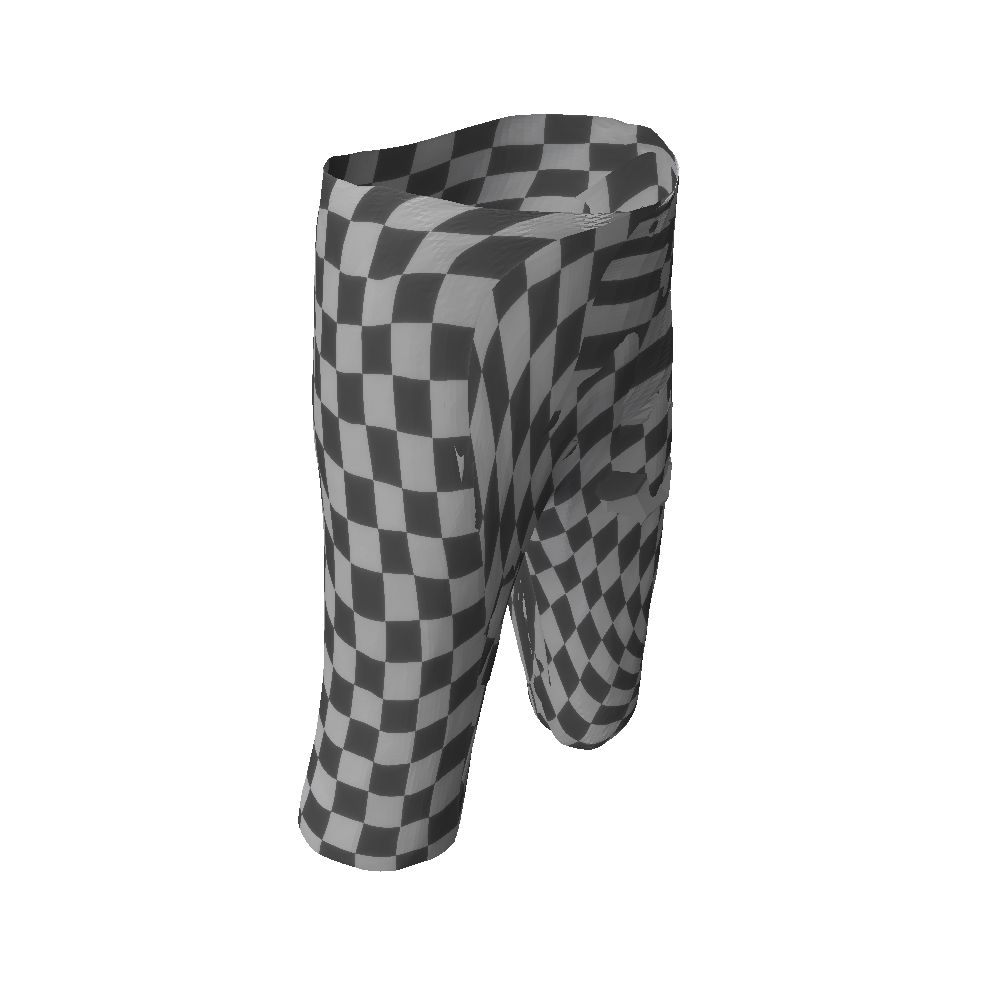}}
            \hfil
			\raisebox{-0.5\height}{\includegraphics[width=0.199\linewidth,trim={4cm 4cm 4cm 7cm},clip]{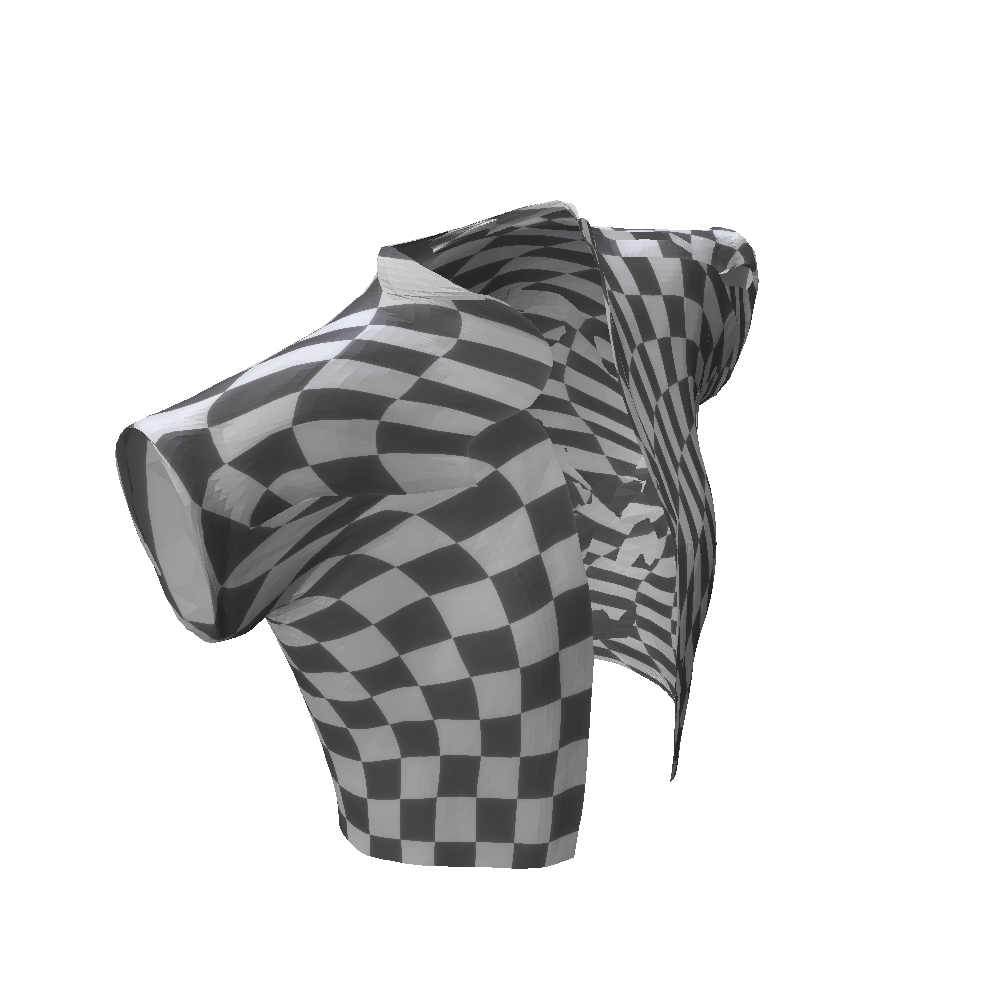}}
            \hfil
            \raisebox{-0.5\height}{\includegraphics[width=0.199\linewidth,trim={4cm 4cm 4cm 4cm},clip]{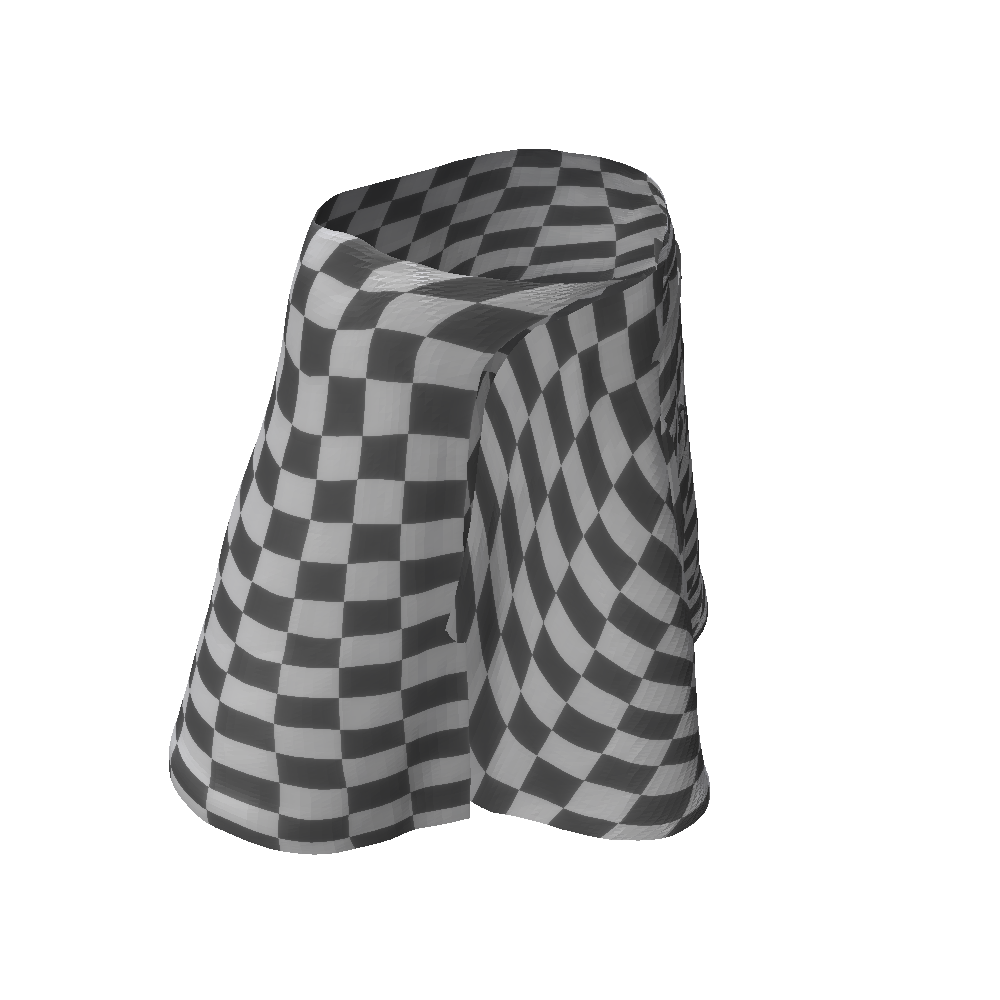}}
		\end{subfigure}
        \begin{subfigure}{1\linewidth}
        	\centering
            \rotatebox[origin=c]{90}{AtlasNet++}
            \raisebox{-0.5\height}{\includegraphics[width=0.199\linewidth,trim={4cm 3cm 4cm 3cm},clip]{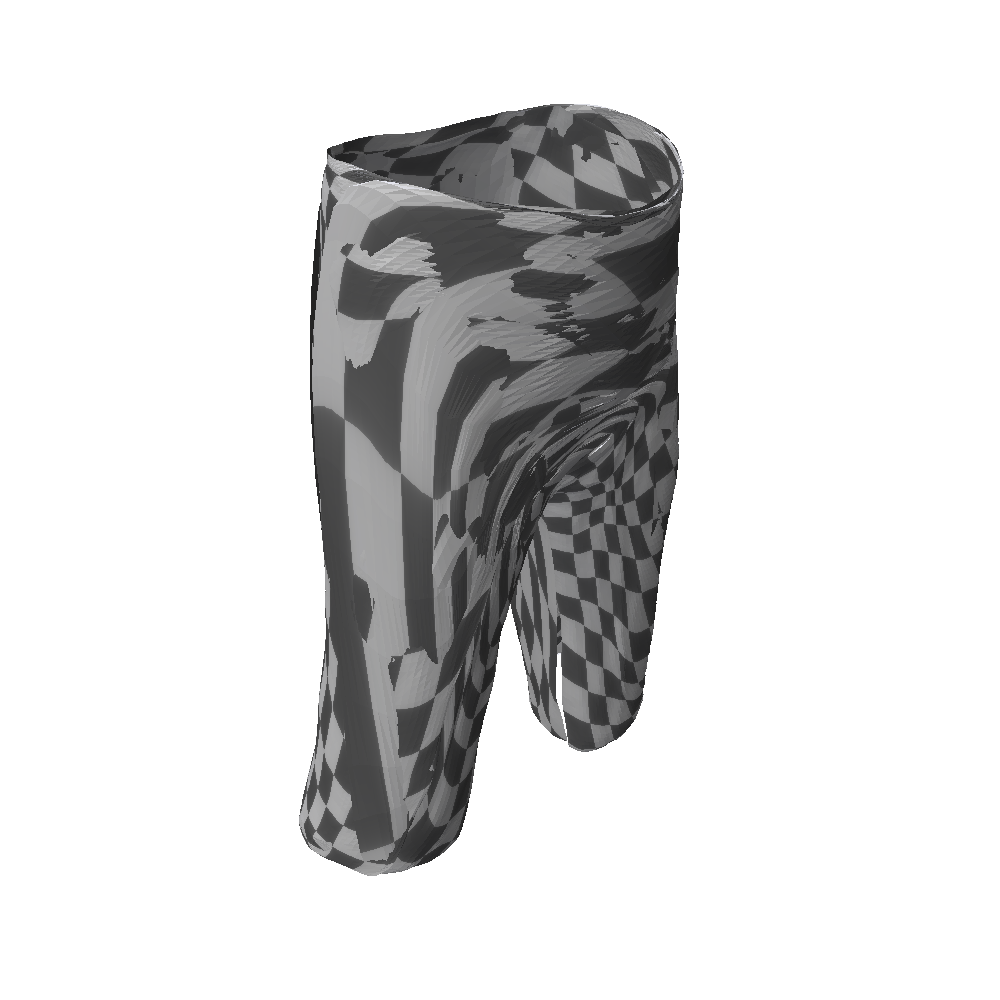}}
            \hfil
			\raisebox{-0.5\height}{\includegraphics[width=0.199\linewidth,trim={4cm 4cm 4cm 7cm},clip]{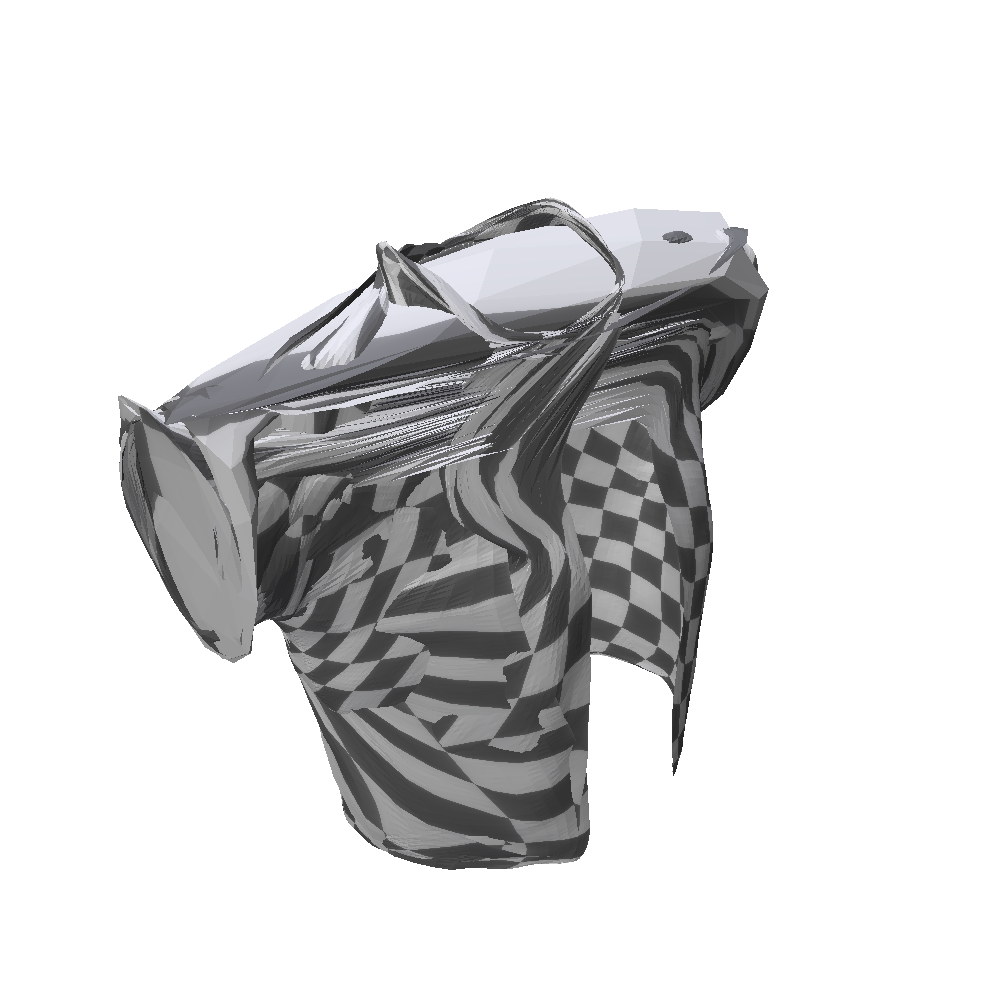}}
            \hfil
            \raisebox{-0.5\height}{\includegraphics[width=0.199\linewidth,trim={4cm 4cm 4cm 4cm},clip]{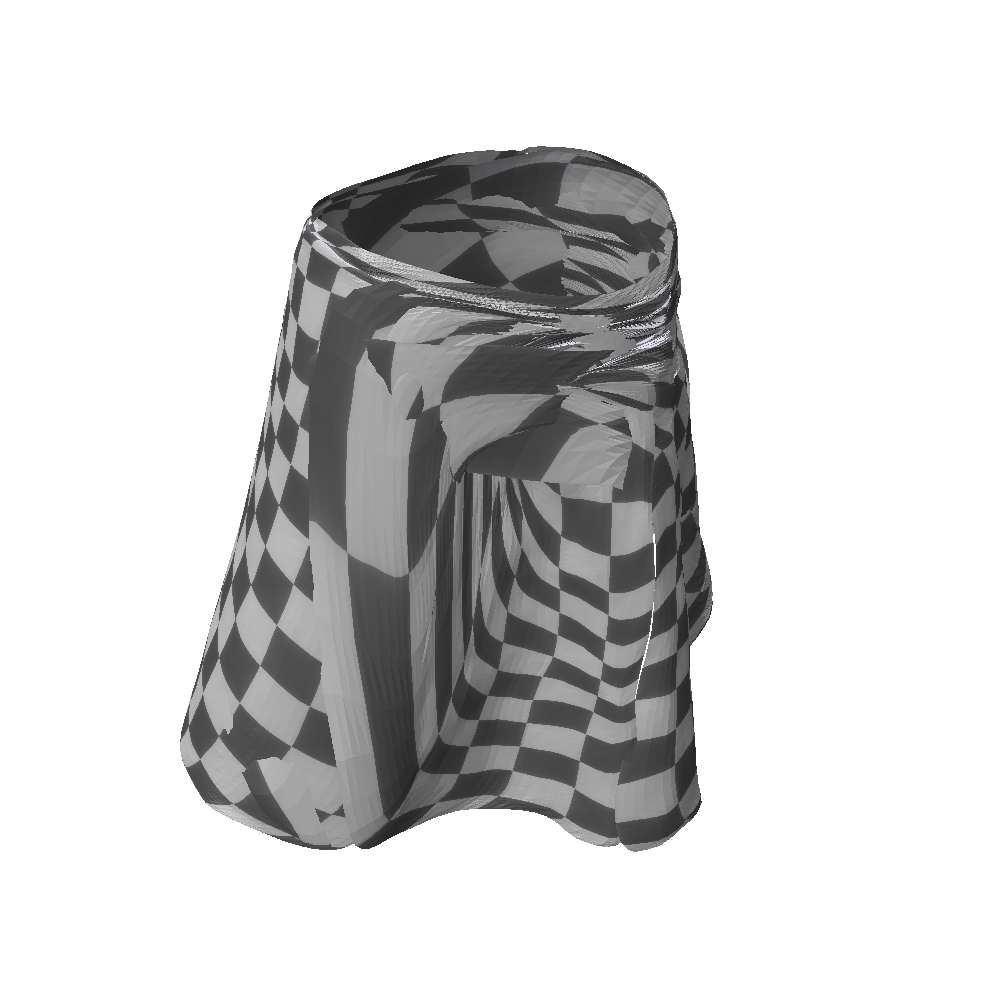}}
		\end{subfigure}
        \begin{subfigure}{1\linewidth}
        	\centering
            \rotatebox[origin=c]{90}{DSP}
            \raisebox{-0.5\height}{\includegraphics[width=0.199\linewidth,trim={4cm 3cm 4cm 3cm},clip]{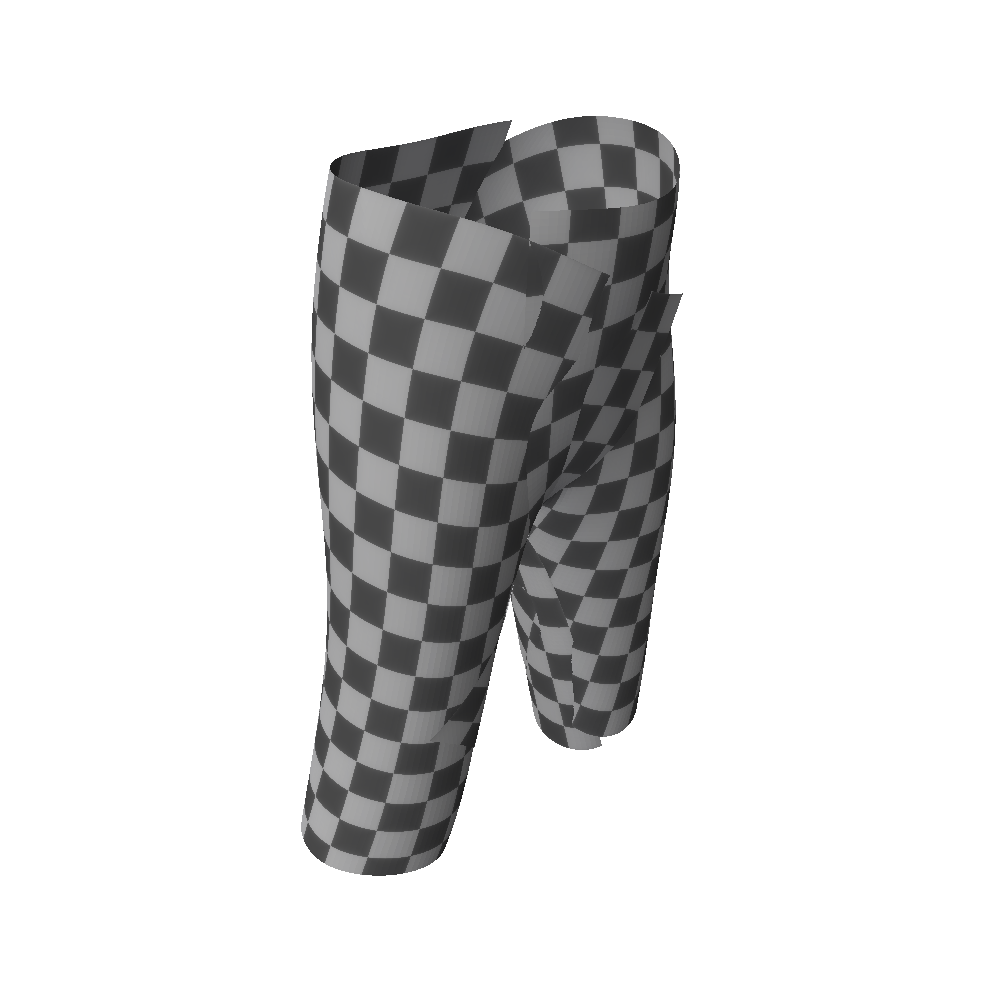}}
            \hfil
			\raisebox{-0.5\height}{\includegraphics[width=0.199\linewidth,trim={4cm 4cm 4cm 7cm},clip]{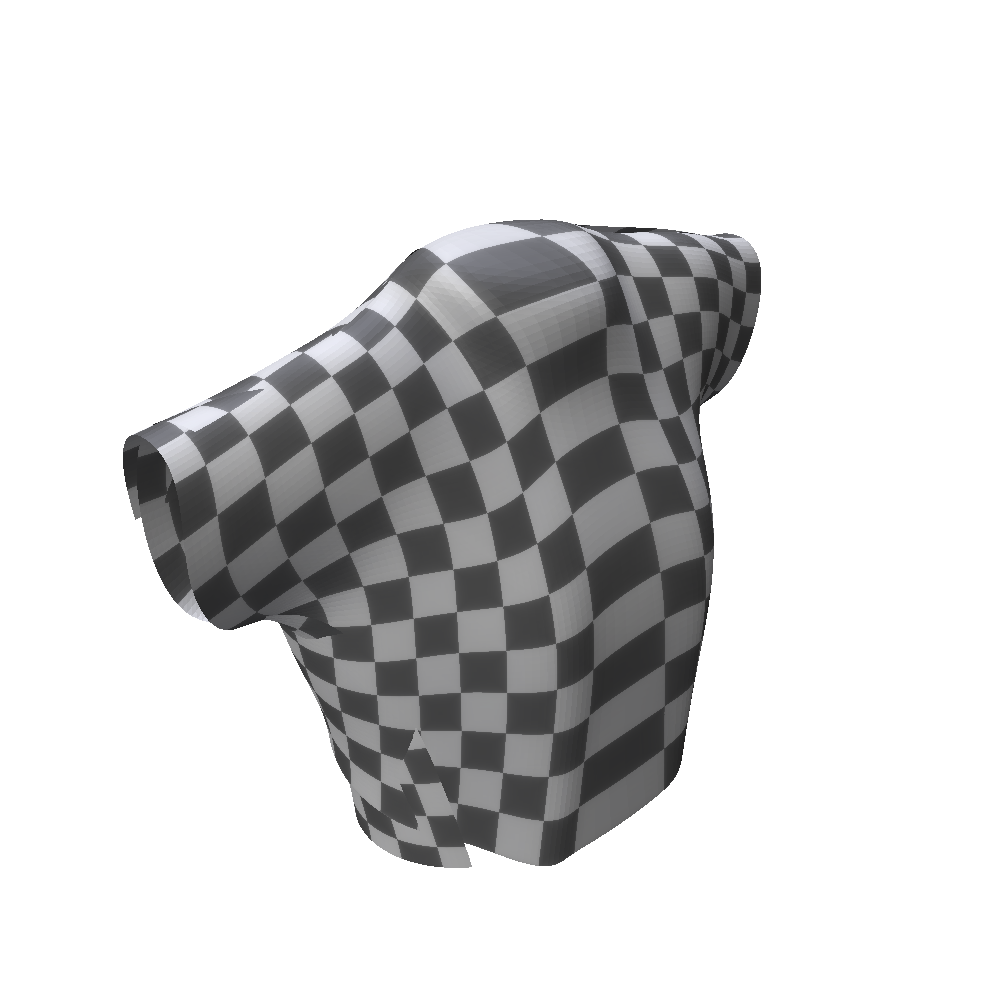}}
            \hfil
            \raisebox{-0.5\height}{\includegraphics[width=0.199\linewidth,trim={4cm 4cm 4cm 4cm},clip]{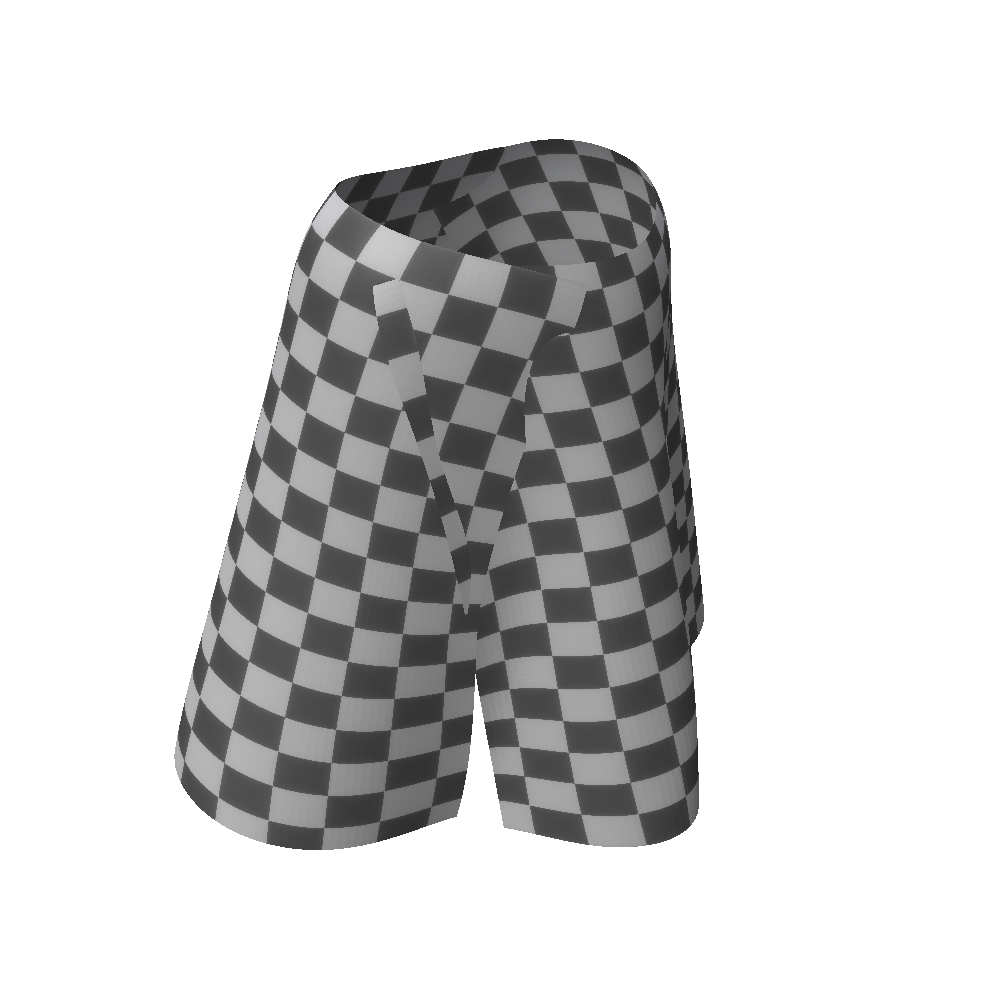}}
		\end{subfigure}
        \begin{subfigure}{1\linewidth}
        	\centering
            \rotatebox[origin=c]{90}{TearingNet}
            \raisebox{-0.5\height}{\includegraphics[width=0.199\linewidth,trim={4cm 3cm 4cm 3cm},clip]{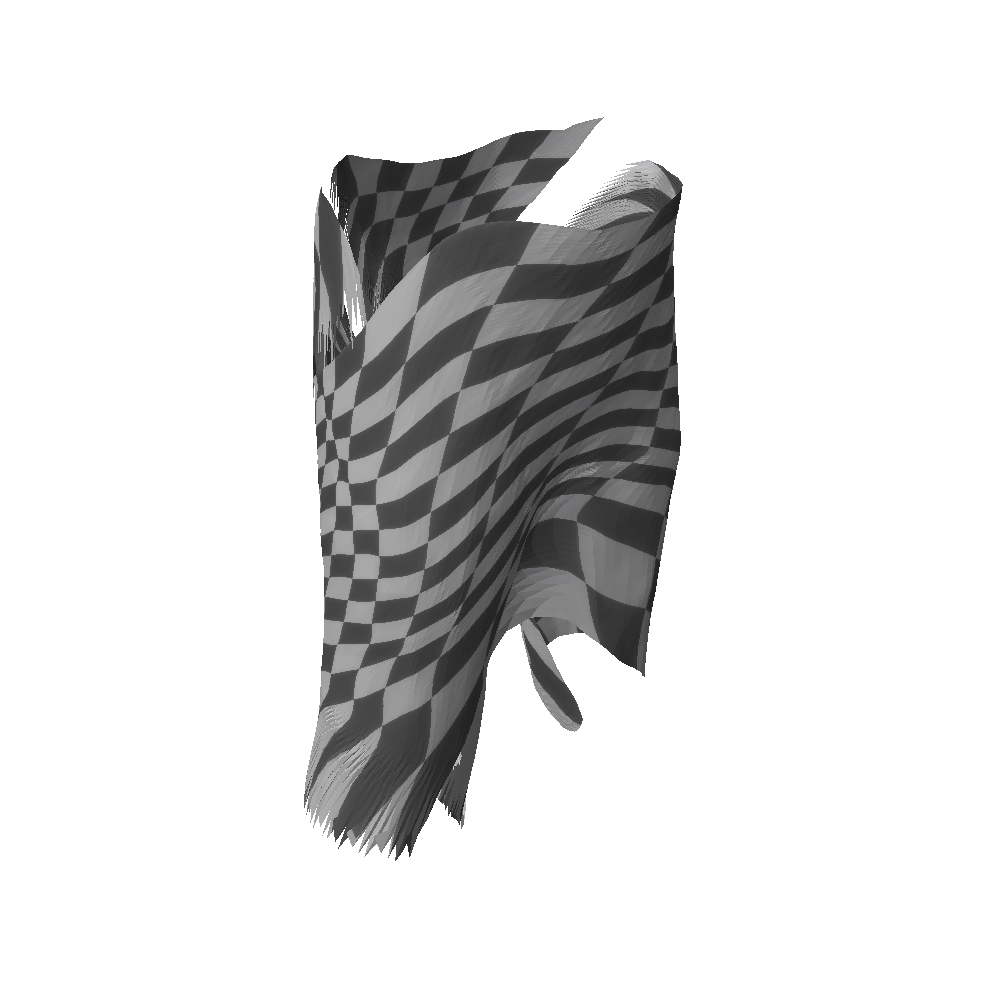}}
            \hfil
			\raisebox{-0.5\height}{\includegraphics[width=0.199\linewidth,trim={4cm 4cm 4cm 7cm},clip]{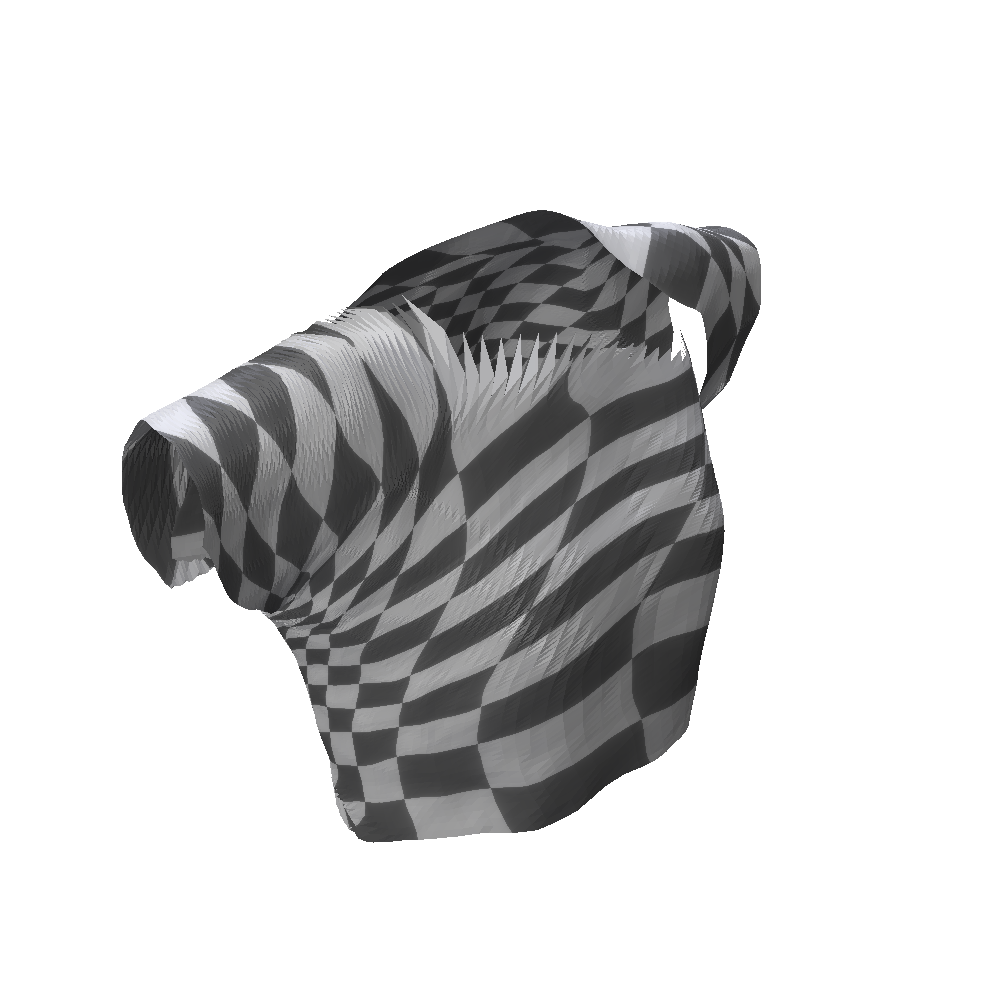}}
            \hfil
            \raisebox{-0.5\height}{\includegraphics[width=0.199\linewidth,trim={4cm 4cm 4cm 4cm},clip]{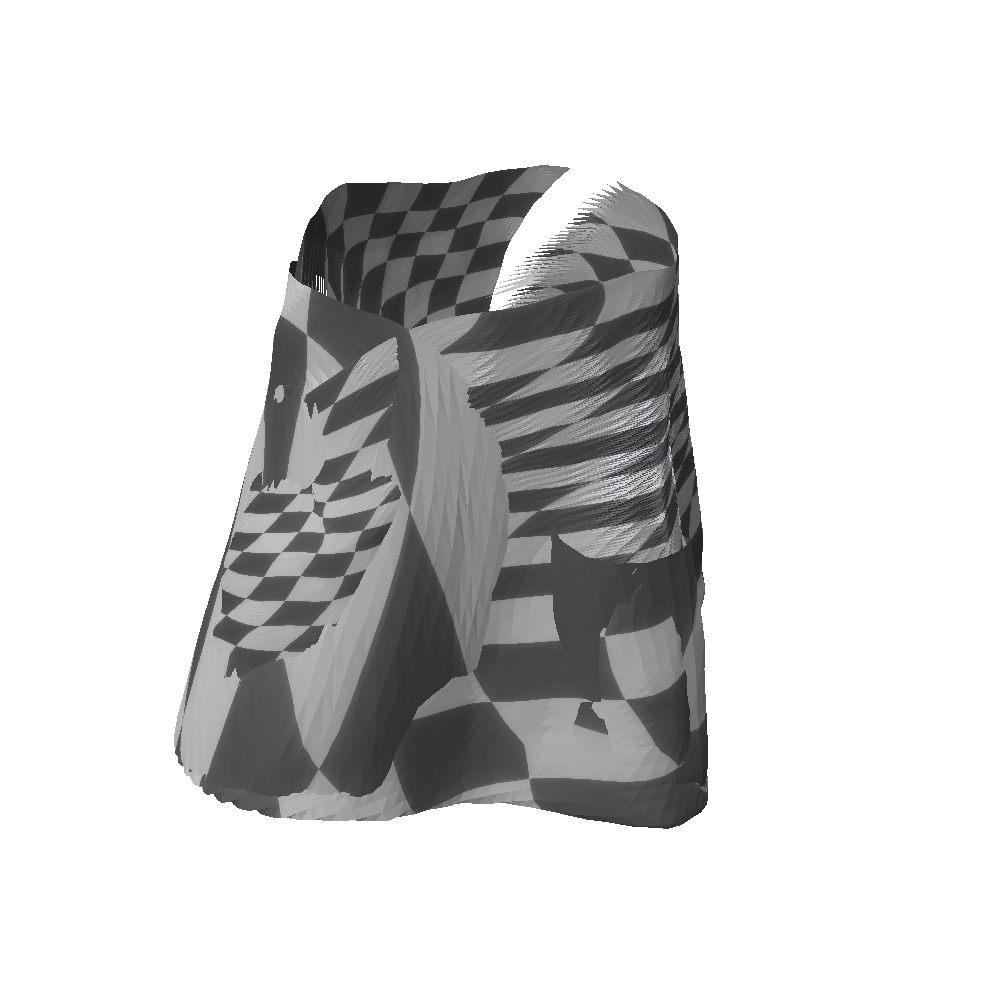}}
		\end{subfigure}
        \begin{subfigure}{1\linewidth}
        	\centering
            \rotatebox[origin=c]{90}{Ours}
            \raisebox{-0.5\height}{\includegraphics[width=0.199\linewidth,trim={4cm 3cm 4cm 3cm},clip]{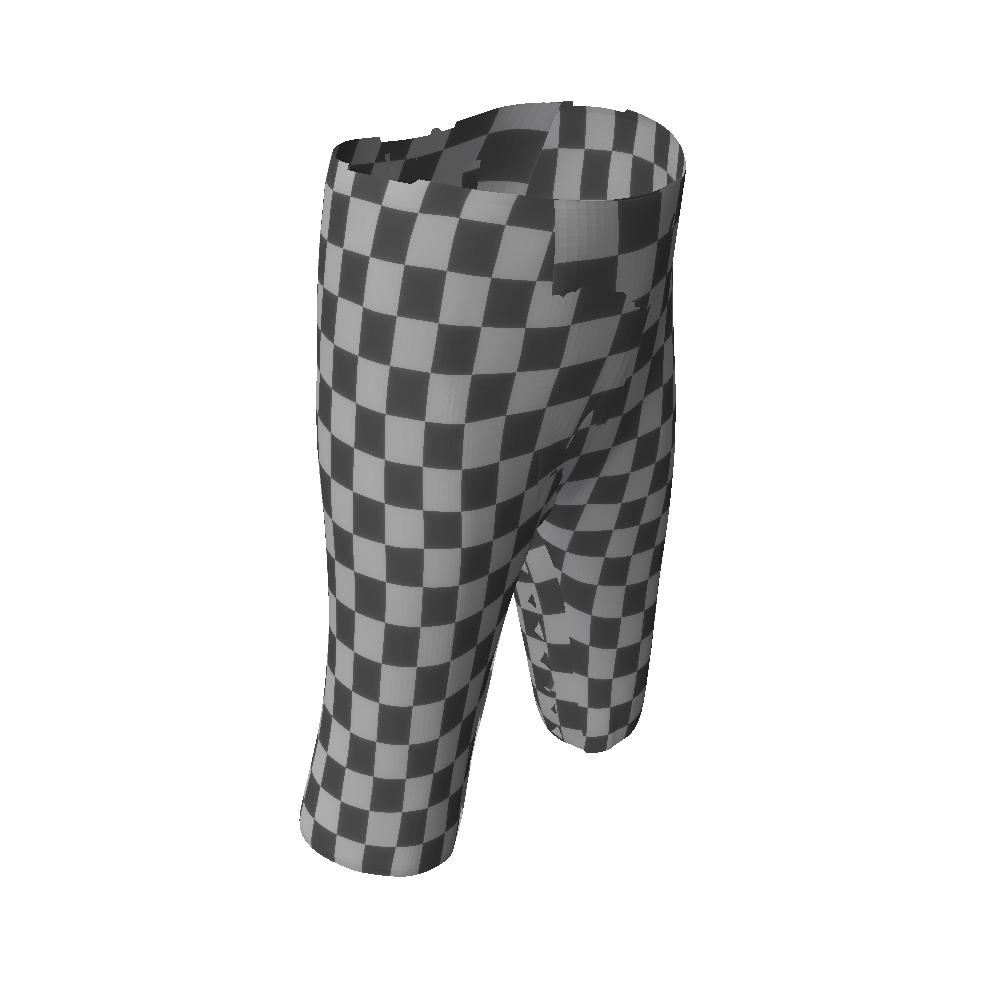}}
            \hfil
			\raisebox{-0.5\height}{\includegraphics[width=0.199\linewidth,trim={4cm 4cm 4cm 7cm},clip]{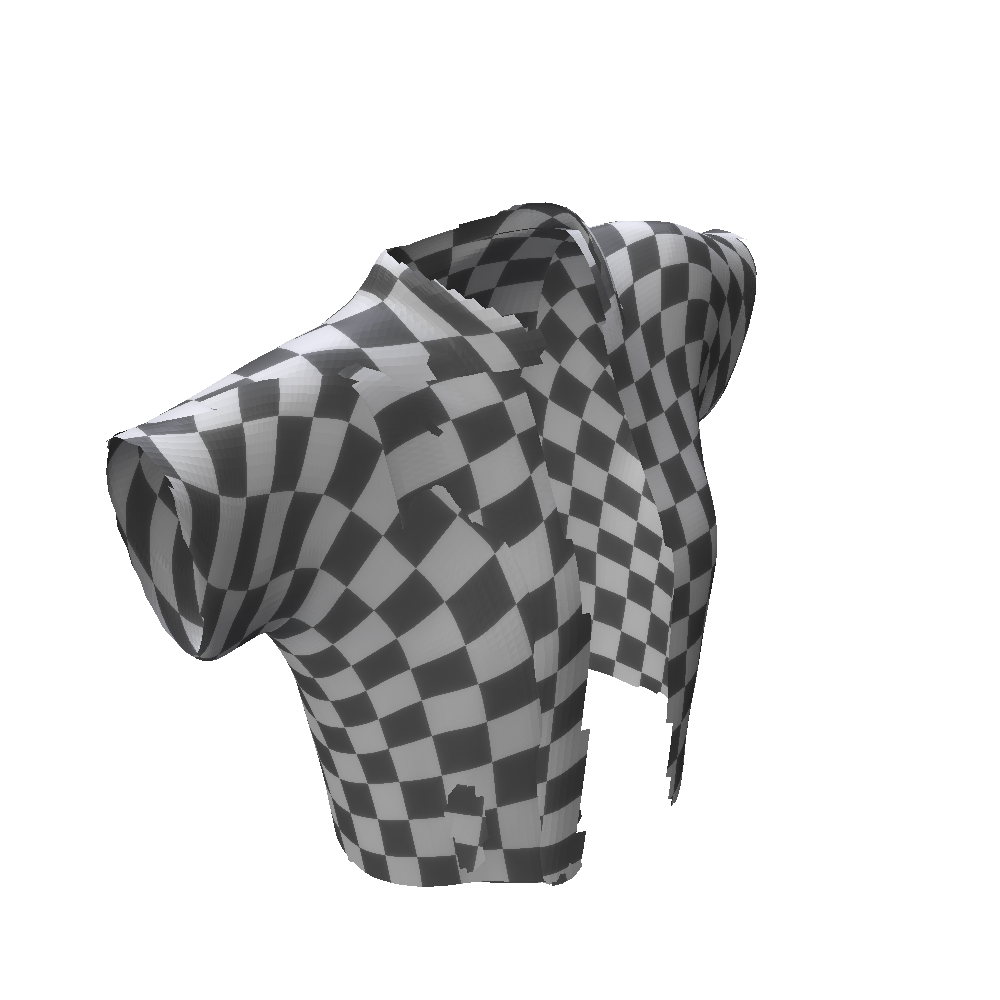}}
            \hfil
            \raisebox{-0.5\height}{\includegraphics[width=0.199\linewidth,trim={4cm 4cm 4cm 4cm},clip]{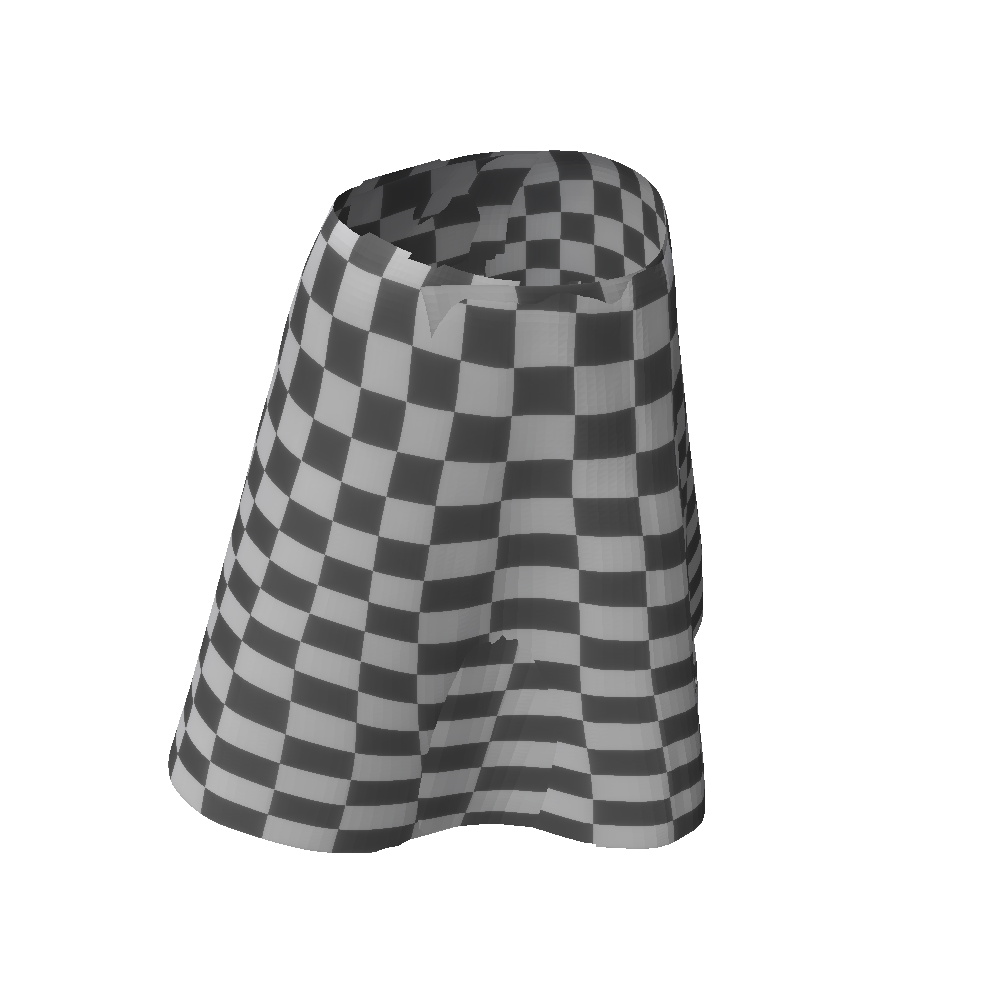}}
		\end{subfigure}
        \begin{subfigure}{1\linewidth}
        	\centering
            \rotatebox[origin=c]{90}{Target}
            \raisebox{-0.5\height}{\includegraphics[width=0.199\linewidth,trim={4cm 3cm 4cm 3cm},clip]{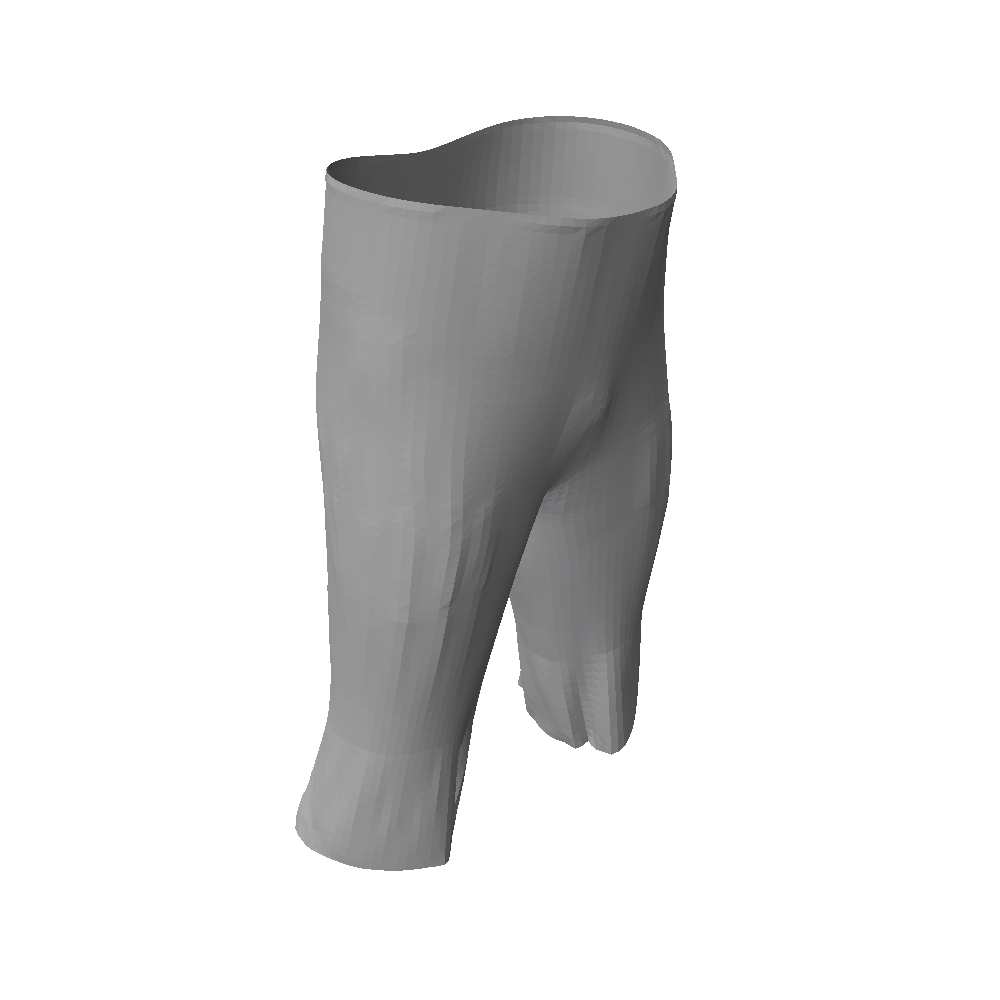}}
            \hfil
			\raisebox{-0.5\height}{\includegraphics[width=0.199\linewidth,trim={4cm 4cm 4cm 7cm},clip]{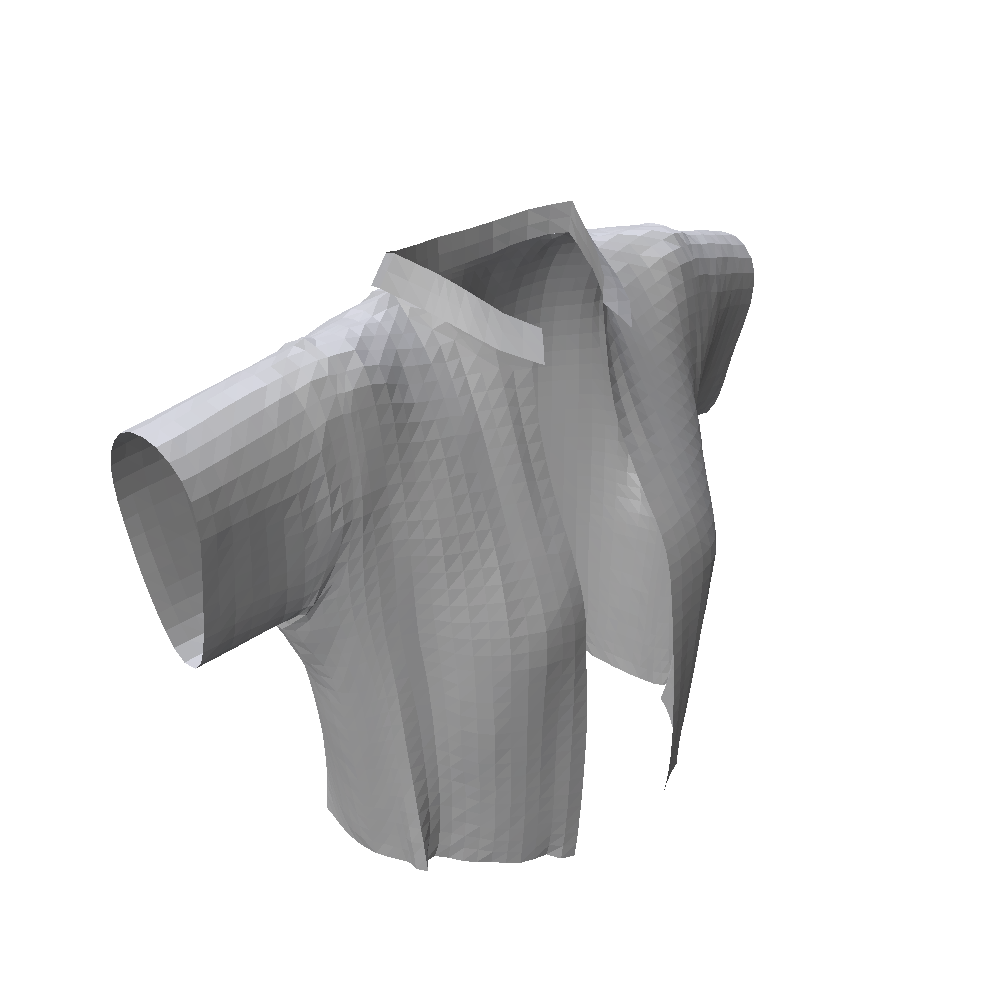}}
            \hfil
            \raisebox{-0.5\height}{\includegraphics[width=0.199\linewidth,trim={4cm 4cm 4cm 4cm},clip]{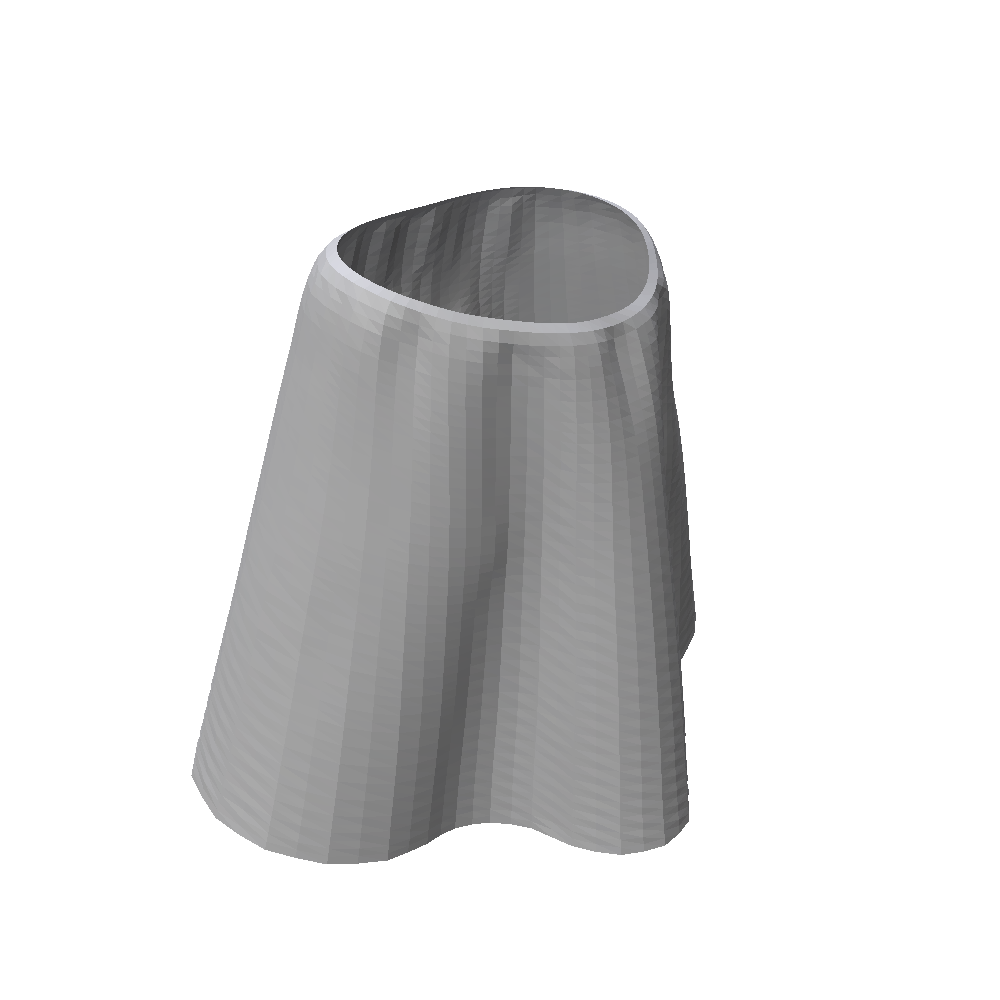}}
		\end{subfigure}
        \begin{subfigure}{1\linewidth}
        	\centering
            \rotatebox[origin=c]{90}{Input}
            \raisebox{-0.5\height}{\includegraphics[width=0.199\linewidth,trim={4cm 3cm 4cm 3cm},clip]{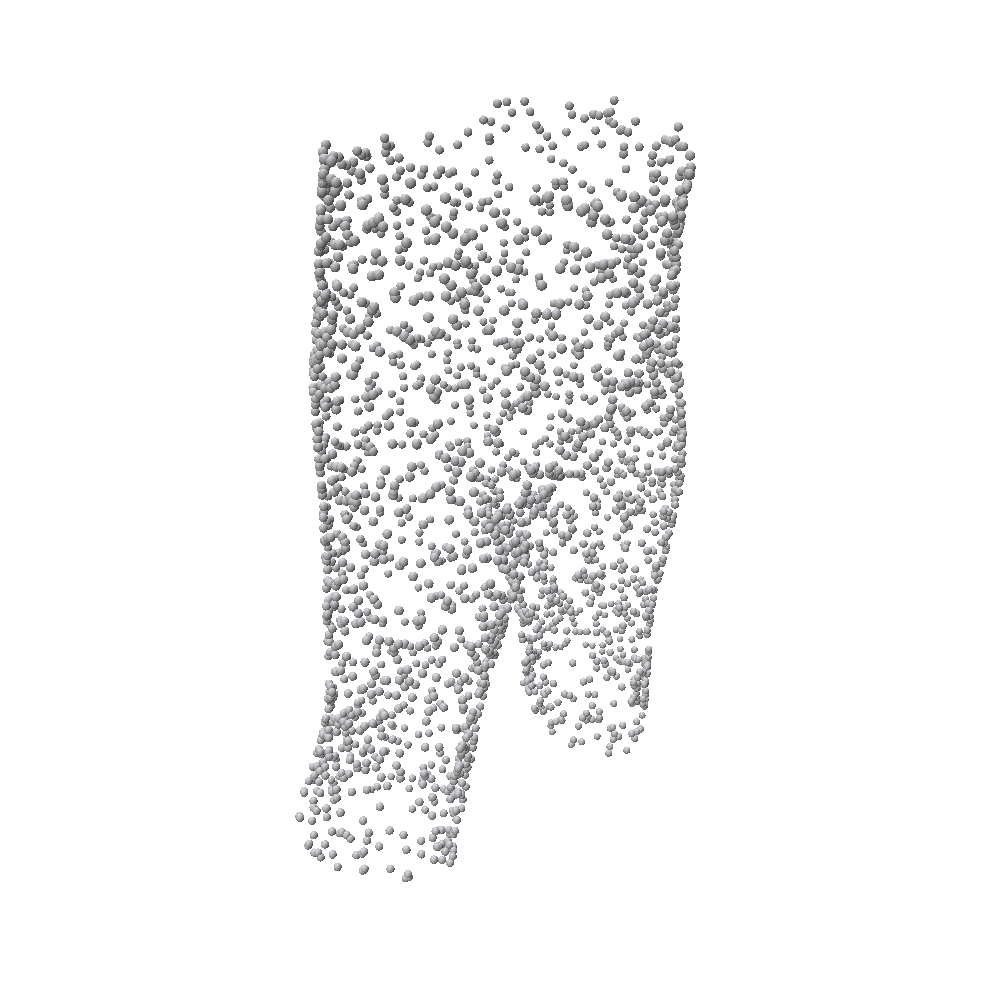}}
            \hfil
			\raisebox{-0.5\height}{\includegraphics[width=0.199\linewidth,trim={4cm 4cm 4cm 7cm},clip]{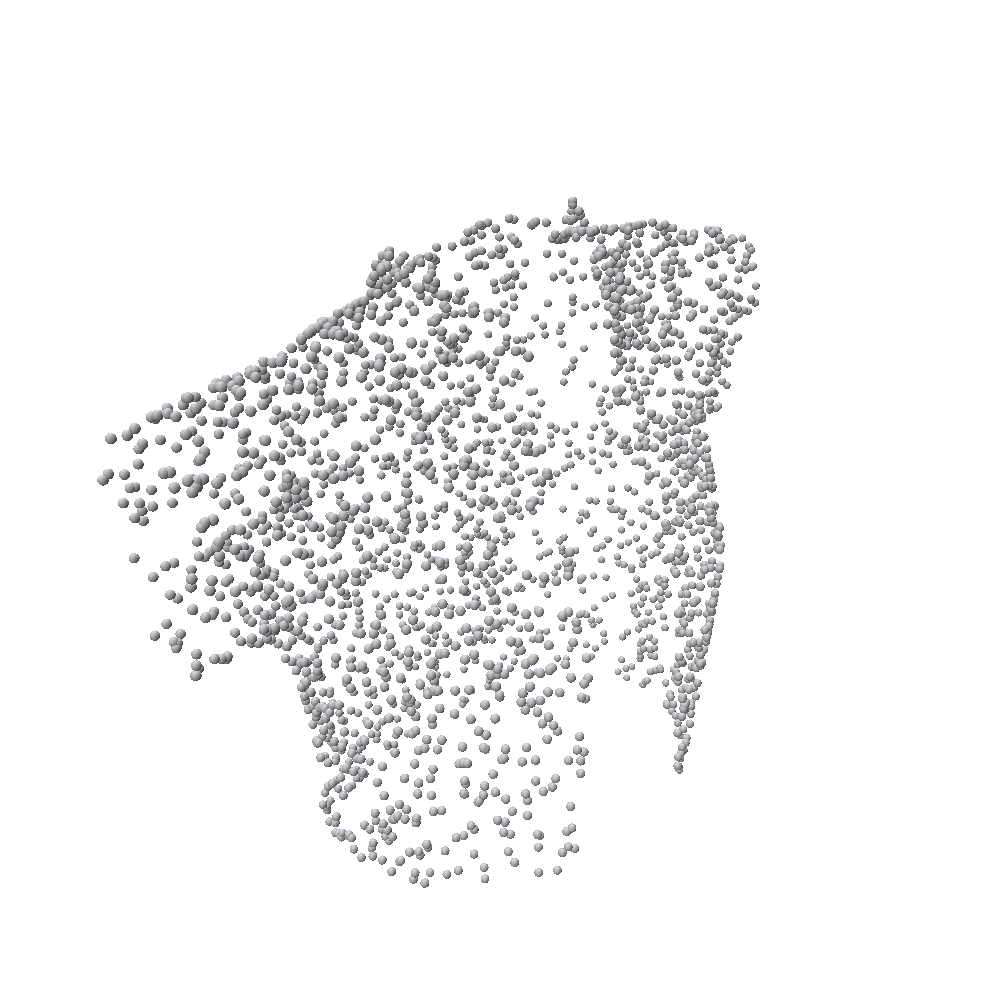}}
            \hfil
            \raisebox{-0.5\height}{\includegraphics[width=0.199\linewidth,trim={4cm 4cm 4cm 4cm},clip]{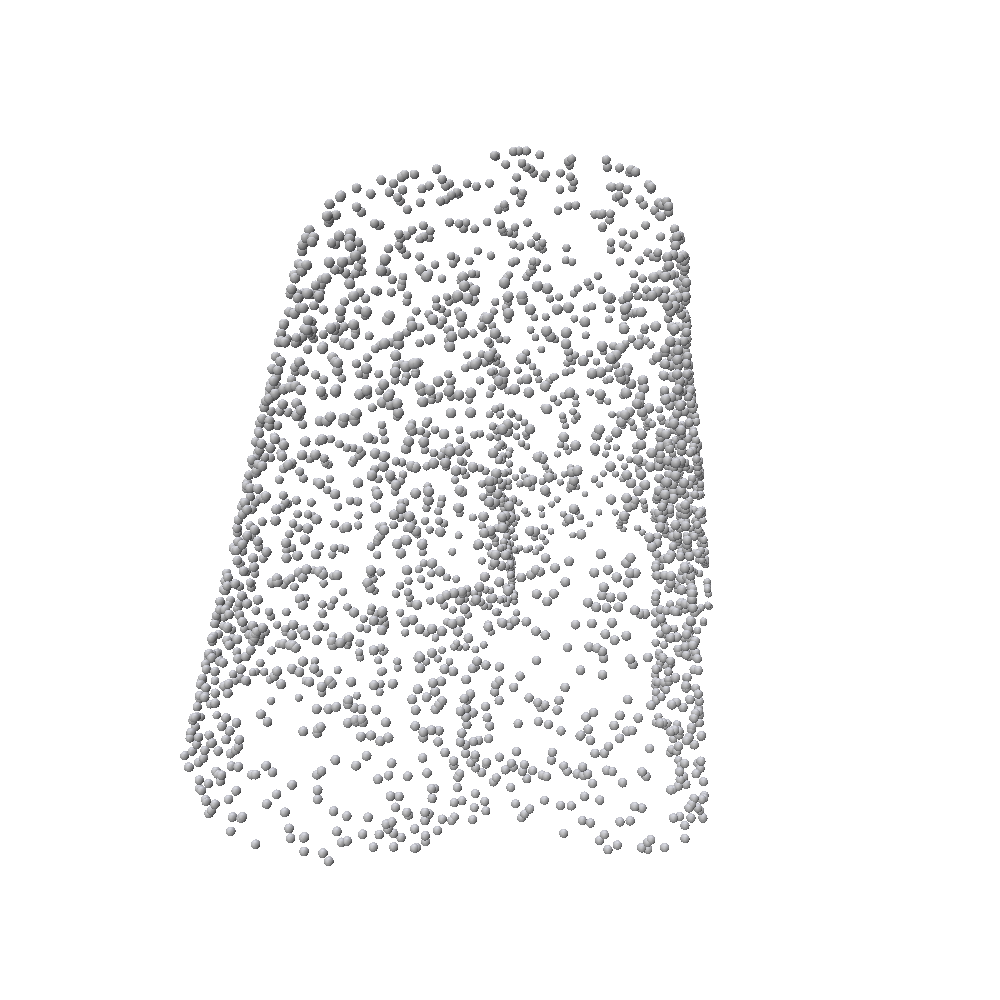}}
		\end{subfigure}
        
        \caption{Distortion of Surface Reconstructions on CLOTH3D++.}
        \label{fig:distortion:cloth3d++}
\end{figure}

\begin{figure}[tp]
\centering 
		\begin{subfigure}{1\linewidth}
        	\centering
            \rotatebox[origin=c]{90}{AtlasNet}
            \raisebox{-0.5\height}{\includegraphics[width=0.245\linewidth,trim={0cm 7cm 2cm 7cm},clip]{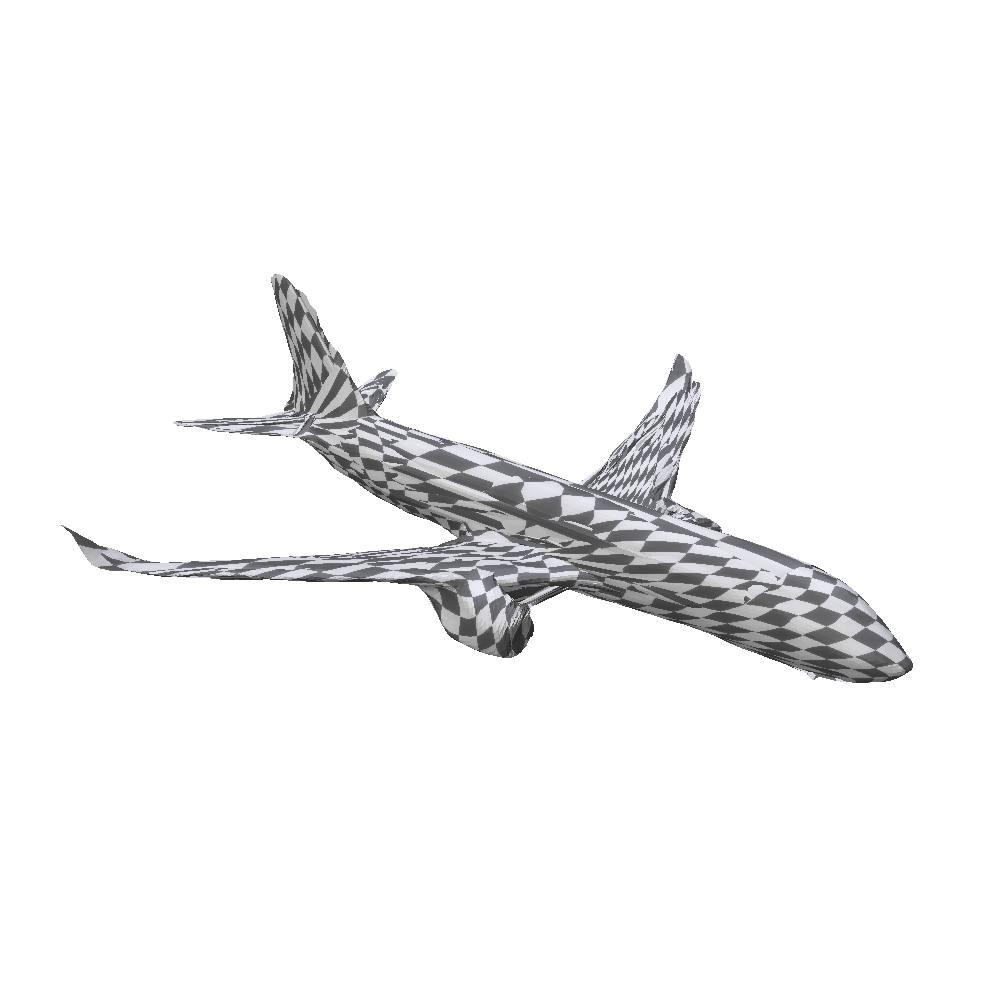}}
            \hfil
			\raisebox{-0.5\height}{\includegraphics[width=0.245\linewidth,trim={2cm 1cm 2cm 7cm},clip]{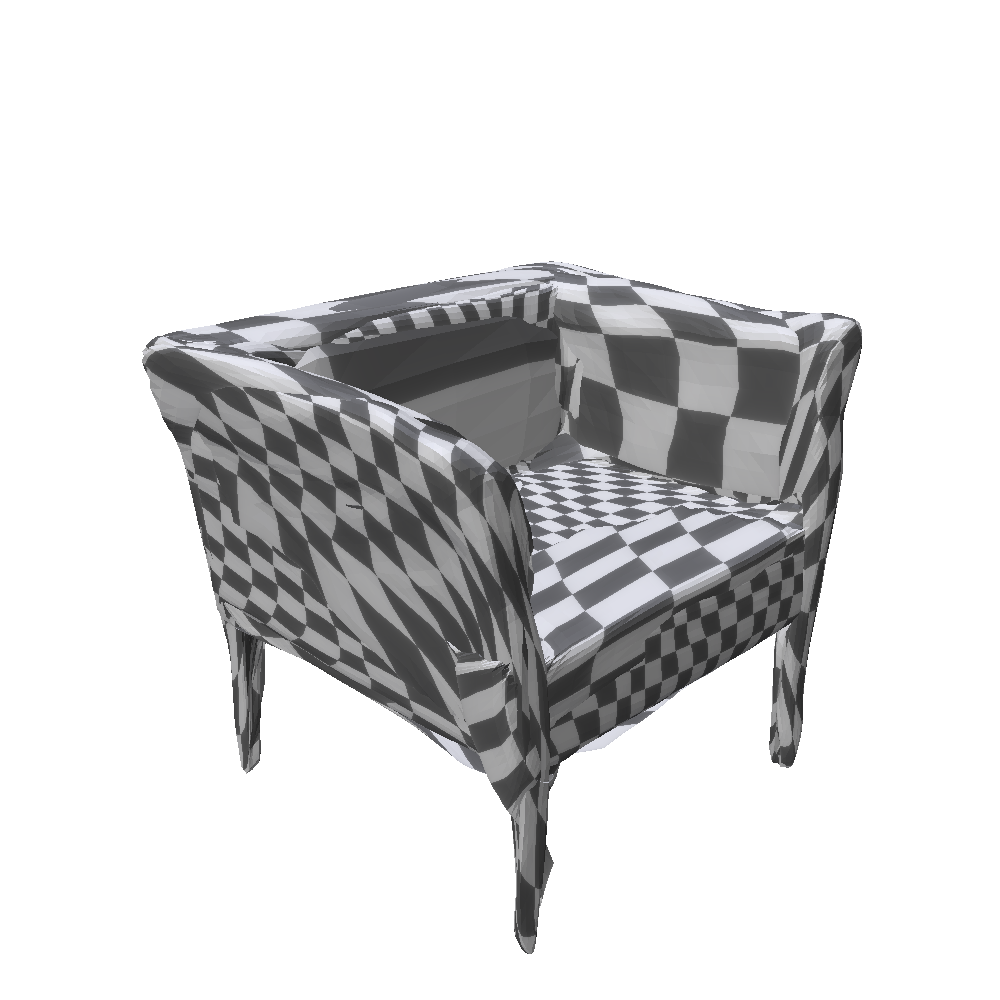}}
            \hfil
            \raisebox{-0.5\height}{\includegraphics[width=0.245\linewidth,trim={4cm 7cm 2cm 7cm},clip]{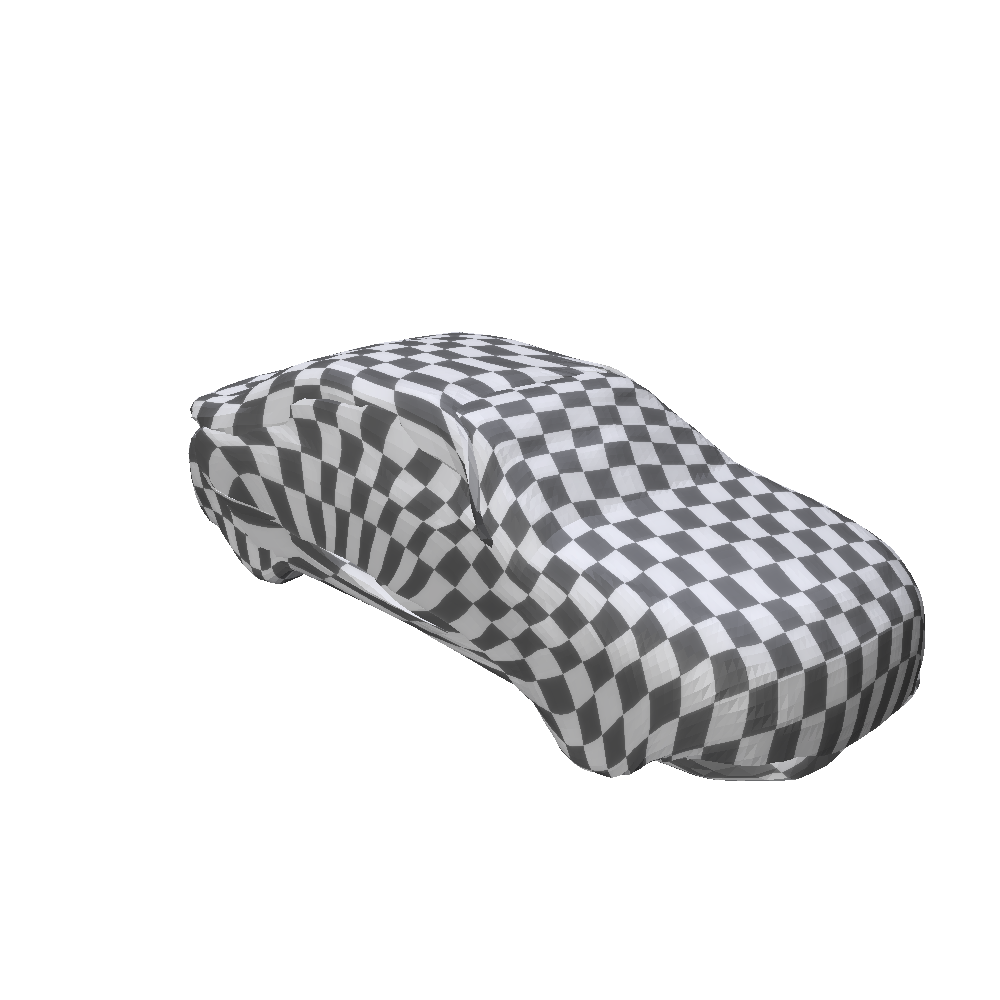}}
		\end{subfigure}
        \begin{subfigure}{1\linewidth}
        	\centering
            \rotatebox[origin=c]{90}{AtlasNet++}
            \raisebox{-0.5\height}{\includegraphics[width=0.245\linewidth,trim={0cm 7cm 2cm 7cm},clip]{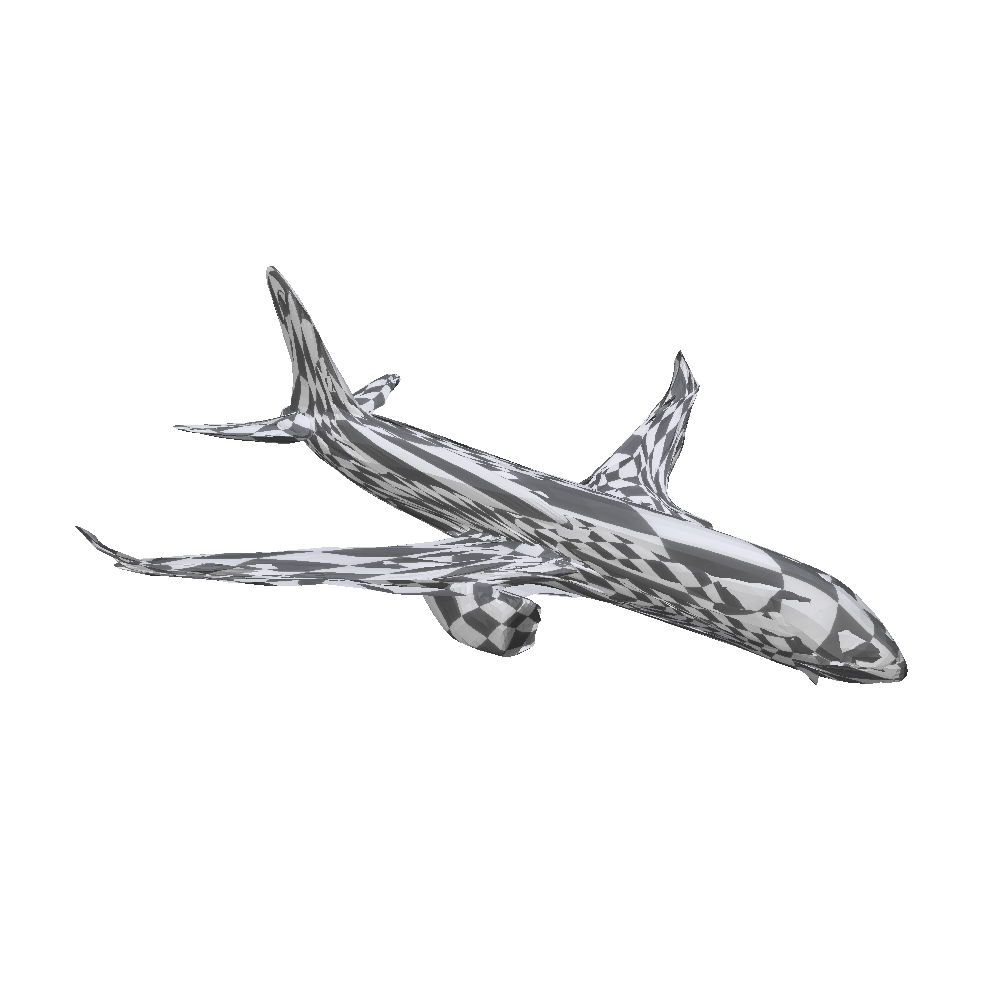}}
            \hfil
			\raisebox{-0.5\height}{\includegraphics[width=0.245\linewidth,trim={2cm 1cm 2cm 7cm},clip]{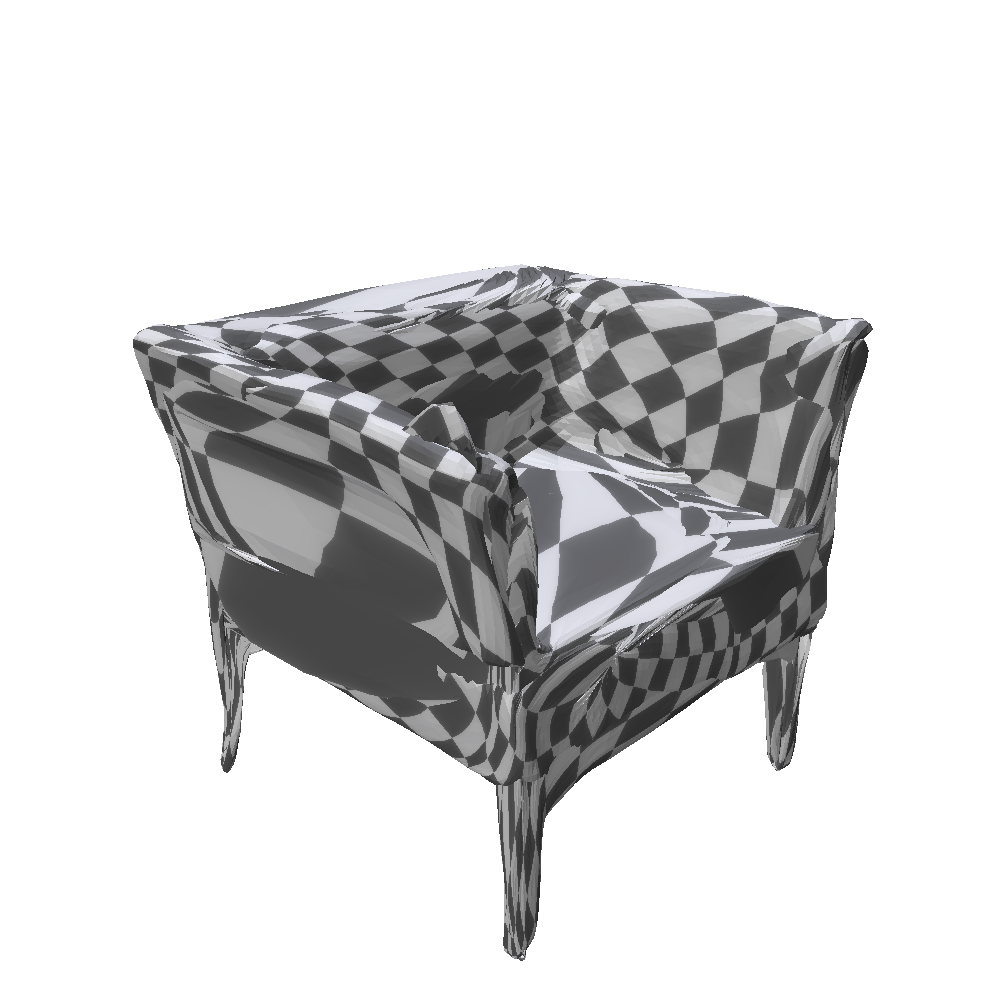}}
            \hfil
            \raisebox{-0.5\height}{\includegraphics[width=0.245\linewidth,trim={4cm 7cm 2cm 7cm},clip]{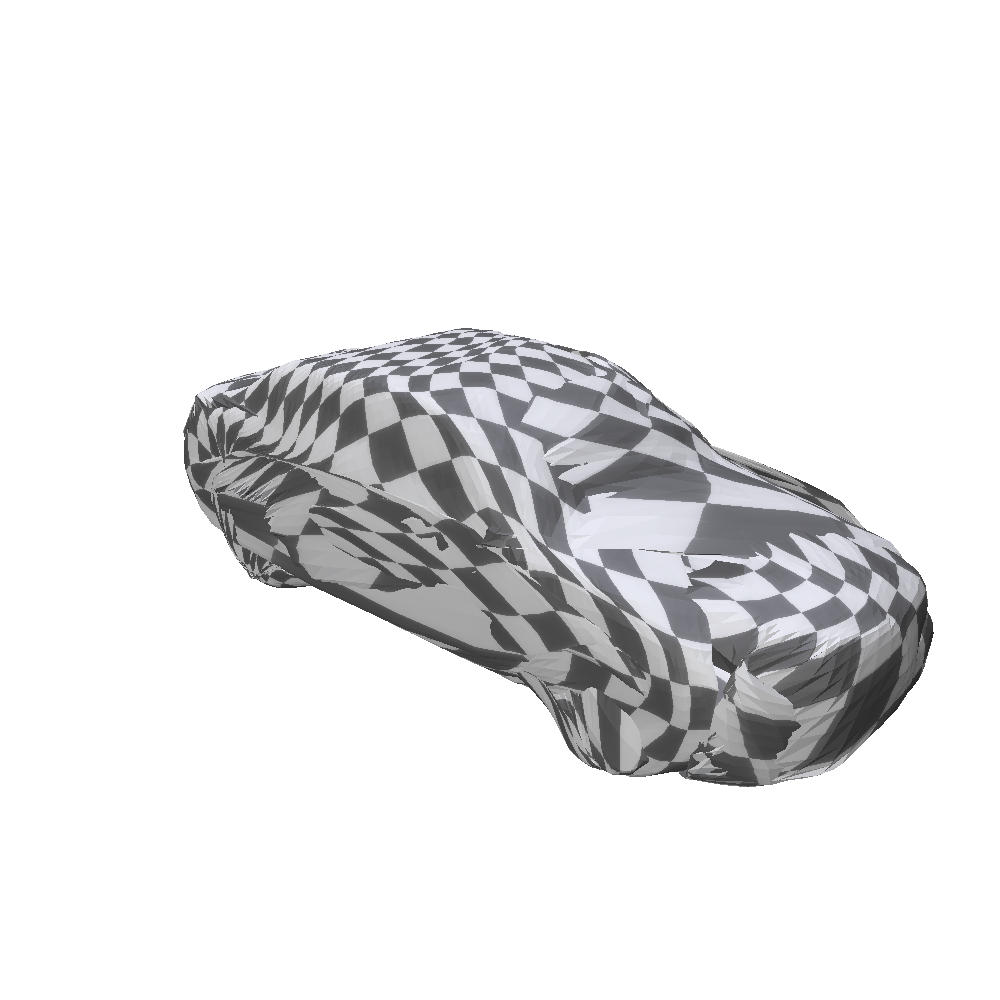}}
		\end{subfigure}
        \begin{subfigure}{1\linewidth}
        	\centering
            \rotatebox[origin=c]{90}{DSP}
            \raisebox{-0.5\height}{\includegraphics[width=0.245\linewidth,trim={0cm 7cm 2cm 7cm},clip]{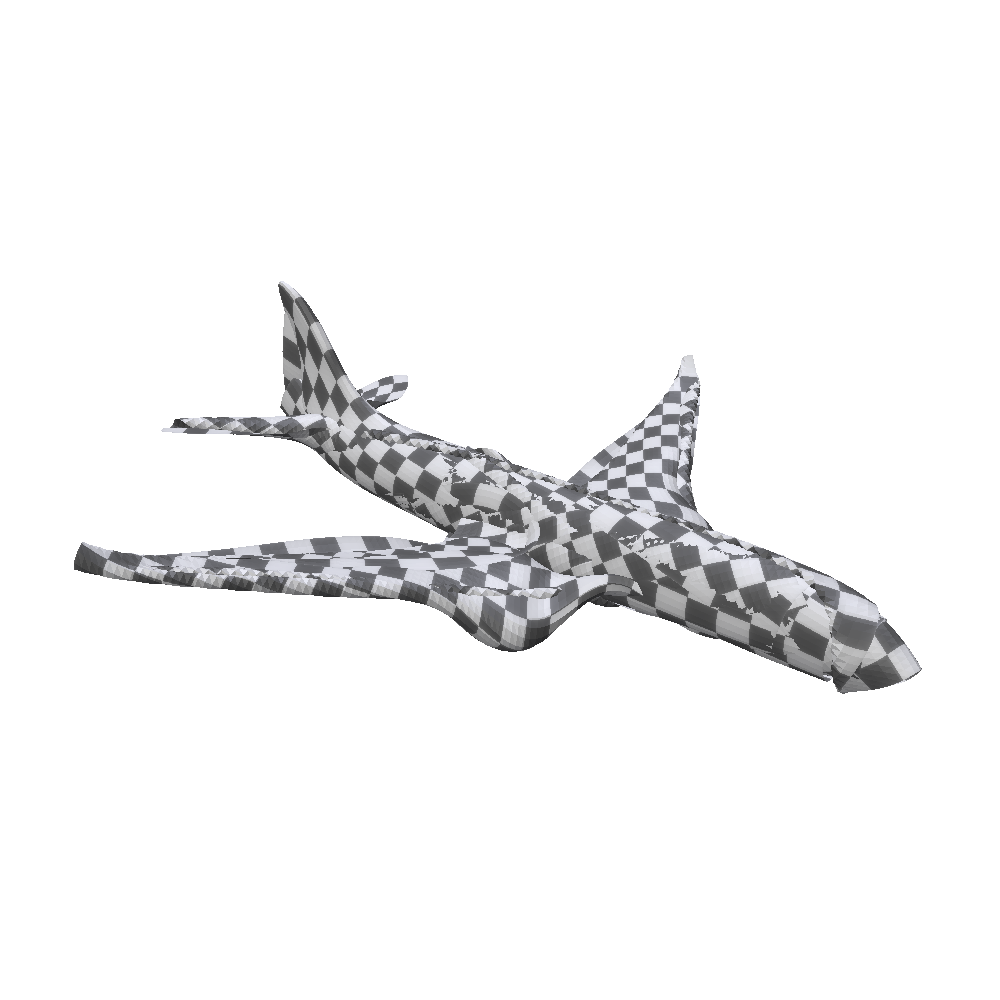}}
            \hfil
			\raisebox{-0.5\height}{\includegraphics[width=0.245\linewidth,trim={2cm 1cm 2cm 7cm},clip]{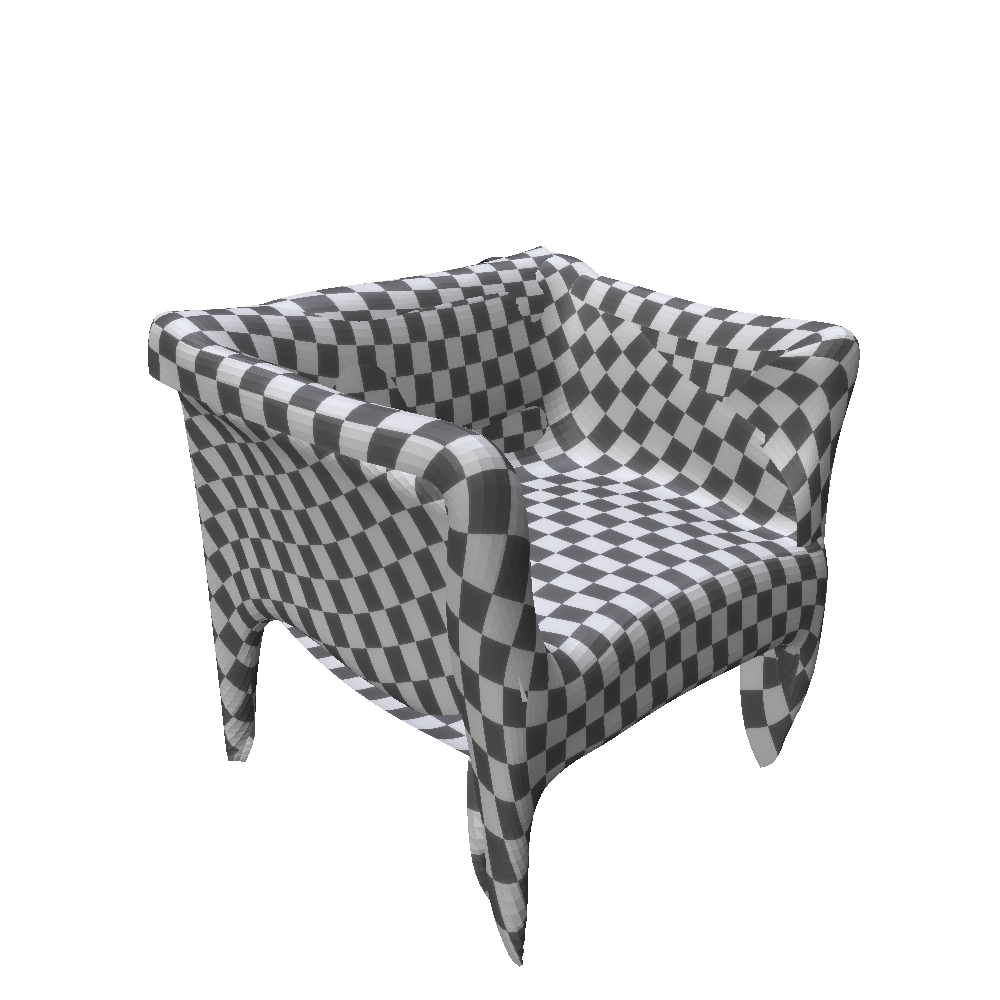}}
            \hfil
            \raisebox{-0.5\height}{\includegraphics[width=0.245\linewidth,trim={4cm 7cm 2cm 7cm},clip]{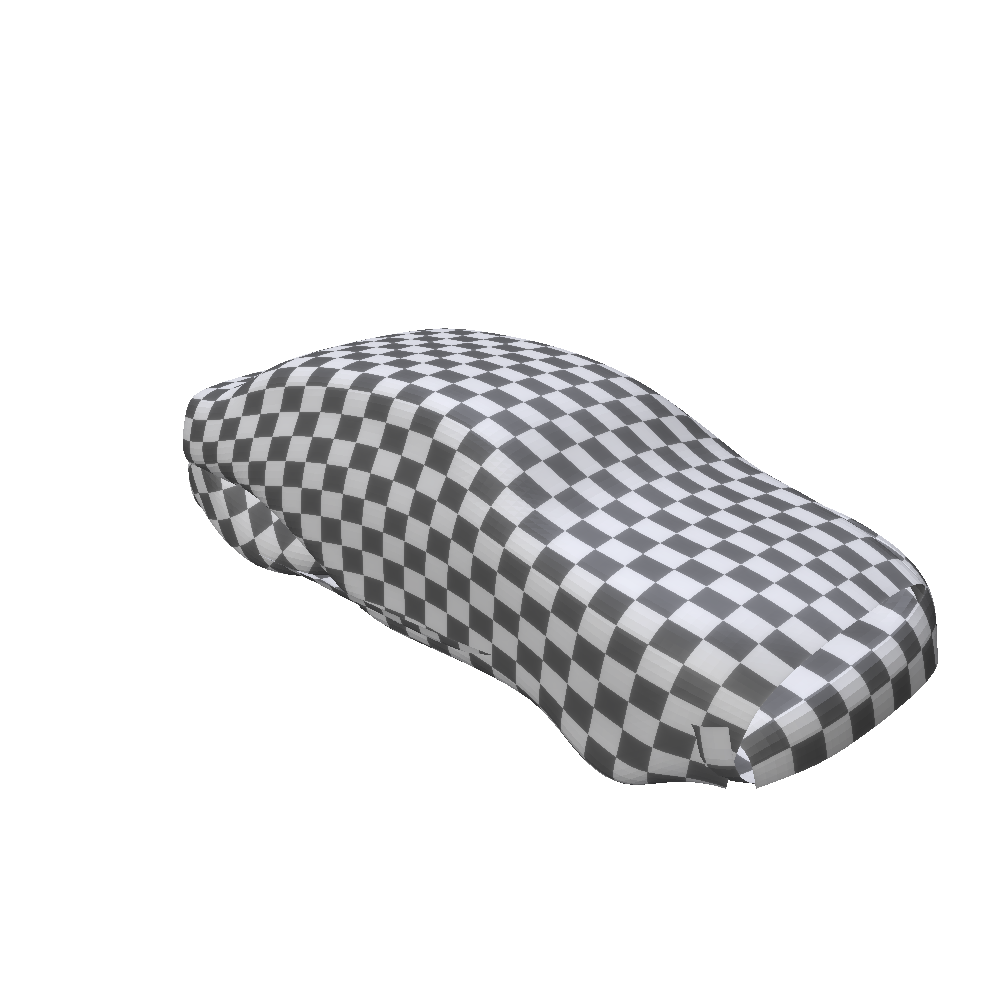}}
		\end{subfigure}
        \begin{subfigure}{1\linewidth}
        	\centering
            \rotatebox[origin=c]{90}{TearingNet}
            \raisebox{-0.5\height}{\includegraphics[width=0.245\linewidth,trim={0cm 7cm 2cm 7cm},clip]{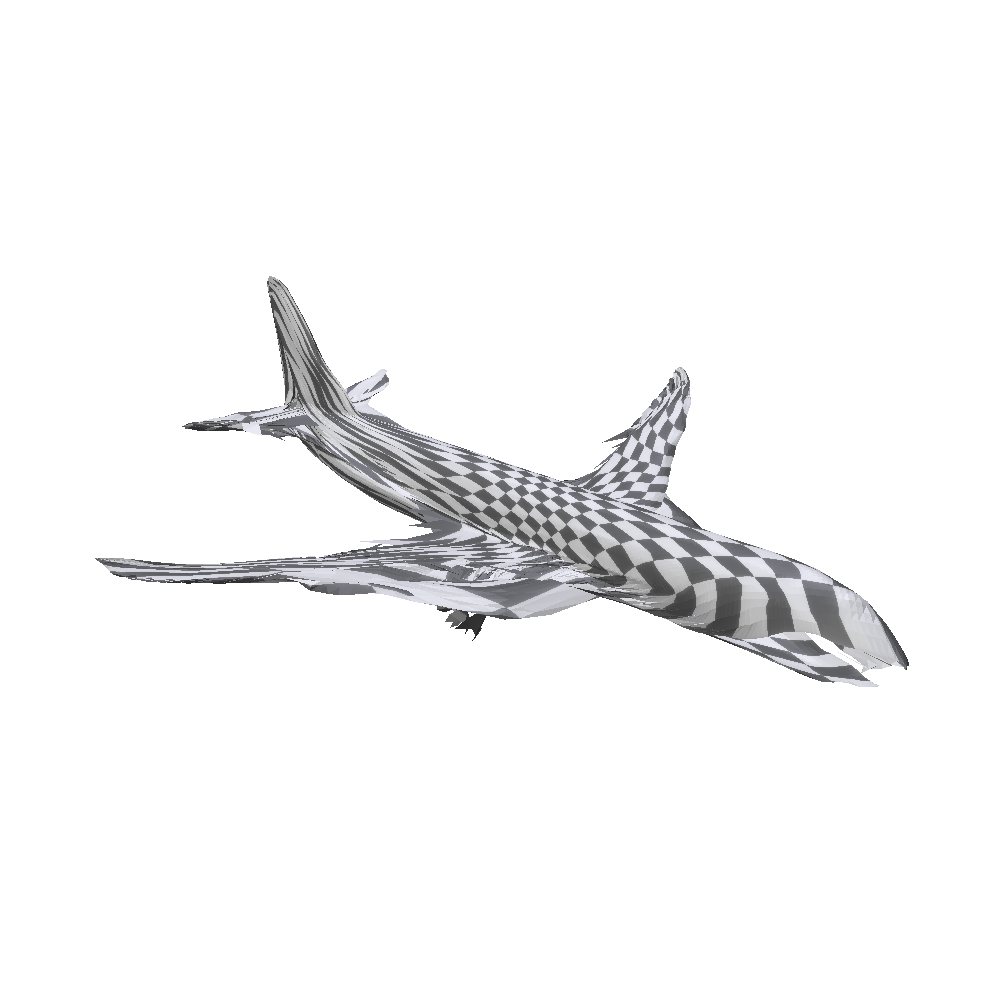}}
            \hfil
			\raisebox{-0.5\height}{\includegraphics[width=0.245\linewidth,trim={2cm 1cm 2cm 7cm},clip]{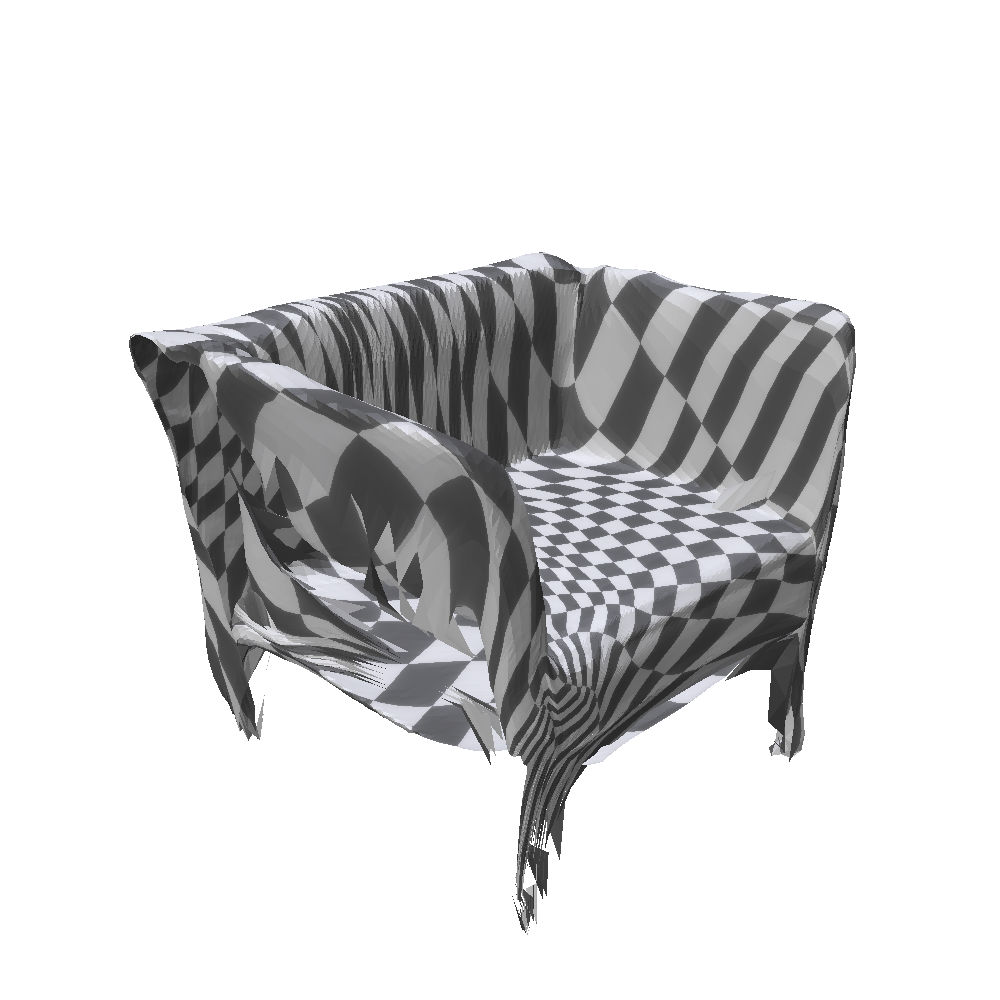}}
            \hfil
            \raisebox{-0.5\height}{\includegraphics[width=0.245\linewidth,trim={4cm 7cm 2cm 7cm},clip]{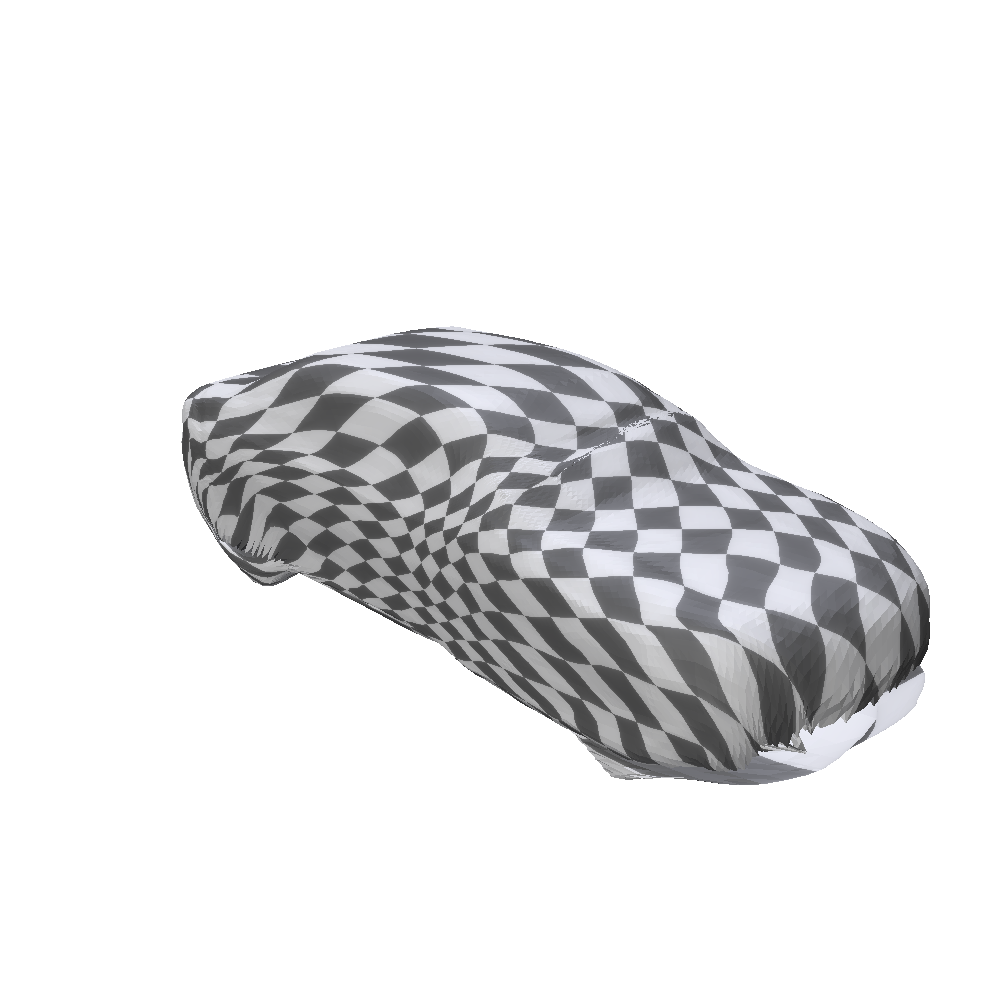}}
		\end{subfigure}
        \begin{subfigure}{1\linewidth}
        	\centering
            \rotatebox[origin=c]{90}{Ours}
            \raisebox{-0.5\height}{\includegraphics[width=0.245\linewidth,trim={0cm 7cm 2cm 7cm},clip]{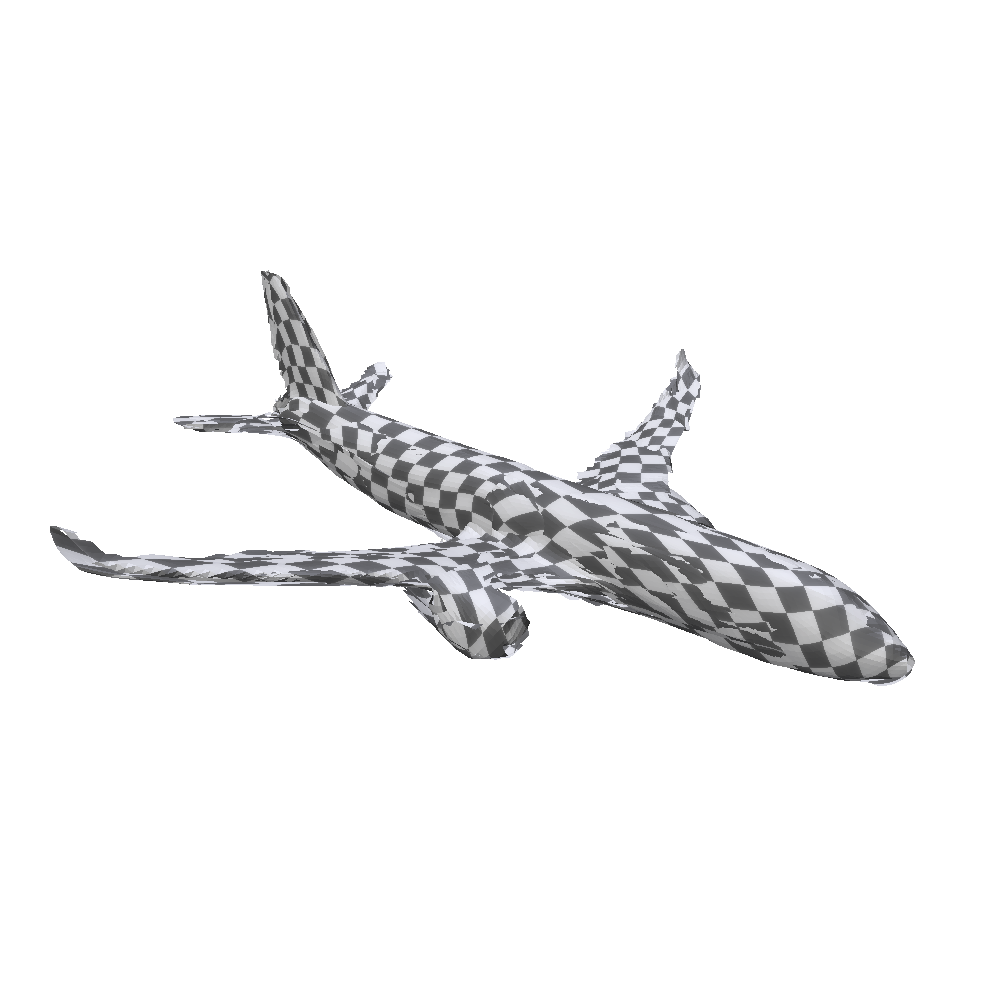}}
            \hfil
			\raisebox{-0.5\height}{\includegraphics[width=0.245\linewidth,trim={2cm 1cm 2cm 7cm},clip]{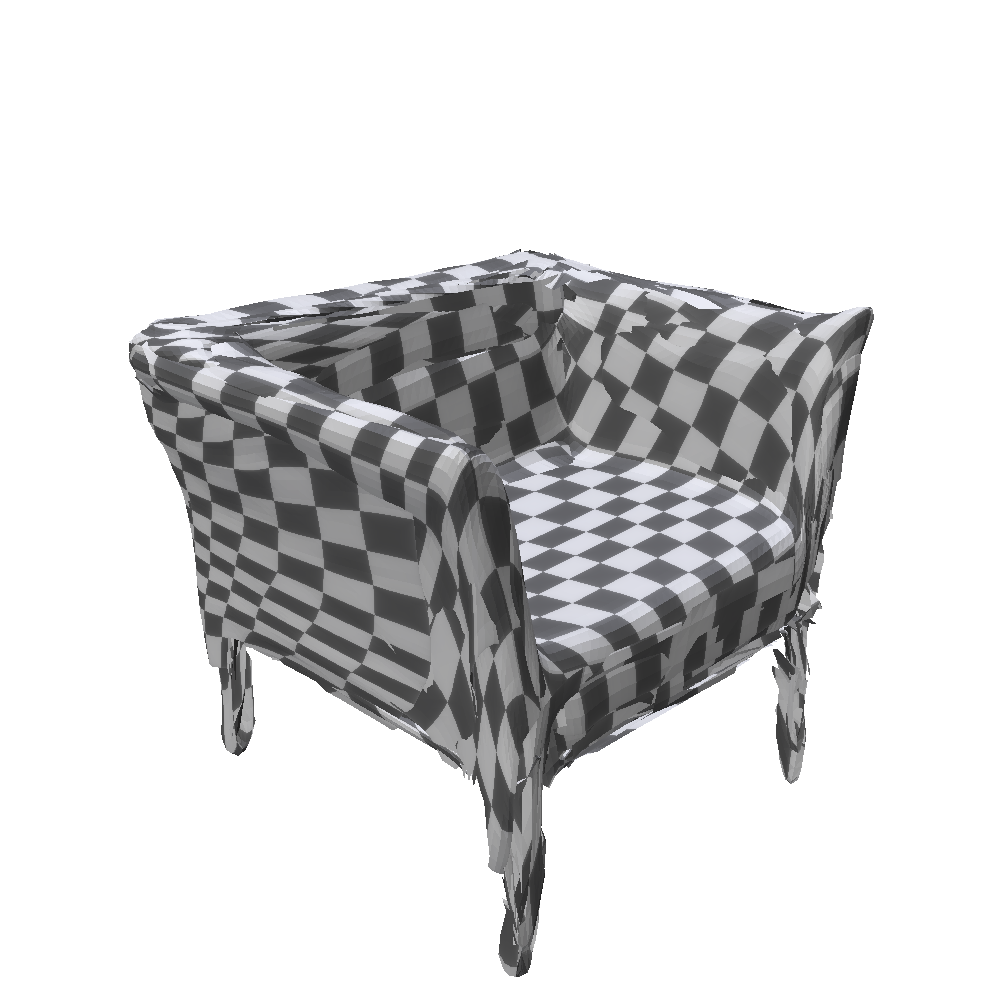}}
            \hfil
            \raisebox{-0.5\height}{\includegraphics[width=0.245\linewidth,trim={4cm 7cm 2cm 7cm},clip]{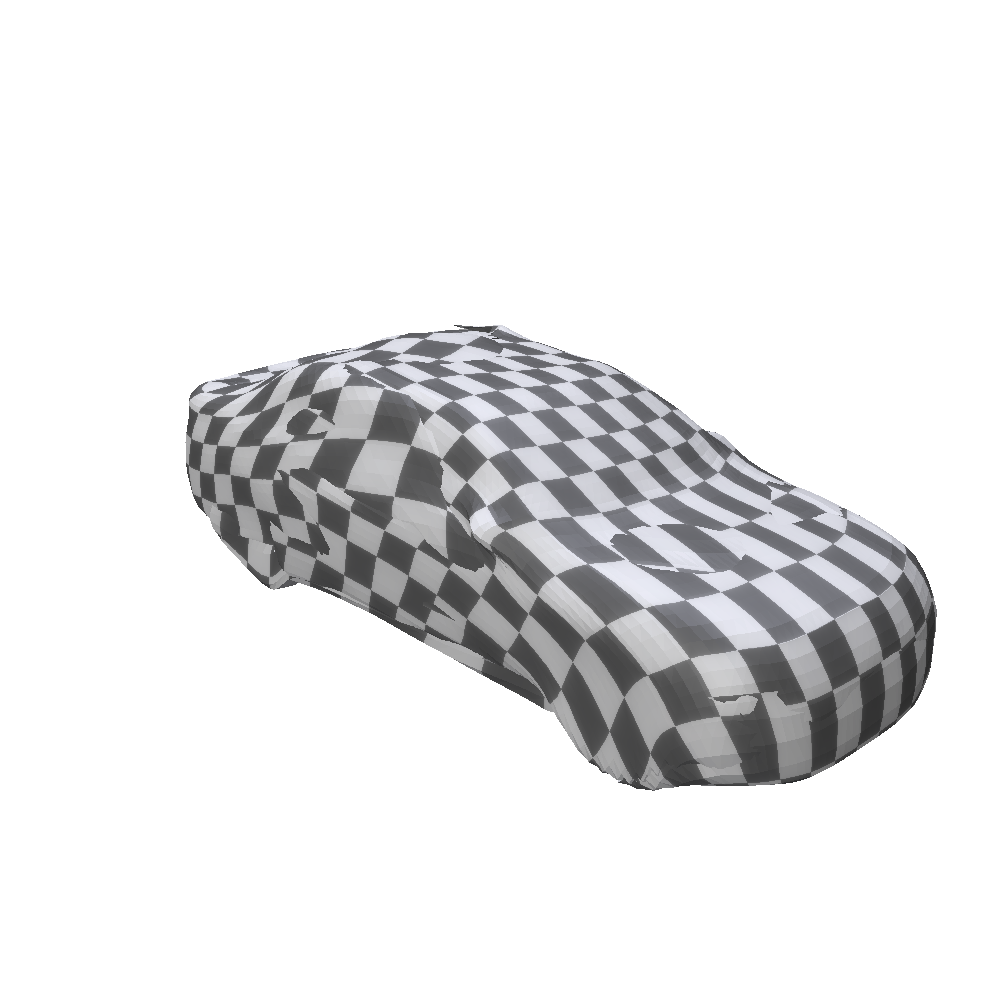}}
		\end{subfigure}
        \begin{subfigure}{1\linewidth}
        	\centering
            \rotatebox[origin=c]{90}{Target}
            \raisebox{-0.5\height}{\includegraphics[width=0.245\linewidth,trim={0cm 7cm 2cm 7cm},clip]{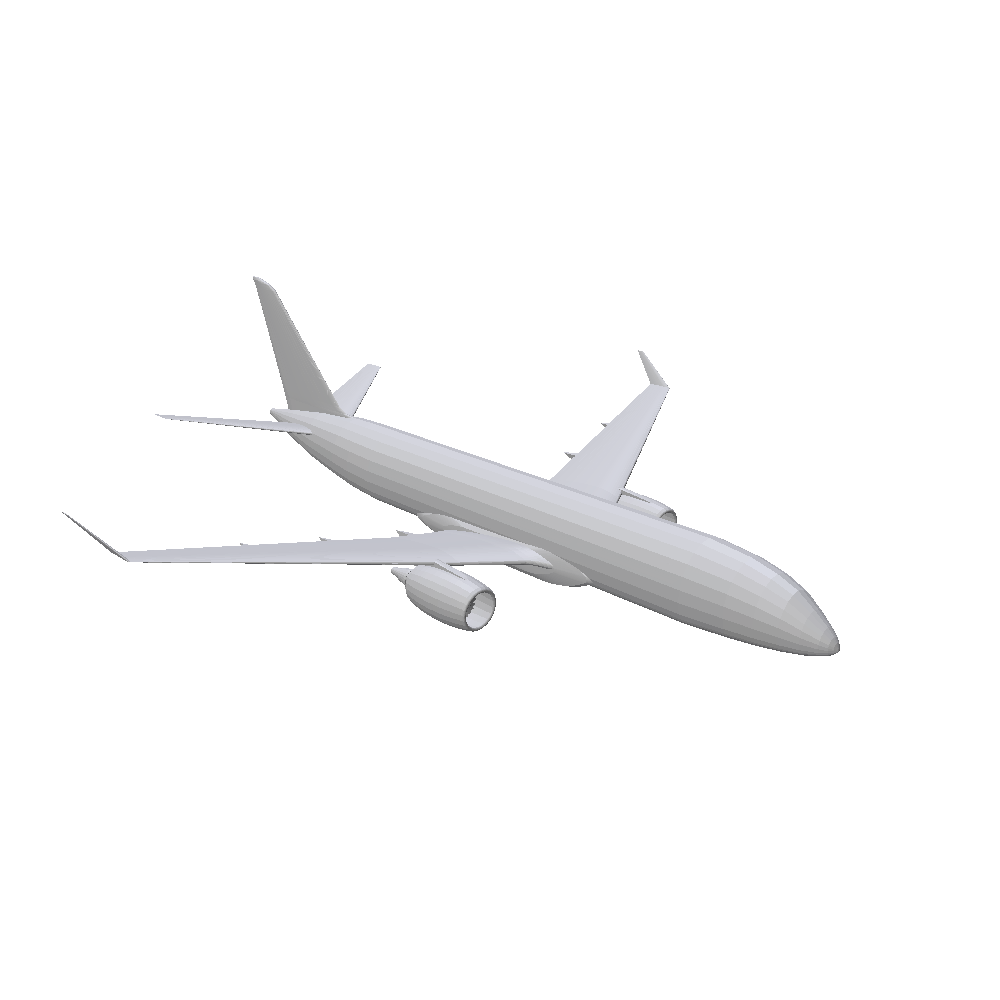}}
            \hfil
			\raisebox{-0.5\height}{\includegraphics[width=0.245\linewidth,trim={2cm 1cm 2cm 7cm},clip]{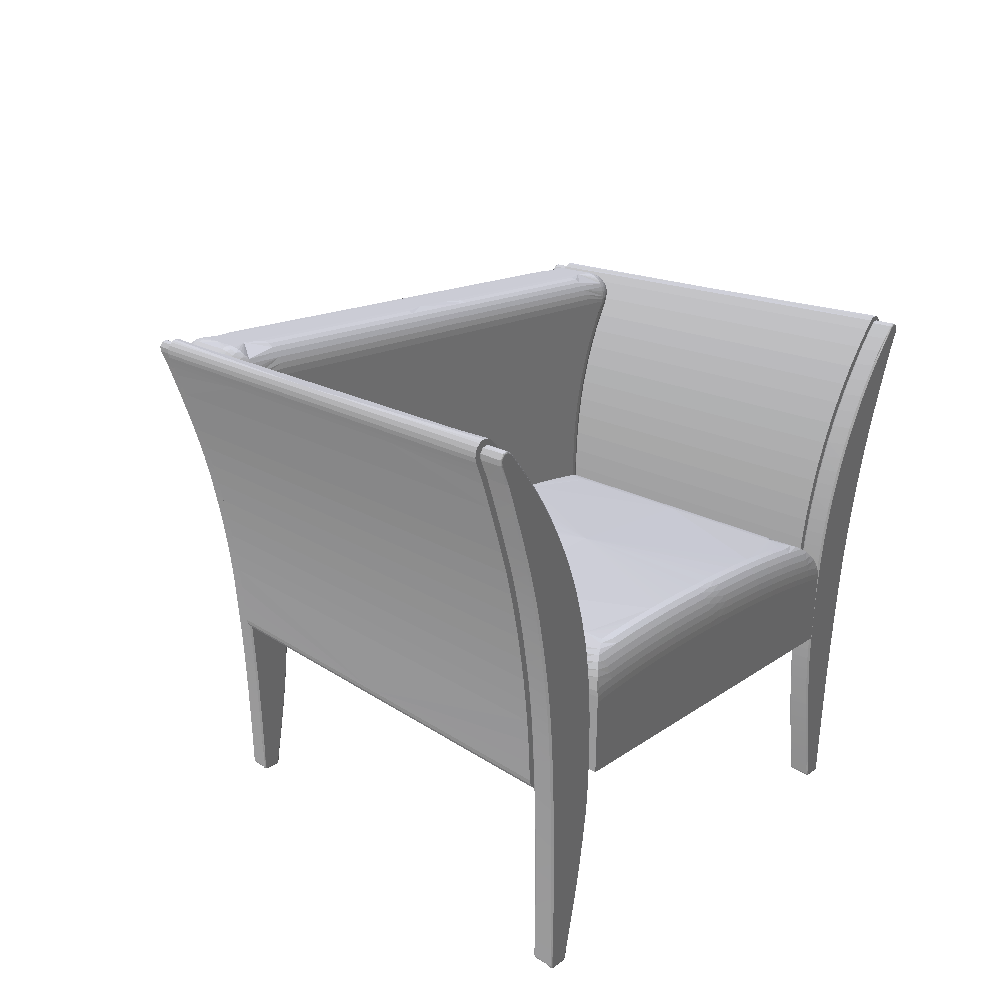}}
            \hfil
            \raisebox{-0.5\height}{\includegraphics[width=0.245\linewidth,trim={4cm 7cm 2cm 7cm},clip]{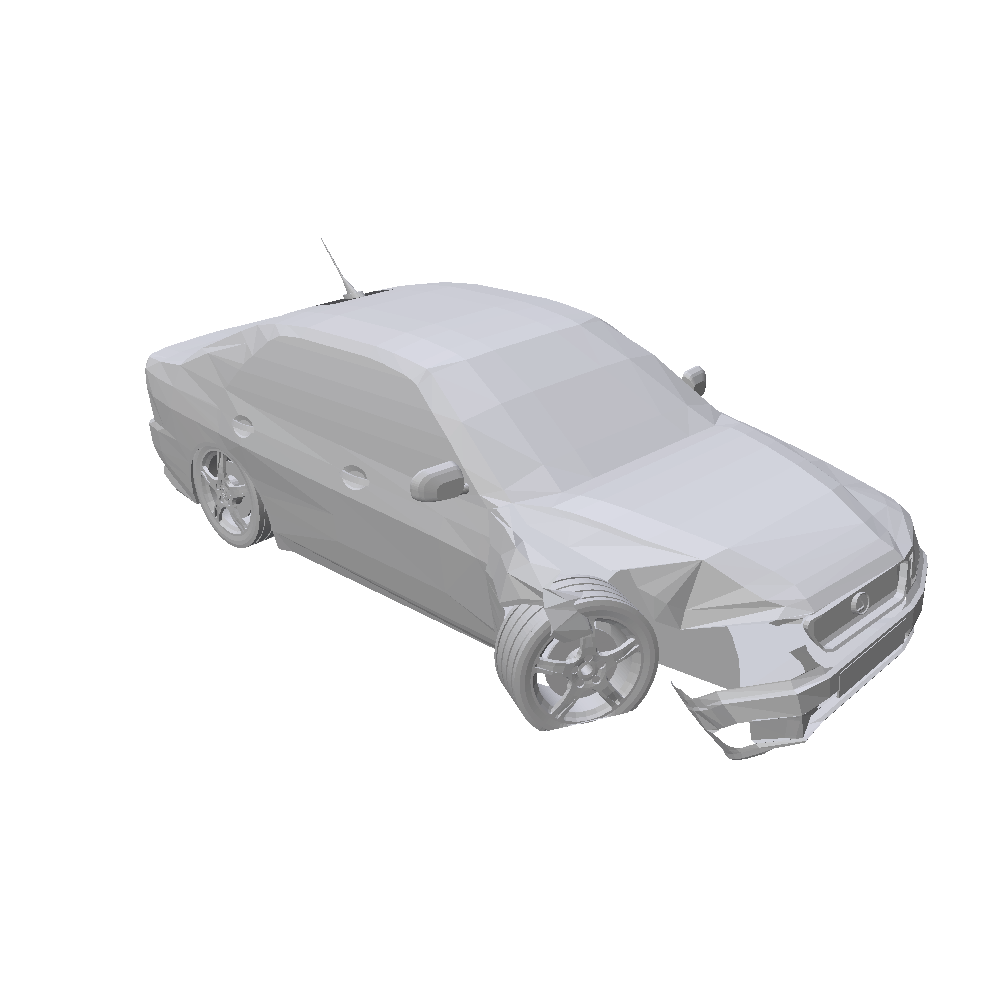}}
		\end{subfigure}
        \begin{subfigure}{1\linewidth}
        	\centering
            \rotatebox[origin=c]{90}{Input}
            \raisebox{-0.5\height}{\includegraphics[width=0.245\linewidth,trim={0cm 7cm 2cm 7cm},clip]{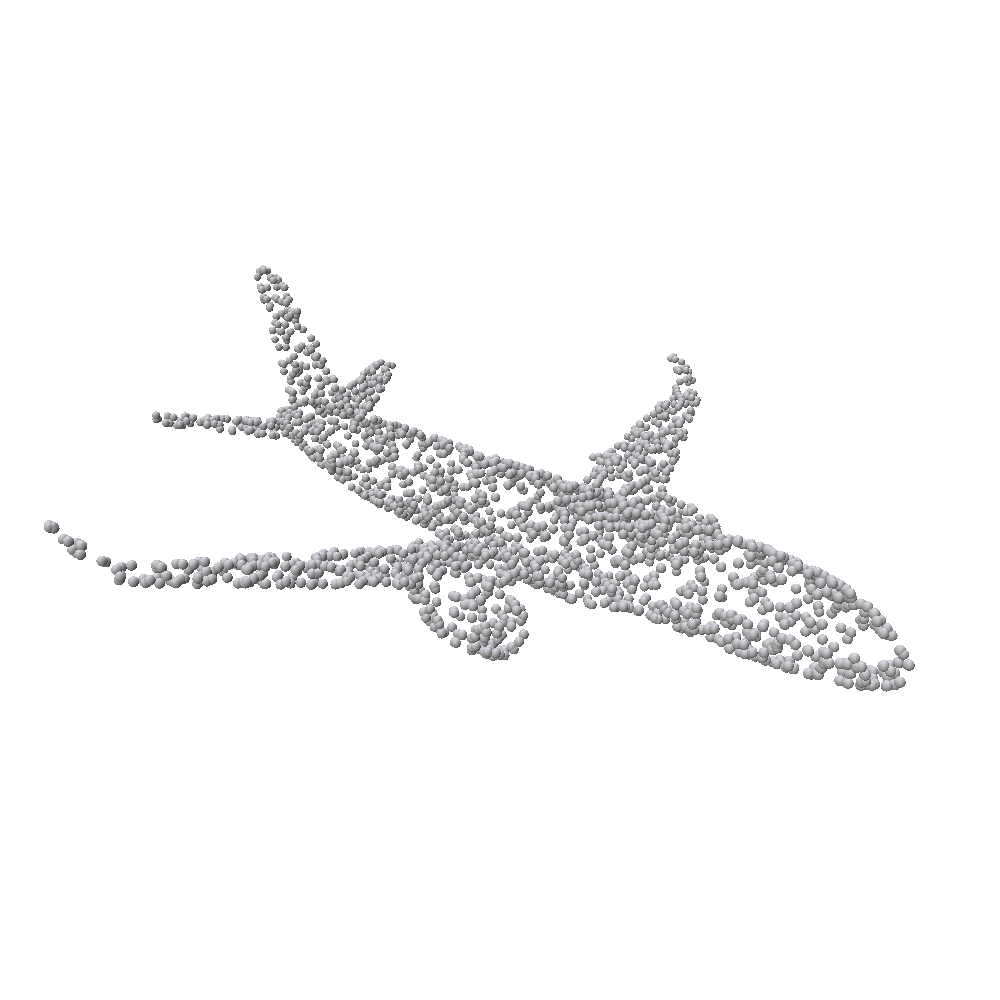}}
            \hfil
			\raisebox{-0.5\height}{\includegraphics[width=0.245\linewidth,trim={2cm 1cm 2cm 7cm},clip]{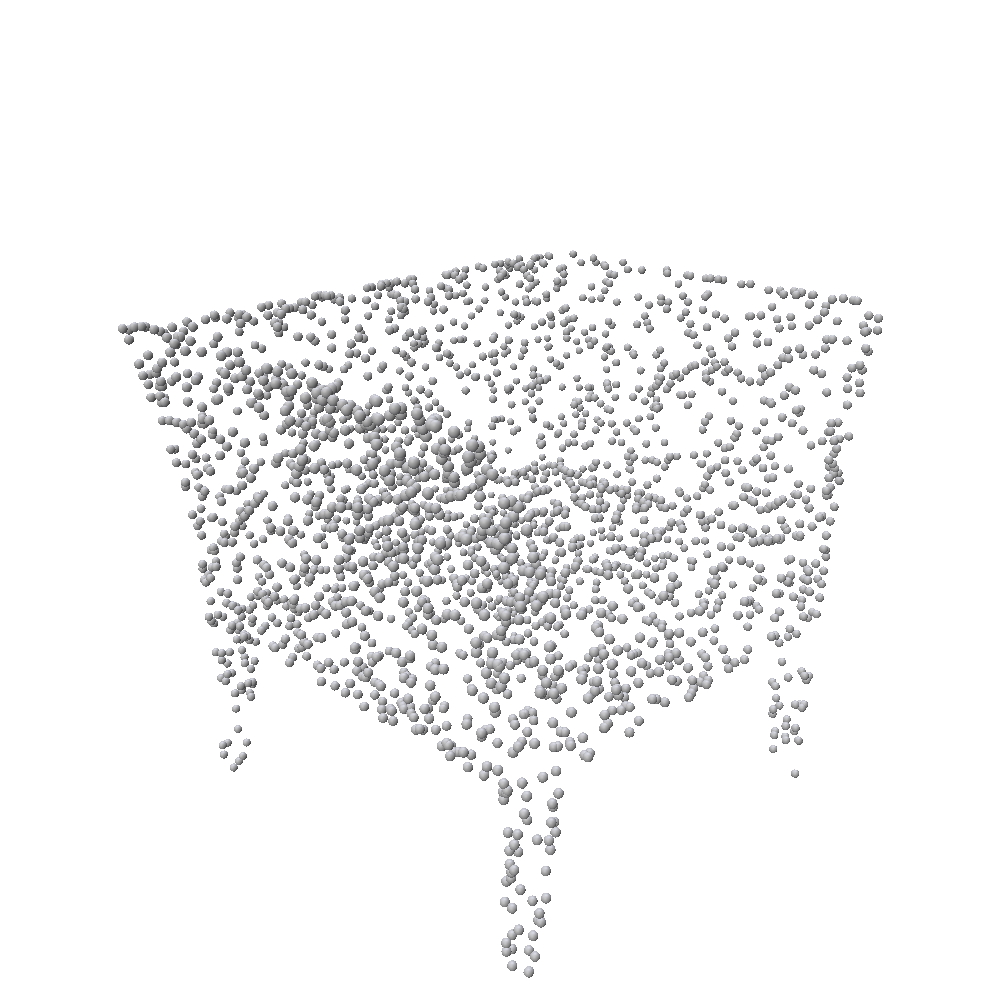}}
            \hfil
            \raisebox{-0.5\height}{\includegraphics[width=0.245\linewidth,trim={4cm 7cm 2cm 7cm},clip]{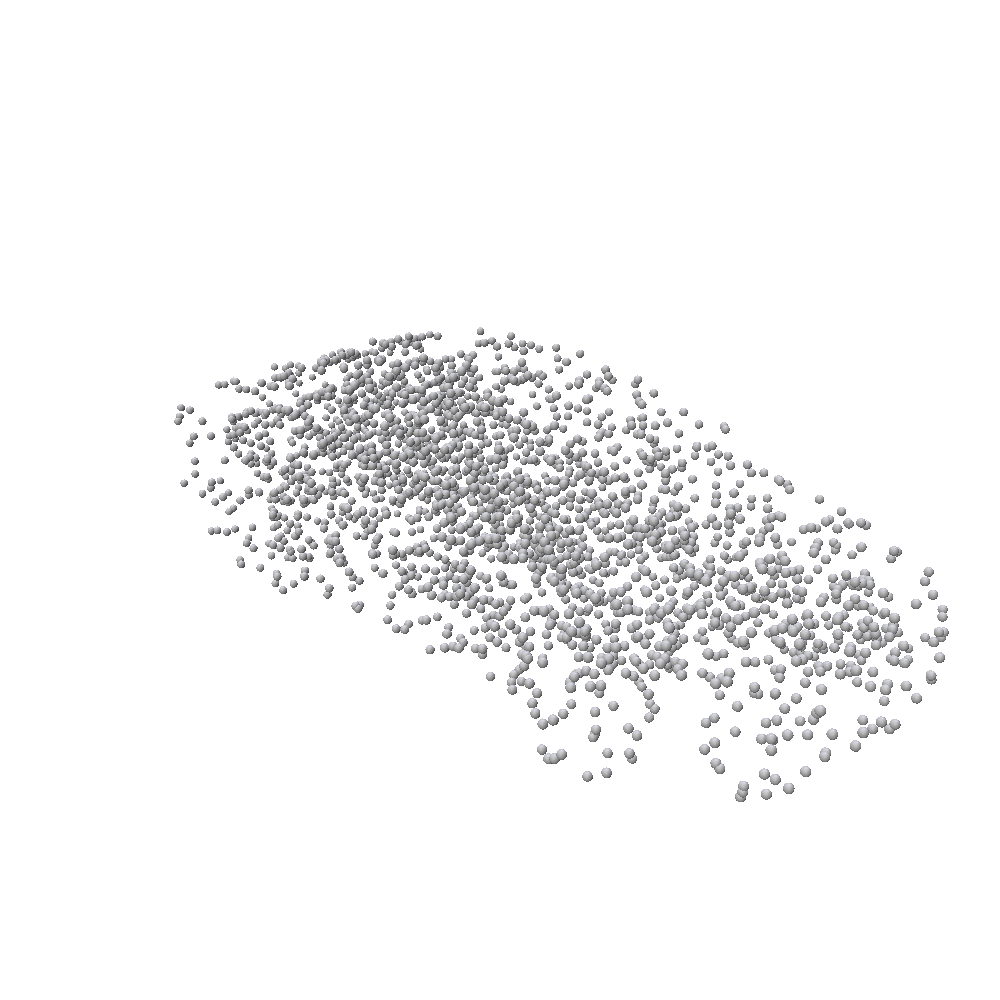}}
		\end{subfigure}
        
        \caption{Distortion of Surface Reconstructions on ShapeNet.}
        \label{fig:distortion:shapenet}
\end{figure}

\par
Fig.~\ref{fig:distortion:cloth3d++} and Fig.~\ref{fig:distortion:shapenet} illustrate the distortion of the reconstructed chart parameterizations in the surface reconstruction experiment on CLOTH3D++ and ShapeNet, respectively. We employ 2 charts on CLOTH3D++ and 3 charts on ShapeNet for all surface representations with the exception of TearingNet where 1 chart is used. The relative level of distortion observed reflect the quantitative metrics previously reported, where DSP and minimal neural atlas exhibit significantly lower distortion compared to other baselines.

\subsection{Artifacts of SCAR Violation}
\label{subsec:results:artifacts}

\begin{figure}[t]
\centering 
		\begin{subfigure}{0.25\linewidth}
    		\centering
			\caption*{Ours}
			\includegraphics[width=1\linewidth,trim={4.5cm 5cm 4.5cm 1cm},clip]{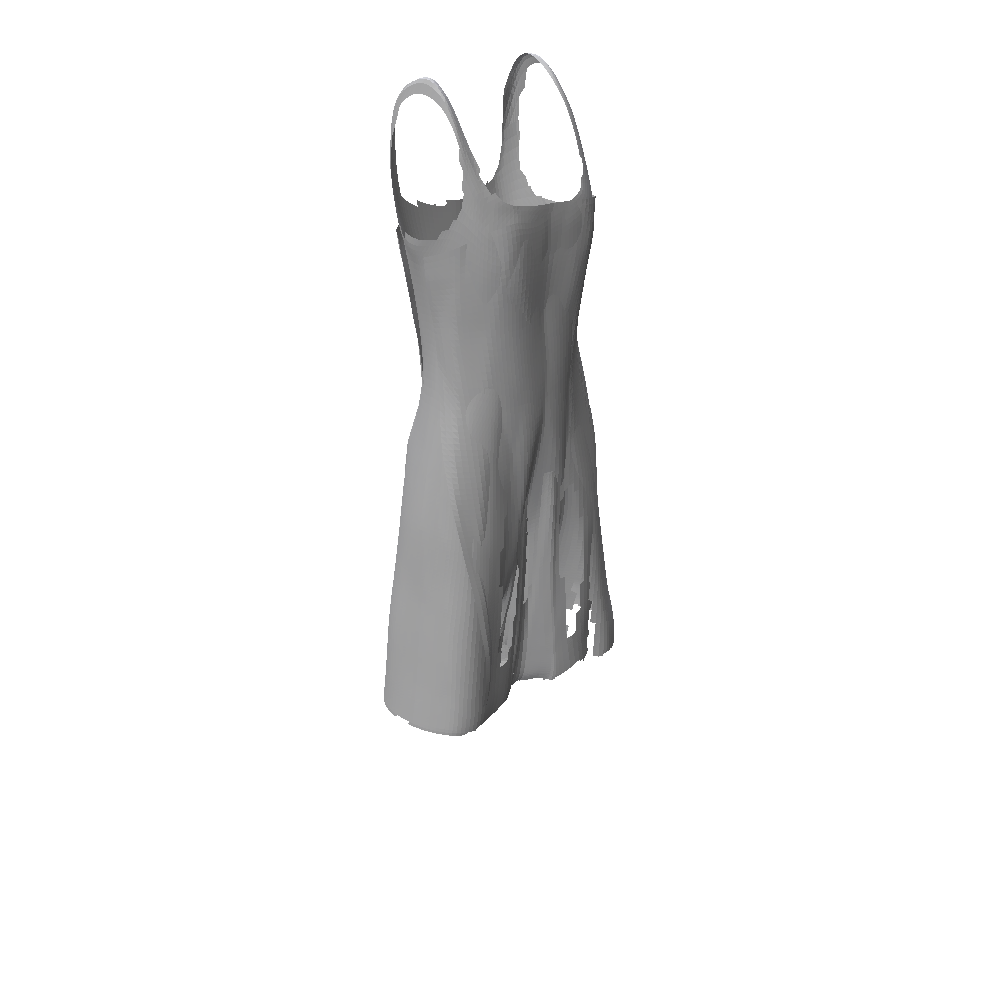}
            \includegraphics[width=1\linewidth,trim={3cm 4.5cm 3cm 5cm},clip]{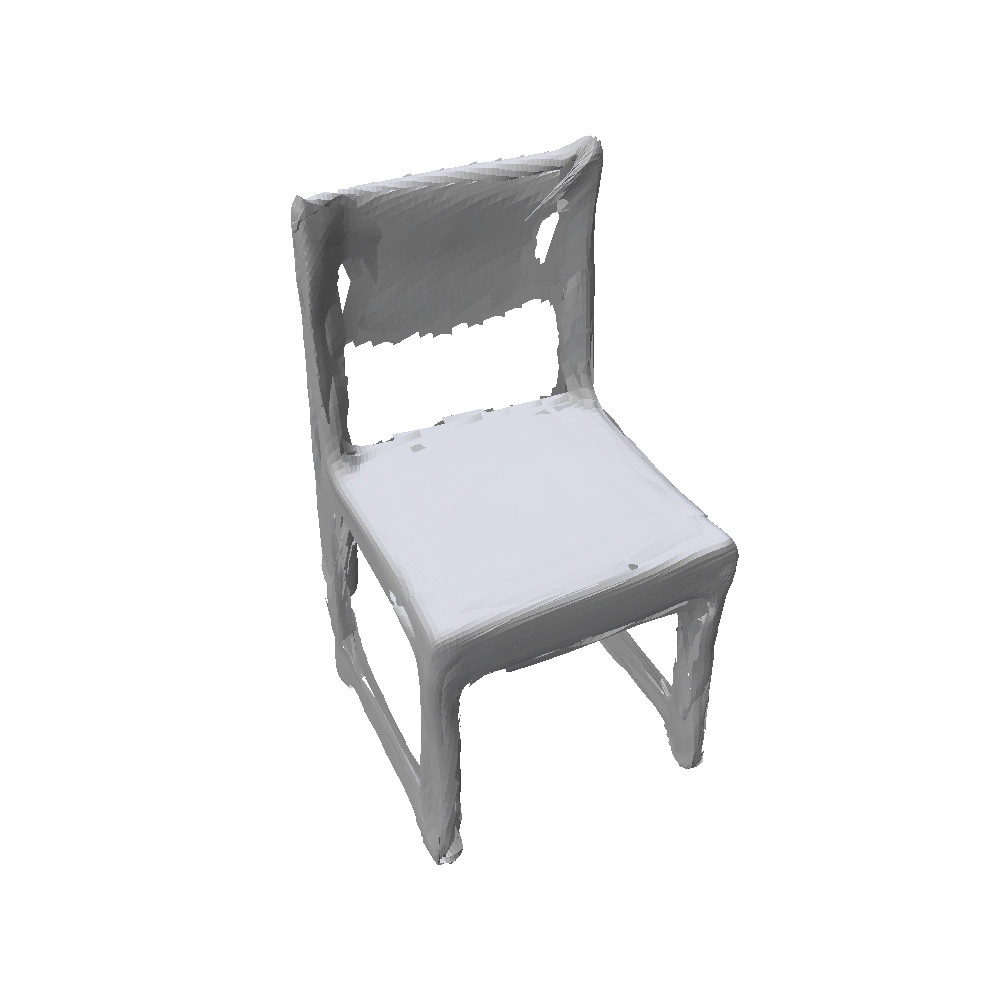}
		\end{subfigure}
        \hfil
        \begin{subfigure}{0.25\linewidth}
    		\centering
			\caption*{Target}
			\includegraphics[width=1\linewidth,trim={3.2cm 5.5cm 3.2cm 1.2cm},clip]{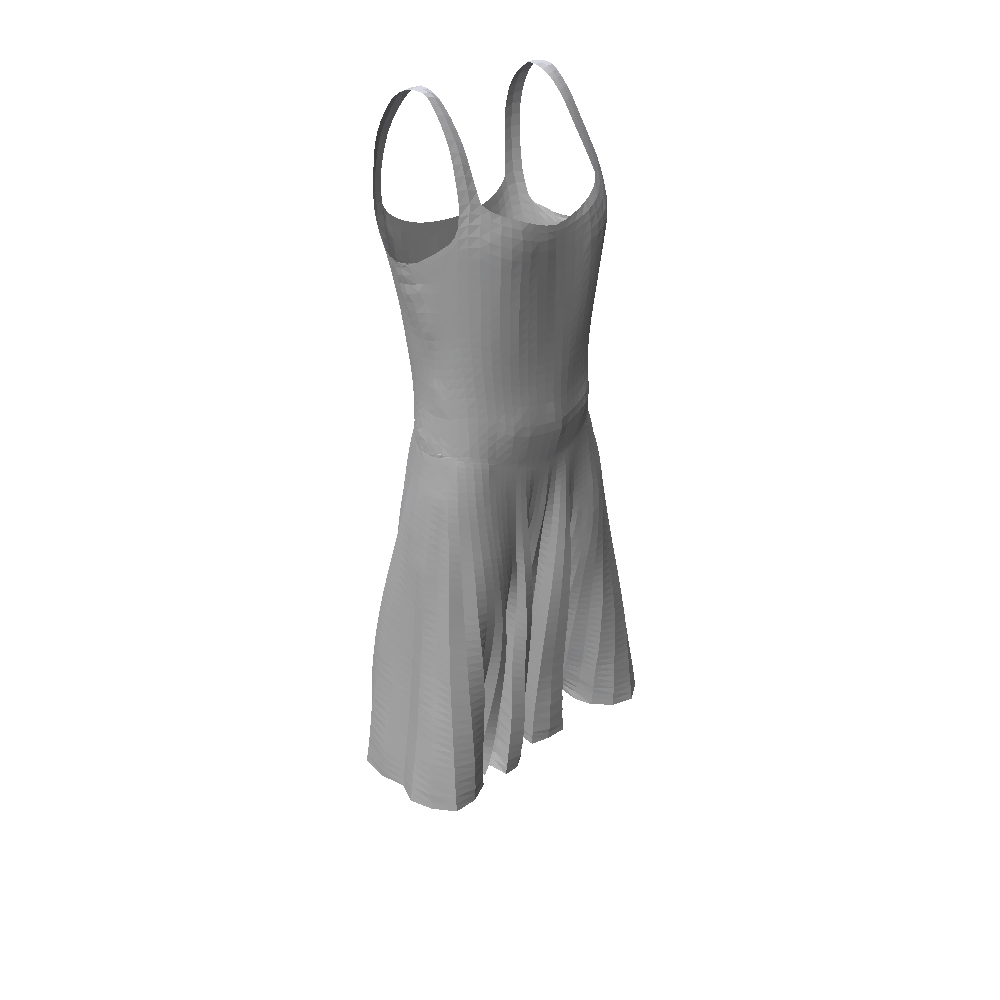}
            \includegraphics[width=1\linewidth,trim={5cm 4.5cm 5cm 5cm},clip]{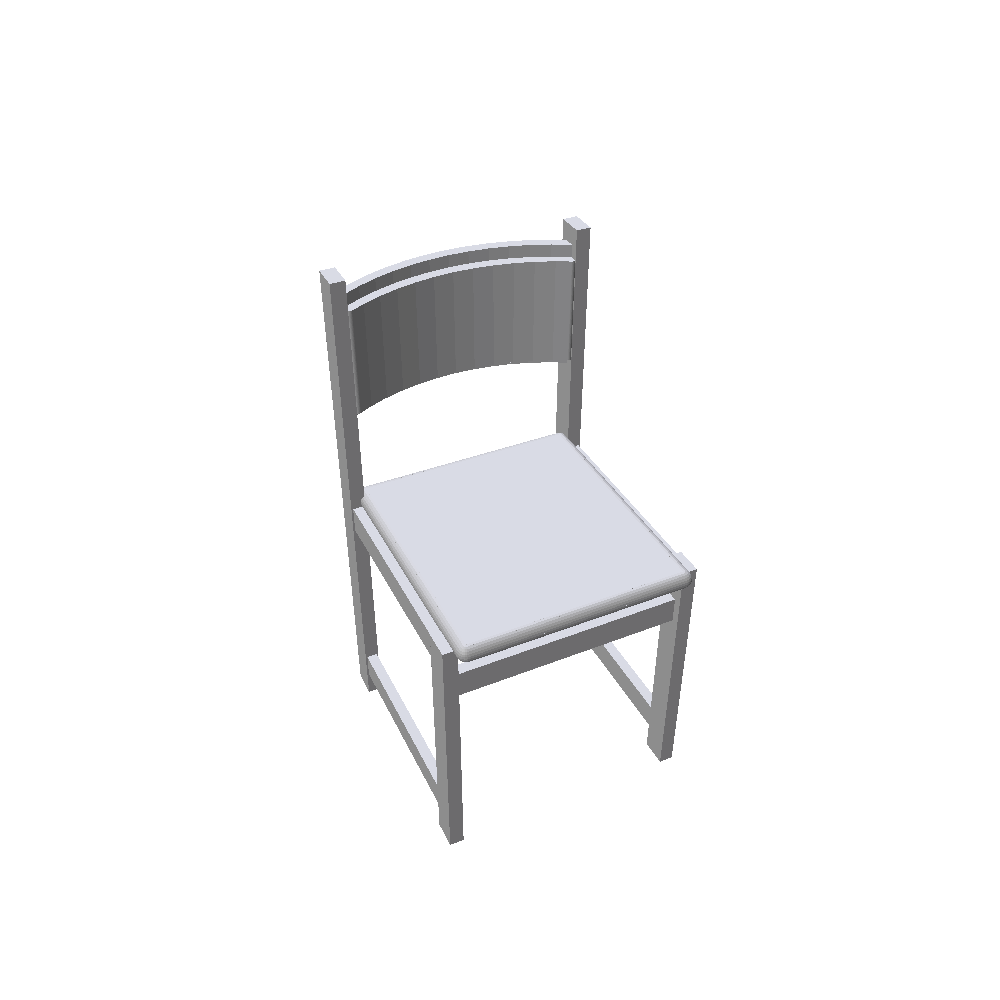}
		\end{subfigure}
        \hfil
        \begin{subfigure}{0.25\linewidth}
    		\centering
			\caption*{Input}
			\includegraphics[width=1\linewidth,trim={4.5cm 5cm 4.5cm 1cm},clip]{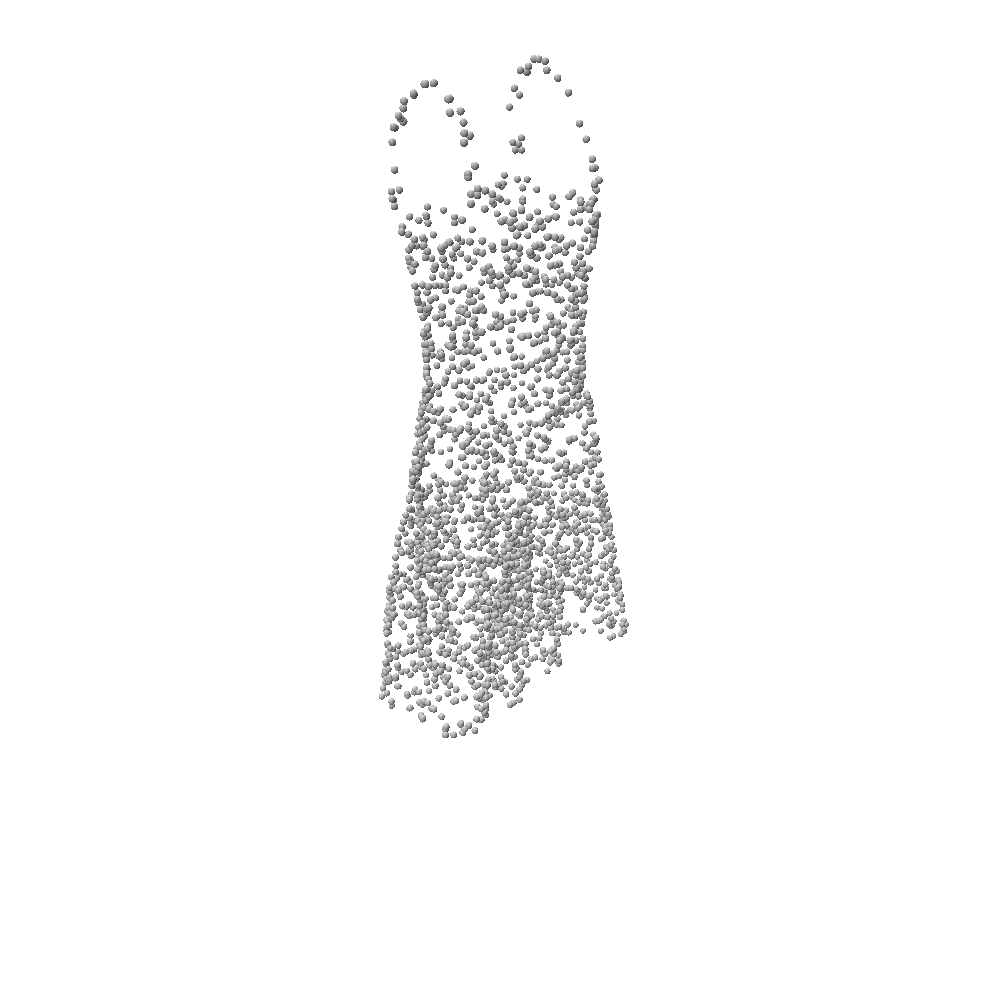}
            \includegraphics[width=1\linewidth,trim={3cm 4.5cm 3cm 5cm},clip]{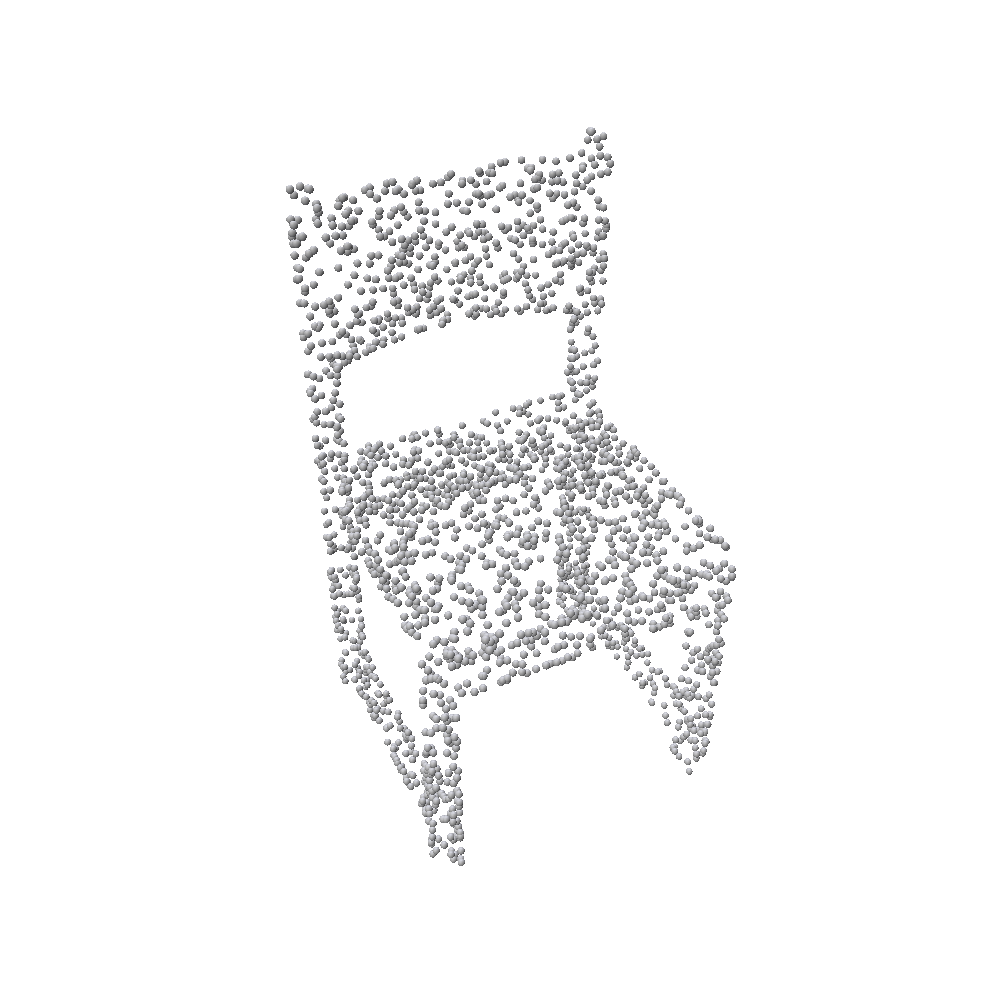}
		\end{subfigure}
        \caption{Artifacts of SCAR Violation.}
        \label{fig:artifacts}
\end{figure}

\par
As depicted in Fig.~\ref{fig:artifacts}, minimal neural atlas suffers from unintended holes on the reconstructed surface. We attribute such artifacts to the severe violation of the SCAR assumption, which is mainly caused by imperfect modeling of the target surface and 
non-matching sampling distribution between the target and maximal surface.

\subsection{Occupancy Rate}
\label{subsec:results:occupancy}

\begin{table}[t]
\centering
\footnotesize
\caption{Occupancy Rates in Surface Reconstruction Experiment.}
\label{table:results:occupancy}

\setlength\tabcolsep{4pt}
\begin{tabular}{@{}lccclccc@{}}
\toprule
\multicolumn{1}{c}{\multirow{2}{*}{\begin{tabular}[c]{@{}c@{}}Surface\\ Representation\end{tabular}}} & \multicolumn{3}{c}{CLOTH3D++} & \phantom{.} & \multicolumn{3}{c}{ShapeNet} \\ \cmidrule(lr){2-4} \cmidrule(lr){6-8} 
\multicolumn{1}{c}{} & 1 Chart & 3 Charts & 25 Charts &  & 1 Chart & 3 Charts & 25 Charts \\ \midrule
Ours w/o $\mathcal{L}_{dist}$ & 94.51 & 96.78 & 94.82 &  & 81.84 & 85.31 & 84.51 \\
Ours & 91.47 & 96.37 & 98.20 &  & 69.50 & 77.48 & 86.50 \\ \bottomrule
\end{tabular}
\end{table}

\par
Table~\ref{table:results:occupancy} shows the mean parametric domain occupancy rates of minimal neural atlas in the surface reconstruction experiment. In general, the occupancy rates are fairly high, which allows for the efficient extraction of the reconstructed surface point cloud and mesh. Furthermore, it can also be observed that the occupancy rate generally increases as the number of charts increases, especially when metric distortion is regularized. This may be attributed to the increased flexibility in forming a cover of the target surface as the number of charts increases.

\subsection{Ablation on Training UV Sample Size}
\label{subsec:results:sample_size}

\begin{table}[t]
\centering
\scriptsize
\caption{Effect of Training UV Sample Size.}
\label{table:ablation:sample_size}

\begin{tabular}{@{}lccclcclclc@{}}
\toprule
\multicolumn{1}{c}{\multirow{3}{*}{\begin{tabular}[c]{@{}c@{}}Surface\\ Representation\end{tabular}}} & \multirow{3}{*}{\begin{tabular}[c]{@{}c@{}}Training UV\\ Sample Size\end{tabular}} & \multicolumn{2}{c}{Point Cloud} & \phantom{} & \multicolumn{2}{c}{Mesh} & \phantom{} & \multirow{3}{*}{\begin{tabular}[c]{@{}c@{}}Metric\\ Distortion\end{tabular}} \multirow{3}{*}{\begin{tabular}[c]{@{}c@{}}$\downarrow$\end{tabular}} & \phantom{} & \multirow{3}{*}{\begin{tabular}[c]{@{}c@{}}Occupancy\\ Rate\end{tabular}} \multirow{3}{*}{\begin{tabular}[c]{@{}c@{}}$\uparrow$\end{tabular}} \\ \cmidrule(lr){3-4} \cmidrule(lr){6-7}
\multicolumn{1}{c}{} &  & CD, $10^{-4} \downarrow$ & F$@1\% \uparrow$ &  & CD, $10^{-4} \downarrow$ & F$@1\% \uparrow$ &  &  &  &  \\ \midrule
\multirow{3}{*}{DSP} & 2500 & 10.85 & 76.29 &  & 12.41 & 74.60 &  & 0.3044 &  & - \\
 & 3333 & 10.71 & 76.50 &  & 12.53 & 74.70 &  & 0.3189 &  & - \\
 & 5000 & 10.79 & 76.39 &  & 11.98 & 74.85 &  & 0.4130 &  & - \\ \midrule
\multirow{3}{*}{Ours w/o $\mathcal{L}_{dist}$} & 2500 & 6.603 & 82.89 &  & 7.153 & 81.07 &  & 11.55 &  & 91.38 \\
 & 3333 & 6.405 & 83.68 &  & 6.996 & 81.75 &  & 9.516 &  & 89.22 \\
 & 5000 & 6.266 & 84.04 &  & 6.875 & 82.22 &  & 10.23 &  & 85.31 \\ \midrule
\multirow{3}{*}{Ours} & 2500 & 6.866 & 82.15 &  & 7.412 & 80.39 &  & 2.441 &  & 87.85 \\
 & 3333 & 6.607 & 82.87 &  & 7.131 & 81.15 &  & 2.403 &  & 83.43 \\
 & 5000 & 6.311 & 83.63 &  & 6.761 & 82.23 &  & 2.189 &  & 77.48 \\ \bottomrule
\end{tabular}
\end{table}

\par
While the UV sample size used for training is typically chosen to be the same as the target point cloud size (2,500 in our experiments), we show in Table~\ref{table:ablation:sample_size} that a larger sample size favors our proposed representation since it leads to better reconstructions and lower distortions when explicitly regularized. It is also important to note that DSP is rather invariant to the increase in training UV sample size. We attribute this to the added influence of occupancy rate to the training, which is absent in other works.

\subsection{Hyperparameter Sensitivity Analysis}
\label{subsec:results:sensitivity}

\begin{table}[t!]
\centering
\scriptsize
\caption{Sensitivity of the Number of Positional Encoding Octaves.}
\label{table:sensitivity:octaves}

\begin{tabular}{@{}cccclcclccccc@{}}
\toprule
\multirow{3}{*}{\begin{tabular}[c]{@{}c@{}}No. of\\ Octaves\end{tabular}} & \phantom{.} & \multicolumn{2}{c}{Point Cloud} & \phantom{..} & \multicolumn{2}{c}{Mesh} & \phantom{..} & \multicolumn{3}{c}{Distortion} & \phantom{.} & \multirow{3}{*}{\begin{tabular}[c]{@{}c@{}}Occupancy\\ Rate\end{tabular}} \multirow{3}{*}{\begin{tabular}[c]{@{}c@{}}$\uparrow$\end{tabular}} \\ \cmidrule(lr){3-4} \cmidrule(lr){6-7} \cmidrule(lr){9-11}
 & & CD, $10^{-4} \downarrow$ & F$@1\% \uparrow$ &  & CD, $10^{-4} \downarrow$ & F$@1\% \uparrow$ &  & Metric $\downarrow$ & Conformal $\downarrow$ & Area $\downarrow$ & &  \\ \midrule
4 & & 6.381 & 83.34 &  & 6.935 & 81.68 &  & 2.100 & 0.7014 & 0.2295 & & 79.07 \\
6 & & 6.311 & 83.63 &  & 6.761 & 82.23 &  & 2.189 & 0.7094 & 0.2521 & & 77.48 \\
8 & & 6.394 & 83.46 &  & 6.882 & 81.96 &  & 2.154 & 0.7056 & 0.2425 & & 77.55 \\ \bottomrule
\end{tabular}
\end{table}

\begin{table}[t!]
\centering
\scriptsize
\caption{Sensitivity of Minimum Interior Rate, $\eta$.}
\label{table:sensitivity:interior}

\begin{tabular}{@{}ccccclcclccccc@{}}
\toprule
\phantom{.} & \multirow{3}{*}{\begin{tabular}[c]{@{}c@{}}$\eta$\end{tabular}} & \phantom{..} & \multicolumn{2}{c}{Point Cloud} & \phantom{...} & \multicolumn{2}{c}{Mesh} & \phantom{...} & \multicolumn{3}{c}{Distortion} & \phantom{..} & \multirow{3}{*}{\begin{tabular}[c]{@{}c@{}}Occupancy\\ Rate\end{tabular}} \multirow{3}{*}{\begin{tabular}[c]{@{}c@{}}$\uparrow$\end{tabular}} \\ \cmidrule(lr){4-5} \cmidrule(lr){7-8} \cmidrule(lr){10-12}
 & & & CD, $10^{-4} \downarrow$ & F$@1\% \uparrow$ &  & CD, $10^{-4} \downarrow$ & F$@1\% \uparrow$ &  & Metric $\downarrow$ & Conformal $\downarrow$ & Area $\downarrow$ & &  \\ \midrule
& 30 &  & 6.318 & 83.67 &  & 6.787 & 82.26 &  & 2.187 & 0.7807 & 0.2517 &  & 76.94 \\
& 40 &  & 6.311 & 83.63 &  & 6.761 & 82.23 &  & 2.189 & 0.7094 & 0.2521 &  & 77.48 \\
& 50 &  & 6.326 & 83.55 &  & 6.764 & 82.17 &  & 2.193 & 0.7099 & 0.2525 &  & 77.96 \\ \bottomrule
\end{tabular}
\end{table}

\subsubsection{Number of Positional Encoding Octaves.}
\label{subsubsec:results:sensitivity:octaves}

\par
While we have demonstrated that it is critical to apply positional encoding on the input maximal point coordinates of $l_{\theta_k}$, Table~\ref{table:sensitivity:octaves} shows that the specific number of octaves adopted in the encoding does not significantly affect the overall performance of our representation.

\subsubsection{Minimum Interior Rate.}
\label{subsubsec:results:sensitivity:interior}

\par
As observed in Table~\ref{table:sensitivity:interior}, minimal neural atlas is also not overly sensitive to the specific minimum interior rate $\eta$ employed for estimating the label frequency $c$ and hence extracting the reconstructed surface.

\end{document}